\documentclass{article} 
\pdfoutput=1
\usepackage[preprint,nonatbib]{nips}
\usepackage{float}
\usepackage{latexsym} 
\usepackage[utf8]{inputenc} 
\usepackage[T1]{fontenc}    

\usepackage{amsmath}
\usepackage{amsthm}
\usepackage{multirow,amssymb,amsmath} 
\usepackage{wrapfig} 
\usepackage{verbatim,graphicx}
\usepackage{xspace} 
\usepackage{algorithm}

\usepackage{algpseudocode}
\usepackage{stmaryrd}
\usepackage{pgfgantt} 
\usepackage{multicol}
\usepackage{times}
\usepackage{color}
\usepackage{hyperref}       
\usepackage{url}            
\usepackage{booktabs}       
\usepackage{amsfonts}       
\usepackage{nicefrac}       
\usepackage{microtype}      %
\usepackage{xcolor}         
\usepackage{subcaption}
\usepackage{enumitem}
\usepackage[title]{appendix}
\definecolor{grey}{rgb}{0.2,0.2,0.5}
\definecolor{dgrey}{rgb}{0.2,0.2,0.5}

\def\scH{\mathcal{H}}

\def\RR{\mathbb{R}}

\def\XX{\mathcal{X}}
\def\FF{\mathcal{F}}
\def\SG{\mathcal{G}}
\def\SH{\mathcal{H}}

\def\betab{\vspace{-0.10in}\begin{tabbing} 
xxx\=xxxx\=xxxxxxxxxxxxxxxxx\=xxxxxxxxxxxxxxxxxxxxxxxx\=xxxxxxxxx\= \kill} 
\def\entab{\end{tabbing}\vspace{-0.12in}}

\newcommand{\norm}[1]{\|#1\|}

\graphicspath{{./figs/}}

\newcommand{\xhdr}[1]{\vspace{0em}\noindent{{\bf #1.}}}
\newcommand{\subscript}[2]{$#1 _ #2$}
\newcommand{\eq}[1]{\begin{equation}\label{#1}}
\newcommand{\Expectation}[2]{\mathbb{E}_{#1}[#2]}
\newcommand{\methodName}{{TGCR}}
\newcommand{\methodNamenl}{{nlTGCR}}
\newcommand{\en}{\end{equation}}
\newcommand{\inner}[2]{\langle #1, #2\rangle}
\def\nref#1{(\ref{#1})} 
\def\RR{\mathbb{R}}

\def\half{\frac{1}{2}}
 
\def\Tr{\text{tr}}
\def\Span{\mbox{Span}}

\newcommand\inv[1]{#1\raisebox{1.15ex}{$\scriptscriptstyle-\!1$}}
\newtheorem{theorem}{Theorem}[section]

\newtheorem*{remark}{Remark}
\newtheorem{proposition}{Proposition}
\newtheorem{lemma}[theorem]{Lemma}
\newtheorem{definition}{Definition}

\providecommand{\norm}[1]{\lVert#1\rVert}
\title{An Efficient Nonlinear Acceleration method that Exploits Symmetry of  the Hessian}

\author{%
   Huan He\\
   Harvard University\\
   \texttt{huan\_he@hms.harvard.edu}\\
   \And
   Shifan Zhao\\
   Emory University \\
   \texttt{szhao89@emory.edu} \\
   \And
   Ziyuan Tang\\
   University of Minnesota \\
   \texttt{tang0389@umn.edu} \\
   \AND
   \quad Joyce C Ho \\
   Emory University  \\
   \texttt{joyce.c.ho@emory.edu} \\
   \And
   Yousef Saad\\
   University of Minnesota \\
   \texttt{saad@umn.edu} \\
   \And
   Yuanzhe Xi \\
   Emory University  \\
   \texttt{yuanzhe.xi@emory.edu} \\
}
\begin{document}
\maketitle
%

%





\begin{abstract}
Nonlinear acceleration methods are powerful techniques to speed up fixed-point iterations. However, many acceleration methods require storing a large number of previous iterates and this can become impractical if computational resources are limited. In this paper, we propose a nonlinear Truncated Generalized Conjugate Residual method (nlTGCR) whose goal is to exploit the symmetry of the Hessian to reduce memory usage. The proposed method can be interpreted as either an inexact Newton or a quasi-Newton method. We  show that,   with the help of global  strategies like residual check techniques, nlTGCR can converge globally for general nonlinear problems and that under mild conditions, nlTGCR is able to achieve superlinear  convergence.  We further analyze the convergence of nlTGCR in a stochastic setting. Numerical results demonstrate the superiority of nlTGCR when compared with several other competitive baseline  approaches on a few problems. Our code will be available in the future. 

\end{abstract}
\def\scH{\mathcal{H}}
\def\diag{\mbox{diag\,}}
\def\Diag{\mbox{Diag\,}}

\section{Introduction}
In this paper, we consider solving the fixed-point problem: 
\begin{equation}\label{eq:fp}
\text { Find } x \in \mathbb{R}^{n} \text { such that } x=H(x) \text {. }
\end{equation}
This problem has received a surge of interest due to its wide range of applications in mathematics, computational science and engineering. 
Most optimization algorithms are iterative, and their goal is to find a related  fixed-point   of the form \eqref{eq:fp}, where  $H: \mathbb{R}^{n} \rightarrow \mathbb{R}^{n}$
is the iteration mapping which can 
potentially be nonsmooth or noncontractive. When the optimization problem is convex, $H$ is typically nonexpansive, and the solution set of the fixed-point problem is the same as that of the original optimization problem, or closely related to it. Consider the simple fixed-point iteration $x_{k+1} = H(x_k)$
which produces a sequence of iterates $\{x_0, x_1, \cdots, x_{K}\}$.
When the iteration converges, its limit is a  fixed-point, i.e.,
$x^{\ast} = H(x^{\ast})$. However, an issue with fixed-point iteration
is that it does not always converge, and when it does, it might reach the
limit very slowly.

To address this issue, a number of acceleration methods have been proposed and studied over the years, such as the reduced-rank extrapolation (RRE) \cite{doi:10.1137/1029042}, minimal-polynomial extrapolation (MPE) \cite{doi:10.1137/0713060}, modified MPE (MMPE) \cite{8145128}, and the vector $\epsilon$-algorithms \cite{doi:10.1137/140957044}.  Besides these algorithms, Anderson Acceleration (AA) \cite{Anderson65} has received enormous
recent attention due to its nice properties and its success in
 machine learnin  applications \cite{https://doi.org/10.48550/arxiv.1805.09639,AARL,Shi2019RegularizedAA,d_Aspremont_2021,Sun2021DampedAM, 9763953,he2022gdaam,wei2022a}. In practice, since computing the 
Hessian of the objective function is commonly difficult or even unavailable, AA can be seen as a practical alternative to Newton’s method \cite{newtonCT}. Also, compared with the classical
iterative methods such as the nonlinear conjugate gradient (CG) method \cite{Hager2005ASO}, no
line-search or trust-region technique is performed in AA, and this is
a big advantage in large-scale unconstrained
optimization. Empirically, it is observed that AA is quite successful in
accelerating convergence. We refer readers to \cite{shanks} for a recent survey of acceleration methods. 

However, classical AA has one undesirable disadvantage in that it is expensive
in terms of memory as well as computational cost, especially in a nonconvex
stochastic setting, where only sublinear convergence can be expected when only
stochastic gradients can be accessed \cite{doi:10.1137/1027074}.
In light of this, a number of variants of AA have been proposed which aim at improving its performance and robustness (e.g.,  
\cite{8682962,wei2022a,AAsto,
https://doi.org/10.48550/arxiv.1805.09639,https://doi.org/10.48550/arxiv.1808.03971}). 
The above-cited works focus on 
improving the convergence behavior of AA, but they do not consider reducing the memory cost. In machine learning, we often encounter practical situations where the number of parameters is quite large and for this reason, it is not practical to use a large number of vectors in the acceleration methods. It is not clear whether or not the symmetric structure of the Hessian can be exploited in a scheme like AA to reduce the memory cost while still maintaining the convergence guarantees. In this paper, we will demonstrate
how this can be accomplished with a new algorithm that is superior to AA in practice.

\xhdr{Our contributions} 
This paper  develops a nonlinear acceleration method, nonlinear Truncated Generalized Conjugate
Residual method (nlTGCR), that takes advantage of symmetry. This work is motivated by the observation that the Hessian of a nonlinear function, or  the Jacobian of a gradient of a mapping, $f$ is symmetric and therefore more effective, conjugate gradient-like schemes can be exploited.

We demonstrate that nonlinear acceleration methods can benefit from the symmetry property of the Hessian. In particular, we study both linear and nonlinear problems and give a systematic analysis of \methodName~ and \methodNamenl. {We show that \methodName~ is efficient and optimal for linear problems.} 
By viewing the method from the angle of an inexact Newton approach,
we also show that adding a few  global convergence strategies ensures that 
 \methodNamenl~can achieve global convergence guarantees.

We complement our theoretical results with numerical simulations on several different problems. The experimental results demonstrate advantages of our methods. To the  best of our knowledge, this is still the first work to investigate and improve the AA dynamics by exploiting symmetry of the Hessian.

\xhdr{Related work}
Designing efficient optimization methods has received much attention. Several recent works \cite{adahessian,subnetwon2,hfdl,subnewton,derezinski2021newtonless} consider second order optimization methods that employ sketching or approximation techniques. { Different from these approaches, our method is a first-order method that utilizes symmetry of the Hessian instead of constructing it. } A variant of inexact Newton method was proposed in \cite{newtonMR} where
the least-squares sub-problems are solved approximately using Minimum Residual method. Similarly, a new type of quasi Newton symmetric update \cite{quasisb}  uses several secant equations in a least-squares sense. These approaches have the same goal as ours. However, they are more closely related to a secant or a multi-secant technique, and as will be argued it does a better job of capturing the nonlinearity of the problem. \cite{wei2022a} proposed a short-term AA algorithm that is different from ours because it is still based on the parameter sequence instead of the gradient sequence and does not exploit symmetry of the Hessian.  

\section{Background}

\subsection{Extrapolation, acceleration, and the Anderson Acceleration procedure}\label{sec:AA} 
Consider a general fixed-point problem and the associated fixed-point iteration as shown in \eqref{eq:fp}. Denote by 
$r_j = H(x_j)- x_j$ the \emph{residual} vector at the
$j$th iteration. 
Classical  extrapolation methods 
including RRE, MPE and the vector $\epsilon$-Algorithm, have been 
designed to accelerate the convergence of the original sequence
by generating a new and independent sequence  of the form: 
$t_j^{(k)}=\sum_{i=0}^{k} \alpha_{i} x_{j+i}$.
An important characteristic of these classical extrapolation methods is
that the two sequences  are not mixed in the sense that no accelerated item $t_j^{(k)}$, is used to produce the iterate $x_j$. These \emph{extrapolation} methods must be
distinguished from \emph{acceleration} methods such as the AA procedure  which aim at generating their own sequences to find a fixed point of a certain mapping $H$. 

AA was originally designed to solve a system of nonlinear equations written in the form $F(x) = H(x)-x=0$
\cite{Anderson65,WalkerNi2011,thni,a4}.  Denote $F_{i}=F(x_i)$. AA starts with an
initial $x_0$ and sets  $x_1=H(x_0)=x_0+\beta F_0$, where
$\beta >0$ is a parameter.  At step $j>1$ we define
${X}_j=[x_{j-m},\ldots, x_{j-1}],$   and
$\bar{F}_j=[ F_{j-m}, \ldots, F_{j-1}]$ along with the differences:
\begin{equation}
\label{eqn:dfdx}
\begin{aligned}
\mathcal{X}_j &= [\Delta x_{j-m}\;\cdots\;\Delta x_{j-1}]\in \RR^{n\times m},\\
\mathcal{F}_j &= [\Delta F_{j-m}\;\cdots\;\Delta F_{j-1}]\in \RR^{n\times m}.
\end{aligned}
\end{equation}
We then  define the next AA iterate as follows: 
    \begin{align}
        x_{j+1} &=  x_j+\beta F_j -(\XX_j+\beta \FF_j)\ \theta^{(j)}  \quad \mbox{where:}  \label{eq:AA}  \\
\theta^{(j)} &=
\text{argmin}_{\theta \in \mathbb R^{m}}\| F_j - \FF_j \ \theta \|_2   \label{eq:thetaj}.
    \end{align}

To define the next iterate in \nref{eq:AA} the algorithm
uses the term $F_{j+1} = F(x_{j+1})$ where $x_{j+1}$ is the current
accelerated iterate. AA belongs to the class of \emph{multi-secant methods}.
Indeed, the approximation \nref{eq:AA} can be written as: 
\eq{eq:AndQN} 
\begin{split}
x_{j+1} &= x_j - [- \beta I  + (\XX_j+\beta\FF_j)(\FF_j^T\FF_j)^{-1}\FF_j^T ] F_j \\ &\equiv  x_j - G_j F_j .    
\end{split}
\en 
Thus,  $G_{j}$ can be seen as an update to
the (approximate) inverse  Jacobian $ G_{j-m} = -\beta I$ \eq{eq:msec}
G_{j} = G_{j-m} + (\XX_j  - G_{j-m} \FF_j)(\FF_j^T\FF_j)^{-1}\FF_j^T, 
\en  and  is the minimizer of $\| G_j + \beta I  \|_F$ under the \emph{multi-secant condition}
of type II \footnote{Type I Broyden conditions involve
  approximations to  the Jacobian,
 while type II conditions  deal with the inverse Jacobian.}
\eq{eq:mscond}
G_j \FF_j = \XX_j.
\en
This link between AA and Broyden multi-secant type updates
was first unraveled by Eyert~\cite{eyert:acceleration96} and expanded upon in
\cite{FangSaad07}. 

\subsection{Inexact and quasi-Newton methods}
Given a nonlinear system of equations $F(x)=0$. Inexact Newton methods \cite{Dembo-al,Brown-Saad}, start with an initial guess $x_0$ and
  compute a sequence of iterates   as follows
  \begin{align} 
 &      \mbox{Solve}  & J(x_j) \delta_j   &\approx -F(x_j) &  \label {eq:inex}\\
 &       \mbox{Set}    & x_{j+1}  & = x_j + \delta_j & \label {eq:updt}
  \end{align}  
  Here, $J(x_j)$ is the Jacobian of $F$ at the current iterate $x_j$.
  In \nref{eq:inex}  the system is solved inexactly, typically by some iterative method. In quasi-Newton methods \cite{quasi:77,quasi:77,quasisb}, the inverse of the Jacobian
is approximated progressively. Because it is the inverse Jacobian that is
approximated, the method is akin to Broyden's second 
(or type-II)  update method. This method replaces
Newtons's iteration: $x_{j+1}  = x_j - \inv{DF(x_j)}  F(x_j)$ with 
$     x_{j+1} = x_j - G_j F(x_j) $  
  where $G_j $  approximates the inverse of the Jacobian $DF(x_j)$ at $x_j$
  by the  update formula $G_{j+1} = G_j + (\Delta x_j - G_j \Delta F(x_j)) v_j^T $
  in which $v_j$ is defined in different ways see \cite{FangSaad07} for details.

\section{Exploiting symmetry}
In the following, we specifically consider the case where the nonlinear mapping
$F$ is the gradient of some
objective function $\phi:\mathbb{R}^n\rightarrow\mathbb{R}$ to be minimized, i.e., $$F(x) = \nabla \phi(x). $$
In this situation, the Jacobian of $F$ becomes $\nabla^2 \phi$  the Hessian of $\phi$. An obvious observation here is that the symmetry of the Hessian is not taken into account in the approximate inverse Hessian update formula
  \nref{eq:msec}. This has only been considered
  in the literature (very) recently (e.g.,
  \cite{Boutet2020,scieur21a,Boutet2021}).  In a 1983 report,
  \cite{Schnabel83} showed that the matrix $G_j$ obtained by a multi-secant method
  that satisfies the secant condition \nref{eq:mscond} is symmetric iff the
  matrix $\XX_j^T \FF_j$ is symmetric. It is possible to explicitly force
  symmetry by employing generalizations of the symmetric versions of
  Broyden-type methods. Thus, the authors of \cite{Boutet2020,Boutet2021}  developed a multisecant version of the Powell Symmetric Broyden (PSB)
  update due to Powell \cite{Powell70} while the article \cite{scieur21a}
  proposed a symmetric multisecant method based on the popular
  Broyden-Fletcher-Goldfarb-Shanno (BFGS) approach as well as the
  Davidon-Fletcher-Powell (DFP) update. However, there are a number of  issues with the symmetric versions of multisecant updates, some of which are  discussed in~\cite{scieur21a}. 
  
 We observe that when we are close to the limit, the condition
   $\XX_j^T \FF_j =  \FF_j^T \XX_j $ is nearly satisfied.
  This is because if $x^{\ast}$ is the limit with 
  $F(x^{\ast}) = 0$ we can write
  \begin{equation}
      \begin{aligned}
              F(x_k) - F(x_{k-1}) &= 
    [ F(x_k) - F(x^{\ast})] \\
    &\quad - [F(x_{k-1}) - F(x^{\ast})] \\
    & \approx \nabla^2 \phi(x^{\ast}) ( x_k \ - \ x_{k-1}) . 
      \end{aligned}
  \end{equation}
  This translates to $\FF_j \approx  \nabla^2 \phi(x^{\ast}) \XX_j $
  from which it follows that
  $ \XX_j^T \FF_j \approx  \XX_j^T   \nabla^2 \phi(x^{\ast}) \XX_j $ which is a
  symmetric matrix under mild smoothness conditions on $\phi$.
  Therefore, the issue of symmetry can be mitigated if we are able to develop nonlinear acceleration methods that take advantage of 
  near-symmetry.

  \subsection{The linear case: Truncated GCR (\methodName)}
  \label{subsec:linearized update}
  We first consider solving the linear system $Ax=b$ with a general matrix   $A$. The Generalized
  Conjugate Residual (GCR) algorithm, see, e.g., 
  \cite{Eis-Elm-Sch,Saad-book2}, solves
  this linear system by
  building a sequence of search directions $p_i$, for $i=0, \cdots, j$ at step $j$
  so that the vectors  $A p_i$ are orthogonal to each other.
 With this property it is easy to generate iterates
 that minimize the residual at each step,
  and this leads to GCR,  see \cite[pp 195-196]{Saad-book2} for details.

Next we will make two changes to GCR. First, we will develop a truncated version in which any given $A p_j$ is orthogonal to
the previous $m$  $Ap_i$'s only. This is dictated by practical
considerations, because keeping all $A p_i$ vectors may otherwise require too much memory.
Second, we will keep a set of vectors
for the $p_i$'s and another set for the vectors $v_i \equiv A p_i$, for $i=1, \cdots, j$ at step $j$ in order to avoid unnecessary additional products of the matrix $A$ with vectors. The
Truncated GCR (\methodName) is summarized in Algorithm \ref{alg:tgcr}.

  \begin{algorithm}[H]
    \centering
    \caption{\methodName~(m)}\label{alg:tgcr}
    \begin{algorithmic}[1]
  \State \textbf{Input}: Matrix $A$, RHS $b$,
  initial  $x_0$. \\
  Set $r_0 \equiv b-Ax_0$; $v = A r_0$; 
  \State $v_0 = v/\| v \|$;   $p_0 = r_0 / \| v \|$; 
\For{$j=0,1,2,\cdots,$ Until convergence} 
\State  $\alpha_j = (r_j, v_j) $
\State $x_{j+1} = x_j + \alpha_j p_j$
\State $r_{j+1} = r_j - \alpha_j v_j$
\State $p = r_{j+1}$;   $ v = A p $;   
\State $i_0 = \max(1,j-m+1)$  
\For{$i=i_0:j$} 
\State $\beta_{ij} :=  (v, Ap_i) $
\State $p := p - \beta_{ij} p_i$;
\State  $v := v - \beta_{ij} v_i$; 
\EndFor
\State $p_{j+1} :=p /\| v\|$ ; \qquad  $v_{j+1} :=v/\|v\|$ ; 
\EndFor
\end{algorithmic}
\end{algorithm}

With $m = \infty $ we obtain the non-restarted GCR method, which is equivalent
to the non-restarted (i.e., full) GMRES.  
However, when $A$ is symmetric, but not necessarily symmetric positive definite, then \methodName~(1) is identical with  \methodName~(m) in exact arithmetic. This leads to big savings both in terms of memory and in computational costs. 
\begin{theorem}
When the coefficient matrix $A$ is symmetric, \methodName~(m) generates the same iterates as \methodName~(1) for any $m>0$.
In addition, when $A$ is positive definite, the $k$-th residual vector $r_k =  b - Ax_k$
satisfies the following inequality
where $\kappa$ is the spectral condition number of $A$:
\eq{eq:convCR}
\| r_k \| \le 2 \left[\frac{\sqrt{\kappa} -1}{\sqrt{\kappa} +1}
\right]^k \|r_0 \| . 
\en
\label{thm:m1}
\end{theorem}

  \subsection{The nonlinear case: \methodNamenl~}
  Assume now that we want to solve the nonlinear problem $
  F(x) = 0.$ We need to make three major changes to Algorithm~\ref{alg:tgcr}.  First, any residual is now the negative of $F(x)$ so Line~2  and Line~7 must be replaced by 
 $r_0 = - F(x_0)$ and $r_{j+1} = - F(x_{j+1})$, respectively.  In
 addition, we originally need to calculate the products $A r_0 $ and $A p$ in Line~2 and
 Line~8 respectively. Here $A$ needs to be replaced by  the Jacobian $J(x_j)$ of $F$ at the
 current iterate. We also use the notation
 $P_j \equiv [ p_{i_0}, \cdots, p_{j}]$,
 and  
 $V_j \equiv [ J(x_{i_0}) p_{i_0}, \cdots,  J(x_{j}) p_{j}]$.  
  The most important
 change is in lines 5-6 where $\alpha_j$ of
 Algorithm~\ref{alg:tgcr} needs to be replaced by a vector $y_j$.  This is because when we write the linear model used in the form of an inexact Newton method:
 \begin{equation}
 \label{eq:JP}
     \begin{aligned}
         F(x_j
  + P_j y) &\approx F(x_j) + [J] P_j y \quad \mbox{where} \\ 
 [J] P_j &\equiv [
  J(x_{i_0}) p_{i_0}, \cdots, J(x_{j}) p_{j} ] = V_j .
     \end{aligned}
 \end{equation}

 The projection method that minimizes the norm
  $\| F(x_j) + [J] P_j y \| = \| F(x_j) + V_j y \| $ of the right-hand side
  determines $y$ in such a way that
  \begin{equation}
  \label{eq:opt}
  \begin{aligned}
  F(x_j) + V_j y \perp \Span
  \{V_j\}  & \rightarrow  (V_j)^T[F(x_j) + V_j y] = 0 \\ & \rightarrow  y = V_j^T
  r_j     
  \end{aligned}
  \end{equation}
  where it is assumed the $v_i$'s are fully orthogonal.
  Note that
  in the linear case, it can be shown that $V_j^T r_j$ has only one nonzero
  component when one assumes that the vectors $J(x_i) p_i$ are fully orthogonal,
  i.e., that $i_0 =1$ always. The nonlinear version of \methodName~(m) is summarized in Algorithm \ref{alg:nltgcr} where  the indication `Use Frechet' means that the
vector $v = J(x) u$ is to be computed as  
$v = (F(x + \epsilon u) - F(x))/\epsilon$ for some small
$\epsilon$.

    \begin{algorithm}[H]
    \centering
    \caption{\methodNamenl~(m)}\label{alg:nltgcr}
    \begin{algorithmic}[1]
  \State \textbf{Input}: $F(x)$,   initial  $x_0$. \\
  Set $r_0 = - F(x_0)$.
  \State Compute  $v = J(x_0) r_0$; \ (Use Frechet)
  \State $v_0 = v/ \| v \| $, $p_0 = r_0/ \| v \| $;
  \For{$j=0,1,2,\cdots,$ Until convergence} 
\State  $y_j = V_j^T r_j $ 
\State $x_{j+1} = x_j + P_j y_j$
\State $r_{j+1} = -F(x_{j+1}) $ 
\State Set: $p := r_{j+1}$;
\State $i_0 = \max(1,j-m+1)$
\State Compute $v =  J(x_{j+1}) p$ (Use Frechet) 
\For{$i=i_0:j$}
\State $\beta_{ij} :=  (v, v_i) $
\State $p := p - \beta_{ij} p_i$
\State $v := v - \beta_{ij} v_i$
\EndFor
\State $p_{j+1} :=p /\| v\|$ ; \qquad  $v_{j+1} :=v/\|v\|$ ; 
\EndFor
\end{algorithmic}
\end{algorithm}

\begin{remark}
\methodNamenl~(m) requires 2 function evaluations per step: one in Line 8 and the
other in {Line~11.}
In the situation when computing the Jacobian is inexpensive, then one can compute $J p$ in Line~11 as a matrix-vector product and this will reduce the number of function
evaluations per step from 2 to 1. {The inner loop in Line 12-16 corresponds to exploiting symmetry of Hessian.} At a given step, \methodNamenl~(m) attempts to approximate a Newton step:
$  x_{j+1} = x_j + \delta $ where
$\delta $ is an approximate solution to $ J(x_j) \delta + F(x_j) = 0$. 
\end{remark}

\xhdr{High-Level Clarification} 
At this point, one might ask the question: why not just use an inexact Newton
method whereby the Jacobian system is 
solved with the \emph{linear} GCR or TGCR method? This is where AA
provides an interesting insight on some weaknesses of Newton-Krylov method. A Newton-Krylov method generates a Krylov subspace $\Span \{r_0, J r_0, \cdots, J^k r_0 \}$ at a current iterate -- say $K = x_0$ -- (so $J \equiv J(x_0) \equiv DF(x_0)$) and tries to minimize
$F(x_0 + \delta)$ where $\delta \ \in \ K$, by exploiting the linear model:
$F(x_0 + \delta) \approx F(x_0) + J \delta$. If we generate a basis $V = [v_1, \cdots v_k]$ of $K$
and express $\delta $ as $\delta = V y$ then we would need to minimize
$\| F(x_0) + J V y \| $ which is a small least-squares problem. One usually adds to this
a global convergence strategy, e.g., a linesearch or a trust-region technique
to produce the next iterate $x_1$. 
The problem with this approach is this: \emph{the approximate solution obtained after $k$ steps of a Krylov subspace approach is based on the Jacobian at the initial point $x_0$.}
The intermediate calculation is entirely
linear and based on $J(x_0)$. It is not exploited in any way to produce
intermediate (nonlinear) iterates which in turn could be used to produce more
accurate information on some local Jacobian.  In contrast, a method like
AA (or in fact any of the secant or multisecant methods) will
do just this, i.e., it will use information on the nonlinear mapping
near the most recent approximation to produce the new iterate.  This
distinction is rather important although if the problem is nearly linear, then
it could make little difference.

{In \methodNamenl~, we try to improve on the Newton-Krylov approach, since our starting point is TGCR which is generalized to nonlinear problems. We also take the viewpoint of
improving on AA or multisecant methods by not relying on the
approximation
$F(x_{j+1})- F(x_j) \approx J (x_{j+1}-x_j)  $ mentioned above. This is achieved by adopting the projection viewpoint. Instead of
minimizing $\| F(x_0) + J  P y\|$ as in the inexact Newton mode,  we would like to now minimize
$\| F(x_k) + J  P y\|$ where $x_k$ is the most recent iterate. This initial idea leads to difficulty since there is not one but 
several $J$ at previous points and a single one of them will not be satisfactory.
Thus, we have a few directions $p_i$ just like the differences $\Delta x_i$ in Anderson,
\emph{but now each $p_i$ will lead to a $J(x_i) p_i$ which - unlike in AA - is accurately
  computed and then saved.}
This feature is what we believe makes a difference in the performance of the algorithm, although
this is something that is rather difficult to prove theoretically. We leave it as future work. Overall, the method described in this paper mixes a number of ideas coming from different horizons. A further high-level discussion and detailed complexity analysis are provided in Appendix \ref{Sec:add}. }

Next, we analyze two possible versions of \methodNamenl~ in the next two sections. In what follows we 
assume that all the $J p_i$'s are computed exactly.

\subsubsection{Linearized update version} 
First, we consider a variant of Algorithm \ref{alg:nltgcr} which we call the ``linearized update version'' -- whereby 
 in Line~8 we update the
residual by using the linear model, namely, we replace Line~8 by
its linear analogue: 8a:   \  $ r_{j+1} = r_j - V_j y_j$. \  {In addition, the matrix-vector product in Line~11
is performed with $J(x_0)$ instead of $J(x_{j+1})$.} 
When $F$ is linear, 
it turns out that $y_j$ has only one nonzero component, namely the last one and this will yield the 
standard truncated GCR algorithm. 
Assume that we perform $k$ steps of the algorithm to produce $x_k$, i.e., that Line~5 is replaced by
\texttt{``for $j = 0, 1, 2, \cdots, k$ do''}.
Then the algorithm is exactly  equivalent to
an \emph{inexact Newton method} in which GMRES (or GCR) is invoked to solve the
Jacobian linear system \cite{Brown-Saad}.
Indeed, in this situation Lines 4-15 of Algorithm~\ref{alg:tgcr} and Lines
5-17 of Algorithm~\ref{alg:nltgcr} are identical.
In other words, in Lines 5-17,
Algorithm~\ref{alg:nltgcr} performs $k$ steps of
the GCR algorithm  for approximately solving the linear systems 
$J(x_0) \delta = -F(x_0)$.
Note that while the update is written in progressive form
as $x_{j+1} = x_j + \alpha_j p_j$, \emph{the right-hand side does
not change during the algorithm and it is equal to $r_0 = - F(x_0)$.
}
In effect $x_k$ is updated from $x_0$ by adding a vector from the span of $P_k$. {See the related global convergence result shown in Theorem \ref{thm:linearizedtgcr} in the Appendix, for a version of this algorithm that includes a line-search.} A weakness of this linear update version is that the Jacobian is not evaluated at the most recent update but at $x_0$, which in practice is the iterate 
at each restart.

\subsubsection{Non-linear update version with residual check}
Next we consider the `nonlinear-update version' as described in Algorithm~\ref{alg:nltgcr}. This version explicitly enforces the
linear optimality condition of GCR, as represented by the Equation \nref{eq:opt}. In this section, we will analyze the convergence of \methodNamenl~ through the function $\phi(x) = \frac{1}{2}\norm{F(x)}^2$.

In order to prove the global convergence of \methodNamenl, we need to make a small modification to Algorithm \ref{alg:nltgcr} because
  as implemented in Algorithm~\ref{alg:nltgcr} $P_j$ is fixed and the solution obtained at this step may not necessarily satisfy the following \emph{residual check} condition which is often used in inexact Newton methods \cite{Dembo-al, Brown-Saad2,Eisenstat-Walker94} to guarantee the global convergence:
\eq{eq:eta2nd}
  \| F(x_j) + [J] P_j y \|  \le \eta  \| F(x_j) \|, 
  \en
 where $\eta < 1  $ is a parameter.
 
  The residual norm on the left-hand
  side of \nref{eq:eta2nd} is readily available at no additional cost and this can help devise globally converging strategies,
  by monitoring to what extent \nref{eq:eta2nd} is satisfied. If \eqref{eq:eta2nd} is not satisfied, we can either use a line-search technique \ref{alg:linesearch} or restart the process and take the next iterate as the output of the fixed point iteration mapping $H$. When the residual check condition is implemented after Line 8 in Algorithm \ref{alg:nltgcr}, we can prove the global convergence of \methodNamenl~ in the next theorem. {Similar global strategies have also been proposed in \cite{JunziBoyd20, inexactN,practicalIN,BFGS_e,hansasy,ouyangAA}. }
\begin{theorem}[Global convergence of \methodNamenl~ with  residual check] 
Assume $\phi$ is twice differentiable and $F(x)$ is L-lipschitz. If the residual check is satisfied $\norm{J(x_n) P_ny_n + F(x_n)} \leq \eta_n \norm{F(x_n)}$ where $0\leq\eta_n \leq \eta < 1$ and $J(x_n)$ is non-singular and its norm is bounded from above for all n, then $P_ny_n$ produced in line 7 of Algorithm \ref{alg:nltgcr} is a descent direction and the iterates $x_n$ produced by Algorithm \ref{alg:nltgcr}will converge to the minimizer $x^{*}$:
\begin{align*}
         \lim_{n\rightarrow \infty} \phi(x_n) = \phi(x^{*}) = 0.
\end{align*}

 \label{thm:inexact-Newton-globalmain}
 \end{theorem}
  In the next theorem, we prove that \methodNamenl~ can achieve superlinear and quadratic convergence under mild conditions. 
 \begin{theorem}[Superlinear and quadratic convergence of \methodNamenl~]
  With the same setting as Theorem \ref{thm:inexact-Newton-globalmain}. Assume $\nabla^{2}\phi$ is L-lipschitz. Consider a sequence generated by Algorithm \ref{alg:nltgcr} such that residual check is satisfied $\norm{J(x_n) P_ny_n + F(x_n)} \leq \eta_n \norm{F(x_n)}$ where $0\leq\eta_n \leq \eta < 1$. Moreover, if the following conditions hold
  \begin{align*}
      \phi(x_n +  P_ny_n) &\leq \phi(x_n) + \alpha  \nabla\phi(x_n)^TP_ny_n\\
      \phi(x_n +  P_ny_n) &\geq \phi(x_n) + \beta  \nabla\phi(x_n)^TP_ny_n
  \end{align*}
  for $\alpha <\frac{1}{2}$ and $\beta >\frac{1}{2}$. Then there exists a $N_s$ such that  $x_n \rightarrow x^{*}$ superlinearly for $n\geq N_s$ if $\eta_n\rightarrow 0$, as $n\rightarrow \infty$.  Moreover, if $\eta_n = O(\norm{F(x_n)}^2)$, the convergence is quadratic.
  \label{thm:super}
 \end{theorem}
{If the property of the function is bad (non-expansive/non-convex), it will be more difficult to satisfy the assumptions of Theorem \ref{thm:inexact-Newton-globalmain} and \ref{thm:super}. For example, in Theorem 3.2, the non-singularity and boundedness is required for $J(x)$. If the function does not satisfy the assumption, say, degenerate at a point, then the algorithm may not converge to a stationary point. }
{
\begin{remark}
This superlinear(quadratic) convergence of nlTGCR does not contradict with the linear convergence of TGCR shown in \ref{thm:m1}. \ref{thm:m1} is obtained from the equivalence between TGCR and CG in that short-term recurrence holds for symmetric matrix. Like CG, TGCR can still have a superlinear convergence rate. In practice, the second stage of convergence of Krylov Space methods is typically well defined by the theoretical convergence bound with $\sqrt{\kappa(A)}$ but may be super-linear, depending on a distribution of the spectrum of the matrix $A$ and the spectral distribution of the error.
\end{remark}
}

 Finally, we analyze the convergence of \methodNamenl~ when the gradient $F$ is subsampled. In the analysis, we make the following five assumptions.

 \paragraph{Assumptions for stochastic setting}
    $\mathbf{A_1:}$ The variance of subsampled gradients is uniformly bounded by a constant $C$, $\operatorname{tr}(Cov(F(x))) \leq C^2, \ \forall x$.

$\mathbf{A_2:}$ The eigenvalues of the Hessian matrix for any sample $|\SH| = \beta$ is bounded from below and above in Loewner order $\mu_{\beta}I \preceq J(x, \SH) \preceq L_{\beta}I$.
    Further more, we require there is uniform lower and upper bound for all subsmaples. That is, there exists $\hat{\mu}$ and $\hat{L}$ such that $0 \leq \hat{\mu} \leq \mu_{\beta} \quad \text{and} \quad L_{\beta} \leq \hat{L}< \infty, \quad \forall \beta \in \mathbb{N}.$ And the full Hessian is bounded below and above
    $\mu I \preceq J(x) \preceq LI,\quad \forall x.$

    $\mathbf{A_3:}$ Hessian is M-Lipschitz, that is
    $ \norm{J(x) - J(y)} \leq M\norm{x-y}, \quad \forall x, y$
    
    $\mathbf{A_4:}$The variance of subsampled Hessian is bounded by a constant $\sigma$.
    \begin{align}
        \norm{\Expectation{\SH}{(J(x;\SH) - J(x))}} \leq \sigma, \quad \forall x
    \end{align}
    $\mathbf{A_5:}$ There exists a constant $\gamma$ such that
   $
        \mathbb{E}[\norm{x_n - x^{*}}^2] \leq \gamma(\mathbb{E}[\norm{x_n - x^{*}}])^2.
   $

\begin{theorem}[{Convergence of stochastic version of \methodNamenl~}] 
Assume  $|\SH_n| = \beta \geq \frac{16\sigma^2}{\mu},~ \forall n$, residual check is satisfied for $\eta_n\leq \eta \leq \frac{1}{4L}$ and assumptions $A1-A5$ hold.
 The iterates generated by the stochastic version Algorithm \ref{alg:nltgcr} converge to $x^{*}$ if $\norm{x_k - x^{*}}\leq \frac{\mu}{2M\gamma}$ and
 \begin{align}
 \label{ineq:stochastic-bound}
     \mathbb{E}\norm{x_{n+1} - x^{*}} \leq \frac{3}{4}\mathbb{E}\norm{x_n - x^{*}}.
 \end{align}
 \end{theorem}

\subsection{Connections with other methods}
This section explores the connection between \methodNamenl~ with inexact Newton and AA. We provide the connection between \methodNamenl~ and quasi-Newton in \ref{sec:qnview}.

\xhdr{1) The inexact Newton viewpoint}
Inexact Newton methods minimize $\| F(x_0) + J(x_0) P_j y \|$
over $y$  by using some iterative method and  enforcing a condition like
\[
  \| F(x_0) + J(x_0) P_j y \|  \le \eta  \| F(x_0) \|
\]
where $\eta < 1  $ is a parameter, see, e.g.,
\cite{Dembo-al,Brown-Saad, Brown-Saad2,Eisenstat-Walker94}. In \methodNamenl, we are trying to solve a similar equation $$F(x_j) + J(x_j) \delta = 0$$  by
minimizing $ \| F(x_j) + [J] P_j y \| $. We can prove the following properties of nlTGCR.
\begin{proposition}
  As defined in  Algorithm~\ref{alg:nltgcr}, $\delta_j = x_{j+1} - x_j = P y_j $ minimizes
  $ \| F(x_j) + \delta \| $ over vectors of the form $\delta = V_j y$,
  where $y \ \in \ \RR^{n_j}$
  and $n_j = j-i_0+1$.
\end{proposition}

As  noted earlier,  in the linear case, the vector $y_j$ has only one nonzero component, namely the top one. In the general case, it is often observed that the other components are not zero but small.  
Let us then suppose that we replace the update
in Lines 6-7 by the simpler form: $\mu_j = v_j^T r_j$, and $x_{j+1} = x_j + \mu_j p_j$.
Then the   direction $\delta_j = x_{j+1} - x_j$  is a descent direction
for $\half \| F(x) \|^2$.

\begin{proposition}
  Assume that  $J(x_j)$ is nonsingular and that $\mu_j  \equiv   v_j^T r_j   \ne 0$.
  Then $\delta_j = \mu_j p_j$ is a descent direction for the function
  $\half \| F(x) \|^2 $ at $x_j$.
\end{proposition}

\xhdr{2) The quasi-Newton viewpoint}
It is also possible to view the algorithm from the alternative angle of
a quasi-Newton approach  instead of  inexact Newton.
In \methodNamenl, the approximate inverse Jacobian $G_j$ at step $j$ is equal to
\eq{eq:Gj}
G_{j} = P_j V_j^T .
\en 
If we apply this to the vector $v_j$ we get
$
  G_{j} v_j = P_j V_j^T v_j =  p_j =   \inv{J(x_j)} v_j .
$ 
So $G_j$ inverts $J(x_j)$ exactly when applied to $v_j$. It therefore satisfies 
the \emph{secant} equation (\cite[sec. 2.3]{FangSaad07}) 
\eq{eq:msecant1} 
  G_j v_j = p_j . 
  \en
This is equivalent to the secant condition
$G_j \Delta f_j = \Delta x_j$ used in Broyden's second update method.

In addition, the update $G_j$ satisfies the `no-change' condition:
\eq{eq:msecant2} 
  G_j q = 0 \quad \forall q \perp v_j  .
  \en
  The usual no-change condition for secant methods is of the form
  $(G_j-G_{j-m}) q = 0 $ for $q \perp \Delta f_j$ which in our case
  would be 
  $(G_j-G_{j-m}) q = 0 $ for $q \perp \ v_j $.
  One can
  therefore consider that we are updating $G_{j-m} \equiv 0$. 
In this sense, we can prove the optimality of \methodNamenl(m).

  \begin{theorem}[Optimality of \methodNamenl]
 The matrix $G_j$ in \eqref{eq:Gj} is the best approximation to the inverse Jacobian $\inv{J(x_j)}$ of $F(x)$ at $x_j$ among  all the matrices $G$ whose range $\operatorname{Range}(G) = \operatorname{Span}\{ V_j \}$ and satisfies the multisecant equation Equation $GV_j = P_j$. That is,
\begin{align}
     \label{eq: Hessian-optimality}
     G_j = \underset{\{G \in \mathbb{R}^{d\times d}|\operatorname{Range}(G) = \operatorname{Span}\{ V_j \}, GV_j = P_j\}}{\operatorname{arg\min}} \norm{GJ(x_i) - I}.
 \end{align}
 \end{theorem}

\xhdr{3) Comparison with Anderson Acceleration}
Let us  set $\beta = 0$ in Anderson Acceleration.
Without loss of generality  and in an effort  to simplify notation we also assume that $i_0 = 1$ each time.
 According to
 (\ref{eq:AA}--\ref{eq:thetaj}), the $j$-th iterate becomes simply
 $x_{j+1} = x_j - \FF_j \theta_j $ where
 $\theta_j$ is a vector that minimizes $\| F_j - \FF_j \theta \|$.
 For \methodNamenl~, we have $x_{j+1} = x_j + P_j y_j $
 where $y_j $ minimizes $ \| F_j + V_j y \| $. So this  is identical with
 Equation \nref{eq:AA} when $\beta = 0$  in which  $P_j \equiv \XX_j$,
 and $\FF_j$ is replaced by $V_j$.

  The most important relation for both cases is the multi-secant relation.
  For Anderson, 
  with  $ G_{j-m} = 0 $,    the multi-secant matrix in \nref{eq:msec} becomes
  \eq{eq:AAb0}
    G_{j} = \XX_j  (\FF_j^T\FF_j)^{-1}\FF_j^T
  \en 
  which can be easily seen to minimizes $ \| G \|_F$ for matrices $G$ that satisfy
the multisecant condition $     G \FF_j = \XX_j $ 
  and the no-change condition
  $
    G_j^T (G_j  - G) = 0 . 
  $ 
    Therefore the two methods differ mainly in the way in which the sets
    $\FF_j / V_j$ ,  and $\XX_j / P_j $ are defined. Let us use the more general notation
    $V_j, P_j$ for the pair of subspaces. 

    In both cases, a vector $v_j$ is related to the corresponding $p_j$ by the fact that
    $ v_j \approx J(x_j) p_j $.
    In the case of \methodNamenl~ this relation is explicitly enforced by
    a Frechet differentiation  (Line 10)-- before we perform an orthogonalization - which
    combines this vector with others -- without changing the span of the new $P_j$ (and also
    $V_j$). 

    In the case of AA, we have  $v_j = \Delta F_{j-1} = F_j - F_{j-1} $ and the relation exploited
    is that
    \begin{equation}
    \label{eq:ApproxAA}
        \begin{aligned}
                f_j & \approx F_{j-1} + J(x_{j-1}) (x_j - x_{j-1}) \to
      \Delta f_{j-1} \\
     & \approx J(x_{j-1}) \Delta  x_{j-1}
        \end{aligned}
    \end{equation}
      
      However, the approximation   $v_j \approx J(x_j) p_j $ in \methodNamenl~ is \emph{more accurate}- because we use an additional
      function evaluation to explicitly obtain  a more accurate approximation (ideally exact
      value)  for  $J(x_j) p_j$. In contrast when $x_j$ and $x_{j-1}$ are not close, then
      \nref{eq:ApproxAA} can be a very rough approximation. This is a key difference between the two
      methods. 

\section{Experimental Results}
This section compares our proposed algorithms \methodName~ and \methodNamenl~ to existing methods in the literature
with the help of a few experiments.
We first compare the 
convergence for linear problems and then for a softmax classification problem in the case where the gradients are either deterministic or stochastic. More experiments and experimental details are available in the Appendix \ref{Sec:moreexp}. 

\subsection{Linear Problems}
\label{sec:linear}
We first compare the performance on linear equations $\mathbf{A}\mathbf{x}=\mathbf{b}$ with Conjugate Gradient \cite{Hestenes1952MethodsOC}, generalized minimal residual method (GMRES) \cite{Saad-Schultz-GMRES} and Anderson Acceleration under different settings. 

\begin{figure*}[ht]
\centering
\begin{subfigure}[b]{0.32\textwidth}
\centering
\includegraphics[width=\linewidth,]{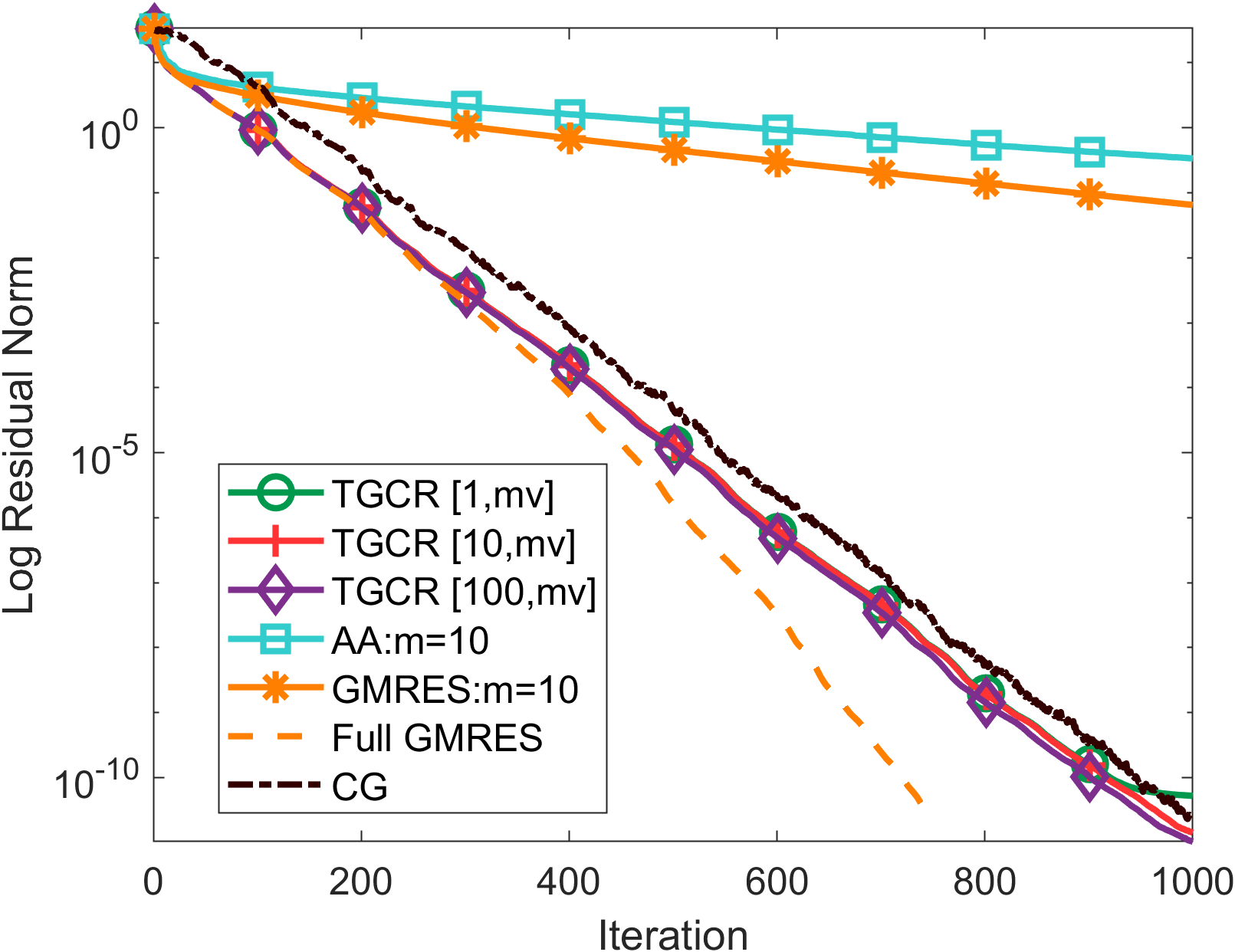}
\subcaption{ $\mathbf{A}$ is SPD} 
\label{fig:linsym}
\end{subfigure}
\begin{subfigure}[b]{0.32\textwidth}
\centering
\includegraphics[width=0.99\linewidth]{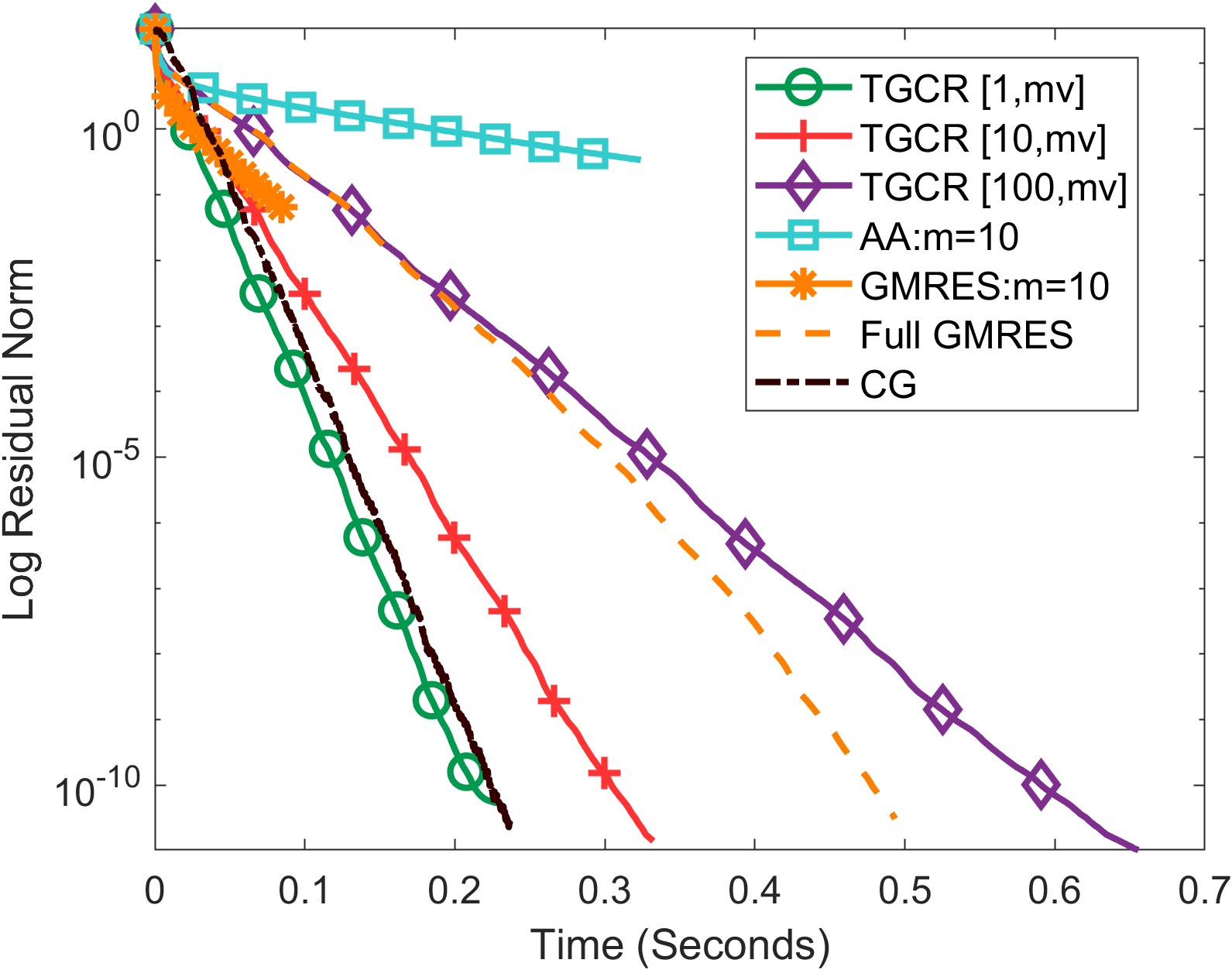}
\subcaption{Time comparison  (\ref{fig:linsym})} 
\label{fig:linsym2}
\end{subfigure}
\begin{subfigure}[b]{0.32\textwidth}
\centering
\includegraphics[width=0.99\linewidth]{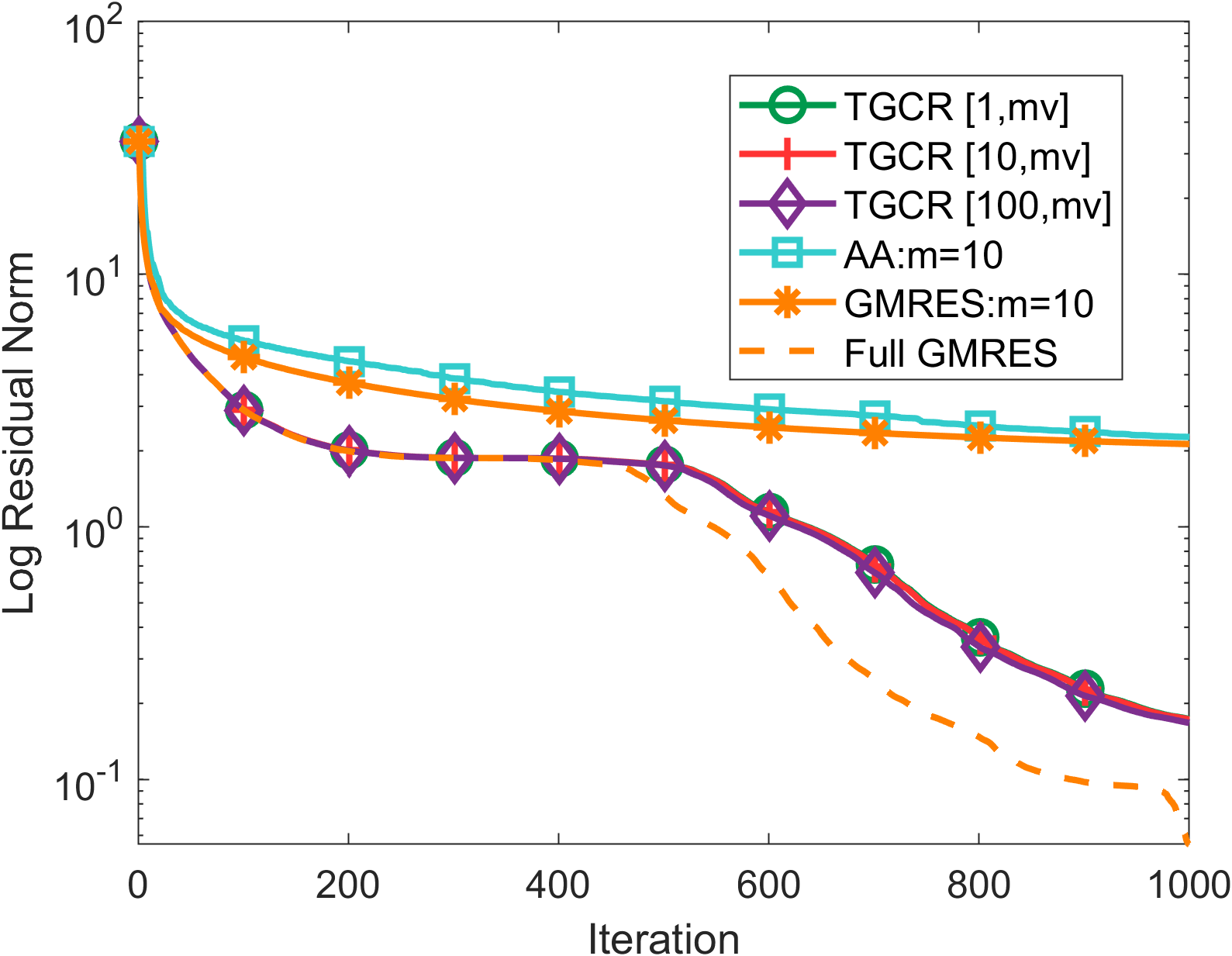}
\subcaption{$\mathbf{A}$ is symmetric indefinite} 
\label{fig:linin}
\end{subfigure}

\caption{\textbf{Linear Systems}  $\mathbf{A}\mathbf{x}=\mathbf{b}, \mathbf{A} \in \mathbb{R}^{1000\times1000}$: \textbf{\ref{fig:linsym}}: Comparison in terms of iteration, \methodName~ [$m$, mv] means  table size$=m$ and moving window (no restart). \textbf{\ref{fig:linsym2}}: Comparison in terms of time for problem in \ref{fig:linsym}. \textbf{\ref{fig:linin}}: Indefinite System. It is well known that CG fails for indefinite systems. The gap between full GMRES and \methodName~ is due to the numerical issue. It can be concluded that \methodName~ is ideal for solving linear systems because of its nice convergence property (compared to CG and AA) as well as the memory-efficient design (compared to GMRES). }
\label{fig:lineareq}
\end{figure*}

\paragraph{Linear Systems.} \ 
The advantages of \methodName~ for linear systems are two-folds. \textbf{1:)} Theorem \ref{thm:m1} shows that \methodName~($1$) is already optimal (equivalent to Conjugate Residual) when $\mathbf{A}$ is symmetric positive definite. A larger table size is unnecessary while AA and GMRES require  more past iterates to converge fast. It can be observed from Figure \ref{fig:linsym} and \ref{fig:linsym2} that \methodName~($1$) requires much less memory and computation overhead to converge compared to GMRES and AA. It also has the same convergence behavior and similar running time as CG. \textbf{2:)} It is easy to show that \methodName~ can converge for indefinite systems while CG fails. Figure \ref{fig:linin} verifies our point. This can be helpful when it is not known in advance if the system is indefinite. The numerical results shown in Figure \ref{fig:lineareq} demonstrate the power of \methodName~ as a variant of  Krylov subspace methods. Figure \ref{fig:lineareq} clearly verifies the correctness of Theorem \ref{thm:m1} that \methodName~(1) is identical with \methodName~$(m)$ in exact arithmetic, which leads to big savings both in terms of memory and in computational costs. We include more experimental results in the Appendix \ref{appendex:lin}. 

\begin{figure*}[]
        \begin{subfigure}[b]{0.25\textwidth}
                \centering
                \includegraphics[width=.85\linewidth]{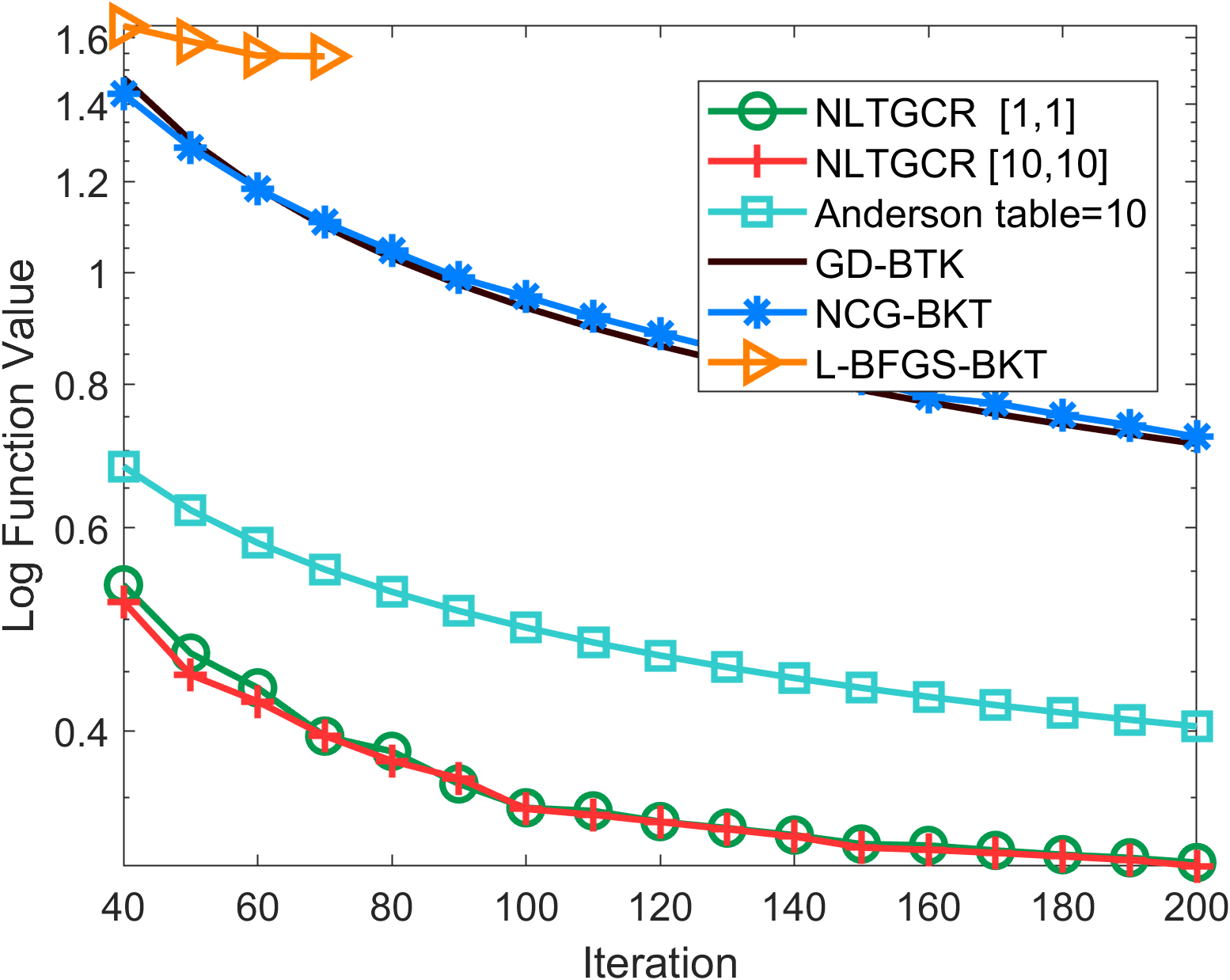}
                \caption{Function Value}
                \label{fig:fval}
        \end{subfigure}%
        \begin{subfigure}[b]{0.25\textwidth}
                \centering
                \includegraphics[width=.85\linewidth]{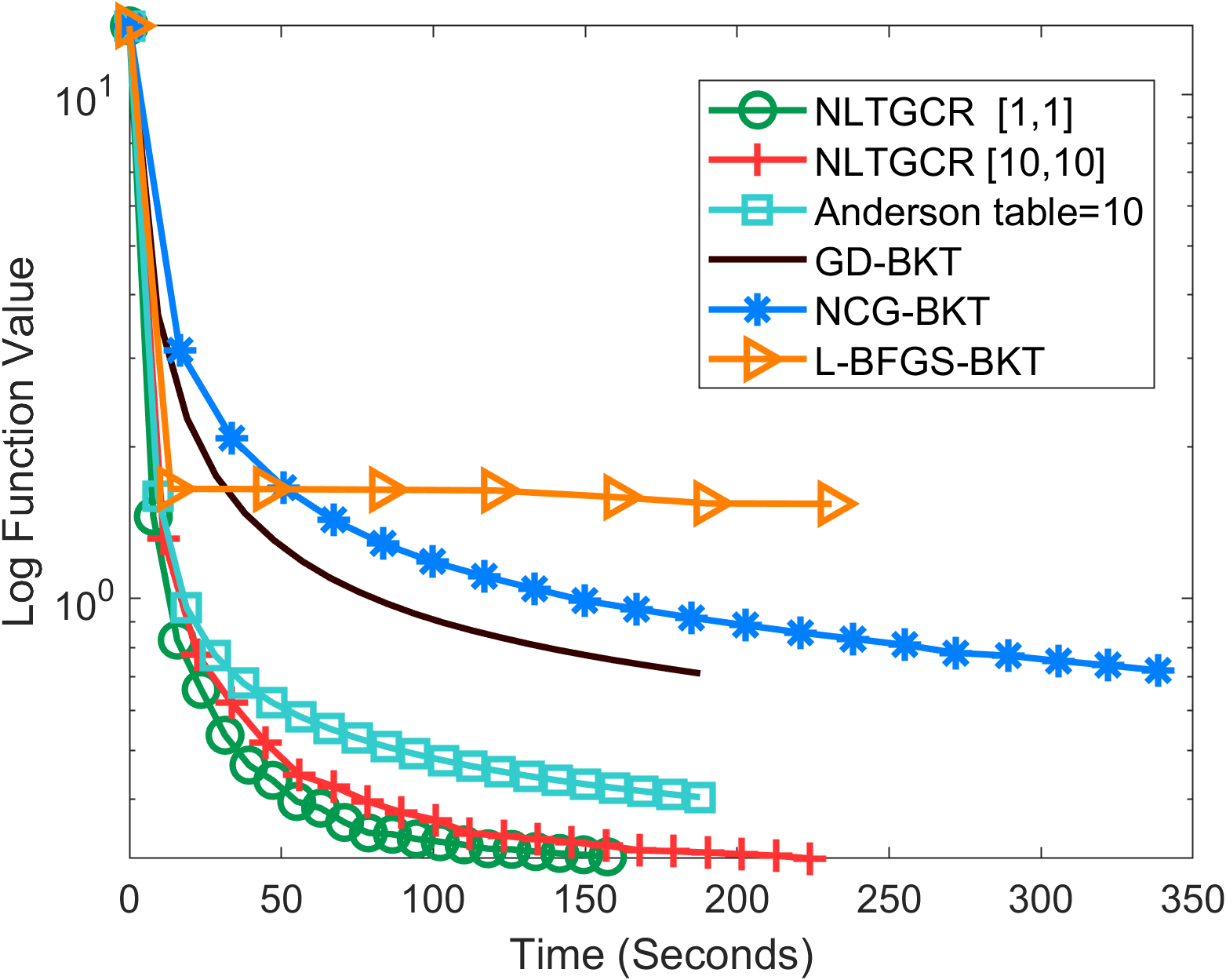}
                \caption{Time Comparison}
                \label{fig:soft_time}
        \end{subfigure}%
        \begin{subfigure}[b]{0.25\textwidth}
                \centering
                \includegraphics[width=.85\linewidth]{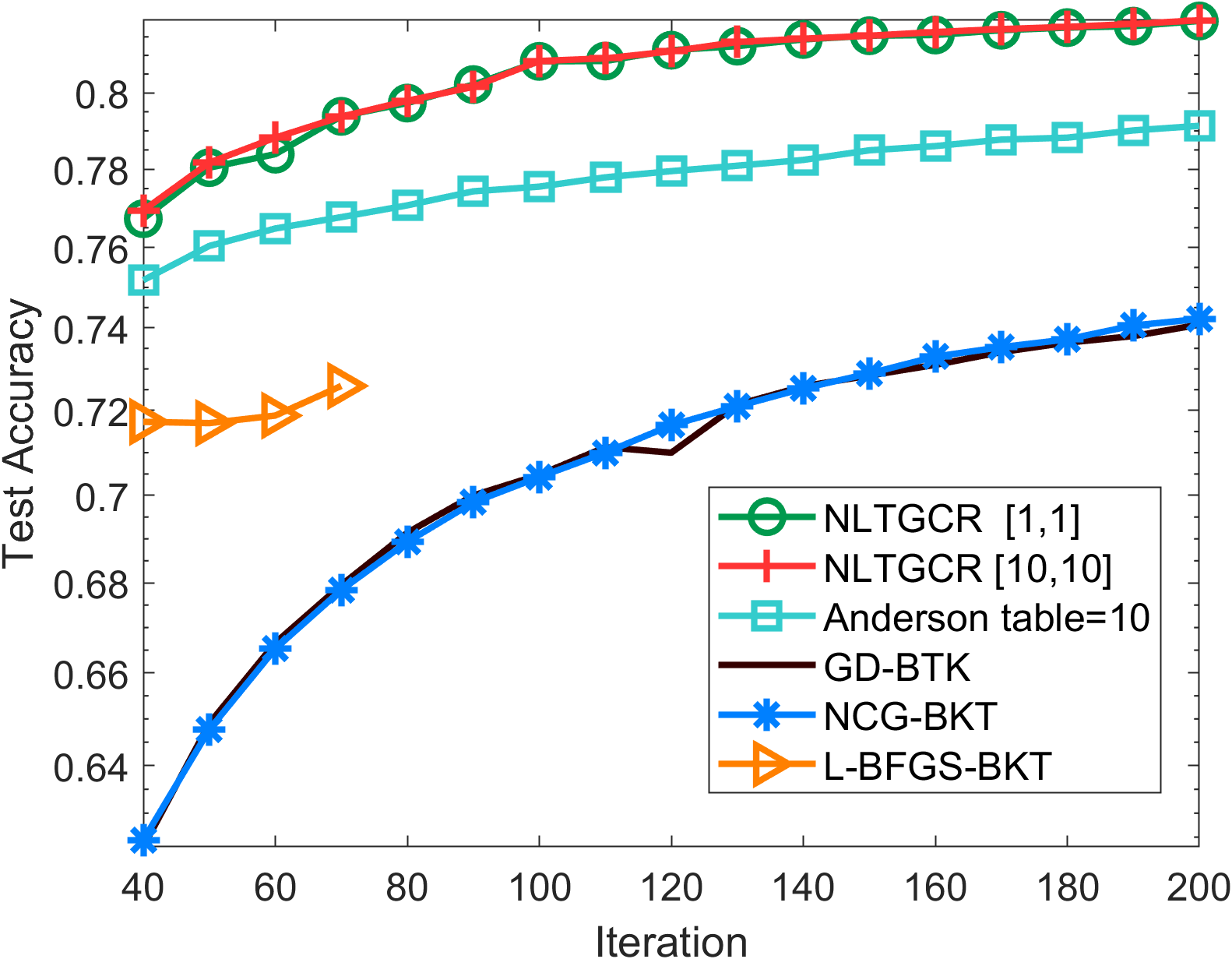}
                \caption{Test Accuracy}
                \label{fig:test_acc}
        \end{subfigure}%
        \begin{subfigure}[b]{0.25\textwidth}
                \centering
                \includegraphics[width=.85\linewidth]{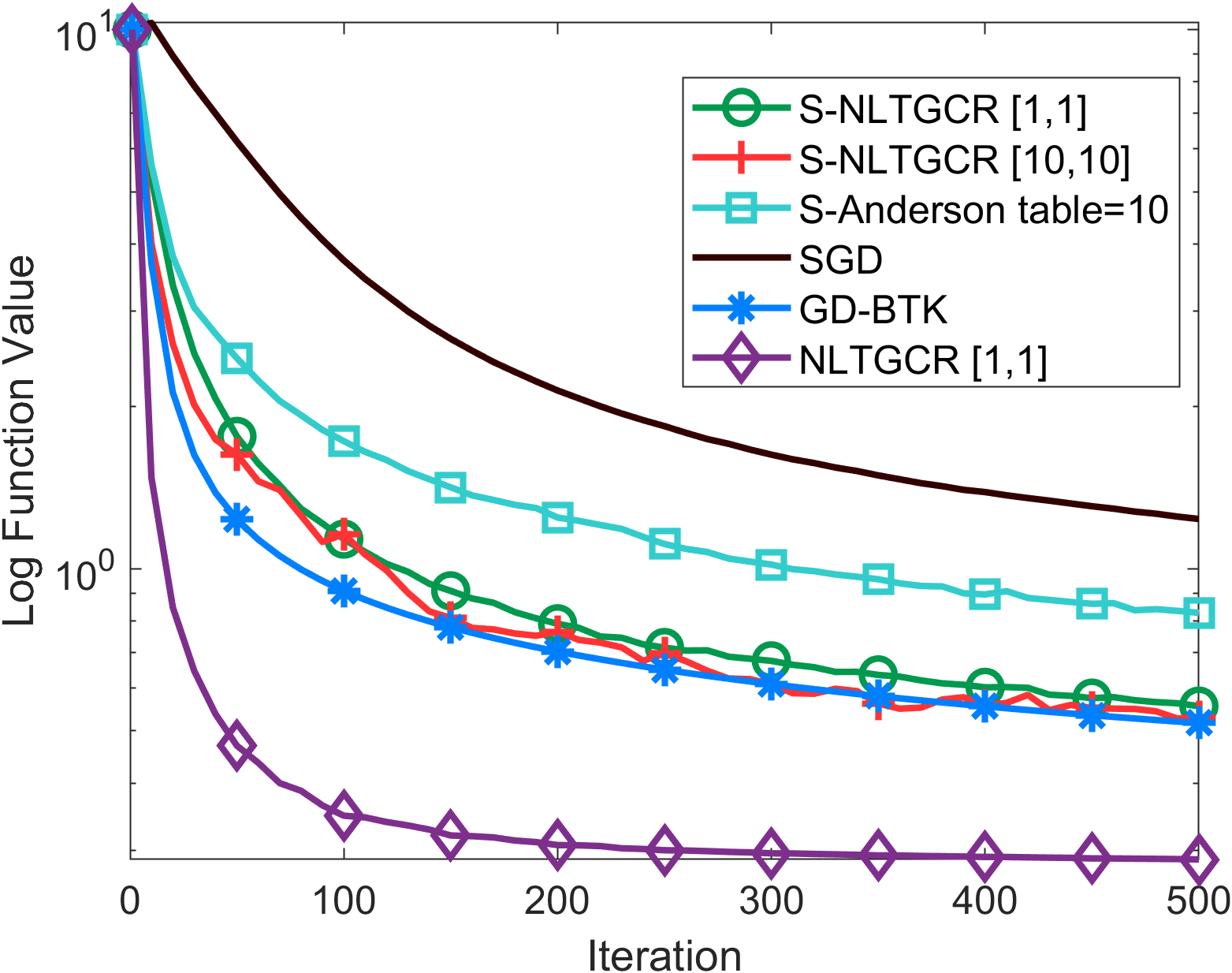}
                \caption{Stochastic Gradients}
                \label{fig:sto}
        \end{subfigure}
        \caption{
        \textbf{Softmax Classification (MNIST Dataset):}  \ref{fig:fval}: Function Value vs. Iterations; \ref{fig:soft_time}: Function Value vs. Time; \ref{fig:test_acc}: Test Accuracy vs. Iterations.  \ref{fig:sto}: Function Value vs. Iterations Stochastic gradients are calculated using a batch size of 500.}\label{fig:softmax}
\end{figure*}

\subsection{Nonlinear Problems: Softmax Classification}
\label{sec:softmax}
Next, we consider a softmax multi-class classification problem shown in \eqref{eqn:softmax} without regularization.
\begin{equation}
f =-\frac{1}{s} \sum_{i=1}^{s} \log \left(\frac{e^{w_{y_{j}}^{T} x^{(i)}}}{\sum_{j=1}^{k} e^{w_{j}^{T} x^{(i)}}}\right),
\label{eqn:softmax}
\end{equation}
where $s$ is the total number of sample, $k$ is the total number
of classes, $x^{(i)}$ is vector of all features of sample $i$, $w_{j}$ is the weights for the $j^{t h}$ class, and $y_{j}$ is the correct class for the $i^{th}$ sample. 
We compare \methodNamenl~ with Gradient Descent (GD), Nonlinear Conjugate Gradient (NCG) \cite{ncg}, L-BFGS \cite{Liu1989OnTL} and Anderson Acceleration using the MNIST dataset \cite{deng2012mnist} and report results in Figure \ref{fig:softmax}. Figure \ref{fig:fval} and \ref{fig:soft_time} plot the objective
value vs. iteration number and wall-clock time respectively. It can be seen that \methodNamenl~ converges significantly faster than baselines even without a line-search strategy. In addition, for this convex and symmetric problem, it is not surprising to observe that \methodNamenl$(1)$ exhibits a similar convergence rate with \methodNamenl$(m)$, which saves even more memory and computation time. Figure \ref{fig:test_acc} shows that \methodNamenl~ greatly outperforms baselines by achieving high test accuracy in the very early stage of training.  Figure \ref{fig:sto} shows the effectiveness of our method in the stochastic setting. `S-' stands for a stochastic version. We use a step size of 0.2 for SGD and a batch size ($B$) of 500 for all stochastic algorithms. It can be observed that \methodNamenl(1) with a small batch size is comparable with the full batch GD with line-search, which confirms 
that \methodName~ takes advantage of symmetry in a very effective way even in the case of stochastic gradients.

\subsection{Deep learning applications}
We then evaluate nlTCGR on several widely-used deep learning applications using different frameworks. We run experiments on image classification using CNN \cite{CNN} and ResNet \cite{He2015}, time series forecasting using LSTM \cite{HochSchm97}, and node classification using GCN \cite{kipf2017semi}. Due to space limitation, we provide full results in Appendix \ref{supsec:DL}. It shows that nlTGCR(1) outperforms baselines (SGD, Nesterov, and Adam) for the above DL experiments, highlighting its effectiveness in large-scale and stochastic non-convex optimization.



\section{Conclusion}
This paper describes an efficient nonlinear acceleration method that takes advantage of the  symmetry of the Hessian.
 We studied the convergence properties of the proposed method and established a few connections with existing methods. The numerical results suggest that nlTGCR  can be a competitive iterative algorithm from both theoretical and practical perspectives. We plan to
 conduct a more detailed theoretical and experimental investigation of the method for a nonconvex stochastic setting.

\xhdr{Social Impact}
This work does not present any foreseeable societal consequence.
\bibliography{strings,eig,saad,local,local1,msecant,accelbib}
\bibliographystyle{abbrv}
\clearpage
\begin{appendices}
\section{Additional Discussion}
\label{Sec:add}
\subsection{High-Level Clarification}
The method described in this paper mixes a number of
ideas coming from different horizons. Some of the high-level discussion provided below is
expanding further in later sections.

\xhdr{Linear case: TGCR}
Our initial idea was motivated by considering the linear case, in an attempt to  exploit
Conjugate-gradient like methods for solving a linear system $Ax=b$.
When $A$ is symmetric,  it is known
that it is possible to minimize the objective function $f(x) = \| b -Ax\|_2^2$
on the $k$-th Krylov subspace $\Span \{r_0, A r_0, \cdots, A^{k-1} r_0 \}$ by a
nice algorithm that uses a short-term recurrence. This
algorithm, called the Conjugate Residual algorithm, is quite similar to the
Conjugate Gradient but its residual vectors are conjugate (instead of being
orthogonal) and its search directions are $A^T A-$ conjugate (instead of being
$A$-conjugate). Its generalization to the nonsymmetric case, called the
Generalized Conjugate Residual method, is easy to obtain by enforcing these two
properties. Enforcing the $A^T A$ conjugacy of the $p_i$'s is the same as
enforcing the orthogonality of the vectors $A p_i$ and this is expensive when we
have many vectors. For this reason, practical versions of the algorithm are
\emph{truncated}, i.e., the orthogonalization is enforced against only a few
previous directions. The result is the TGCR(m) algorithm (Algorithm 1) -- which has been known since
the 1980s. It is
clear that \emph{we expect that when the matrix $A$ is nearly symmetric} TGCR(m)
will perform nearly as well as the full version GCR - because when $A$ is
symmetric, taking $m=1$ will yield full orthogonality of the $Ap_i$s (Theorem \ref{thm:appendTGCR}). 

\xhdr{Nonlinear case: Newton Krylov} 
Suppose now that we have to solve the nonlinear system $F(x) = 0$ (in optimization 
$F$ is just the gradient of the objective function).
At this point, we may ask the question: why not just use an inexact Newton
method whereby the Jacobian system is 
solved with the \emph{linear} GCR or TGCR method? This is where Anderson acceleration
provides an interesting insight on some weaknesses of  Newton-Krylov method.
A Newton Krylov method generates a Krylov subspace $\Span \{r_0, J r_0, \cdots, J^k r_0 \}$
at a current iterate -- say $K = x_0$ -- (so $J \equiv J(x_0) \equiv DF(x_0)$) 
and tries to minimize
$F(x_0 + \delta)$ where $\delta \ \in \ K$, by exploiting the linear model:
$F(x_0 + \delta) \approx F(x_0) + J \delta$. If we generate a basis $V = [v_1, \cdots v_k]$ of $K$
and express $\delta $ as $\delta = V y$ then we would need to minimize
$\| F(x_0) + J V y \| $ which is a small least-squares problem. One usually adds to this
a global convergence strategies, e.g., a linesearch or a trust-region technique
to produce the next iterate $x_1$. 
The problem with this approach is this: \emph{the approximate solution obtained after
  $k$ steps of a Krylov subspace approach is based on the Jacobian at the initial point $x_0$.}
The intermediate calculation is entirely
linear and based on $J(x_0)$. It is not exploited in any way to produce
intermediate (nonlinear) iterates which in turn could be used to produce more
accurate information on some local Jacobian.  In contrast, a method like
Anderson acceleration (or in fact any of the secant or multisecant methods) will
do just this, i.e., it will tend to use information on the nonlinear mapping
near the most recent approximation to produce the new iterate.   This
distinction is rather important although if the problem is nearly linear, then
it could make little difference.

\xhdr{Nonlinear case: Anderson and nlTGCR}
Anderson acceleration can be viewed as a form of
Quasi-Newton method whereby the approximate inverse Jacobian is updated at each step
by using the collection of the previous iterates $x_k, x_{k-1}, \cdots x_{k-m+1}$ and the
corresponding function values $F_k, F_{k-1}, \cdots F_{k-m+1}$. To be more accurate it uses
the differences $\Delta x_j = x_{j+1} - x_j$ and the corresponding $\Delta F_j$ defined in the same way. 
Similarly to  Newton-Krylov, it generates
an approximation of the form $x_k + P y$ where $P$ is a basis of the subspace spanned
by the $\Delta x_j$'s. Notice how the update now is on $x_k$  the latest point generated.
The previous iterates are used to essentially provide
information on the nonlinear mapping and its differential. This information is constantly updated
using the most recent iterate. Note that this is informally stated: Anderson does not
formally get an approximation to the Jacobian. It is based implicitly on exploiting the relation 
$F(x_{j+1})- F(x_j) \approx J (x_{j+1}- x_j)  $. Herein lies a problem that nlTGCR aims at
correcting: this relation is only vaguely verified.  For example,  
\emph{if we take $J$ to be $J(x_j) $, the Jacobian at  $x_j$,  the resulting linear  model is
bound to be extremely inaccurate at the beginning of the iteration.} 

{In nlTGCR, we try to improve on the Newton-Krylov approach, since our starting point is
TGCR which is generalized to nonlinear problems. We also take the viewpoint of
improving on Anderson Acceleration or multisecant methods by not relying on the
approximation
$F(x_{j+1})- F(x_j) \approx J (x_{j+1}-x_j)  $ mentioned above. 
This is achieved by adopting the projection viewpoint. Instead of
minimizing $\| F(x_0) + J  P y\|$ as in the inexact Newton mode,  we would like to now minimize
$\| F(x_k) + J  P y\|$ where $x_k$ is the most recent iterate.
This initial idea leads to a difficulty since there is not one $J$ but 
several ones at previous points and a single one of them will not be satisfactory.
Thus, we have a few  directions $p_i$ just like the differences $\Delta x_i$ in Anderson,
\emph{but now each $p_i$ will lead to a $J(x_i) p_i$ which - unlike in AA - is accurately
  computed and then saved.}
This feature is what we believe makes a difference in the performance of the algorithm -- although
this is something that would be rather difficult to prove theoretically.

\xhdr{The Quasi-Newton viewpoint.}
\label{sec:qnview}
It is also possible to view the algorithm from the alternative angle of
a Quasi-Newton approach  instead of  Inexact Newton.
In this viewpoint,  the inverse of the Jacobian
is approximated progressively. Because it is the inverse Jacobian that is
approximated, the method is akin to Broyden's second update method.

In our case, the approximate inverse Jacobian $G_j$ at step $j$ is equal to
\eq{eq:Gj}
G_{j} = P_j V_j^T .
\en 
If we apply this to the vector $v_j$ we get
$
  G_{j} v_j = P_j V_j^T v_j =  p_j =   \inv{J(x_j)} v_j .
$ 
So $G_j$ inverts $J(x_j)$ exactly when applied to $v_j$. It therefore satisfies 
the \emph{secant} equation (\cite[sec. 2.3]{FangSaad07}) 
\eq{eq:msecant1} 
  G_j v_j = p_j . 
  \en
This is the equivalent to the secant condition
$G_j \Delta f_j = \Delta x_j$ used in Broyden's second update method.
Broyden type-II methods replace
Newtons's iteration: $x_{j+1}  = x_j - \inv{Df(x_j)}  f_j$ with 
$     x_{j+1} = x_j - G_j f_j $  
  where $G_j $  approximates the inverse of the Jacobian $Df(x_j)$ at $x_j$
  by the  update formula $G_{j+1} = G_j + (\Delta x_j - G_j \Delta f_j) v_j^T $
  in which $v_j$ is defined in different ways see \cite{FangSaad07} for details.

In addition, the update $G_j$ satisfies the `no-change' condition:
\eq{eq:msecant2} 
  G_j q = 0 \quad \forall q \perp v_j  .
  \en
  The usual no-change condition for secant methods is of the form
  $(G_j-G_{j-m}) q = 0 $ for $q \perp \Delta f_j$ which in our case
  would be 
  $(G_j-G_{j-m}) q = 0 $ for $q \perp \ v_j $.
  One can
  therefore consider that we are updating $G_{j-m} \equiv 0$.

It is also possible to find a link between the method proposed herein
and the Anderson acceleration,
by unraveling  a relation with multi-secant methods.
Note that equation~\nref{eq:msecant1} is satisfied for (at most) $m$ previous
instances of $j$, i.e., at step $j$ we have ($i_0$ defined in the algorithm)
$ 
  G_j v_i = p_i $ {for} $i=i_0, \cdots, j .  
$ 
  In other words we can also write
\eq{eq:msecant1M}
 G_j V_j = P_j   .   
  \en  
  This is similar to the multi-secant condition
  $ G_j {\cal F}_j ={\cal X}_j $ of Equation
  \nref{eq:mscond} -- see also
  equation (13) of \cite{FangSaad07} where   $  {\cal F}_j $
  and   $  {\cal X}_j $ are defined in \nref{eqn:dfdx}.
  In addition, we clearly also have a multi secant version of the no-change
  condition \nref{eq:msecant2}  seen above, which becomes: 
  \eq{eq:msecant2M}
  G_j q = 0  \quad \forall \quad q  \perp \Span\{ V_j \} .
  \en
  This is similar to the no-change condition represented by
  eq. (15) of  \cite{FangSaad07}, which stipulates that 
  $(G_j - G_{j-m}) q = 0$ for all $q$ orthogonal to the span of the
  subspace $\Span \{ {\cal F}_j \}$ mentioned above, provided we define
 $ G_{j-m}= 0$. 
 
 \subsection{Complexity Analysis}
 Assume that the iteration number is $k$ and the model parameter size is $d$. The full memory AA stores all previous iterations, thus the additional memory is $2kd$. To reduce the memory
overhead, the limited-memory (Truncated) AA$(m)$ maintains the most recent $m$ iterations while discarding the older historical information. In comparison, TGCR and NLTGCR only requires \textbf{the most recent iterate} to achieve optimal performance, thus the additional memory is $2d$. The reduced number of past iterates also saves the orthogonalization costs from TGCR and NLTGCR compared to AA(m). In TGCR and NLTGCR, only one orthogonalization is needed to performed which costs $O(kd)$ while AA(m) requires $O(k^2d)$.  

For TGCR(m), $(2d-1)$ flops are performed in Line 5, $4d$ flops are performed in Lines 6-7 and $m(6d-1)$ flops are performed in the for loop and $2d$ flops are performed in Line 15. If TGCR(m) performs k iterations, the computational complexity is $((6m+8)d-1-m)k$. Thus, TGCR costs $O(mdk)$. For symmetric problems, $m=1$ is guaranteed to generate the same iterates as $m>1$ and TGCR costs $O(dk)$.

Then we analyze the complexity of nlTGCR(m). $m(2d-1)$ flops are performed in Line 6, $2md$ flops are performed in Line 7, two evaluations of $F$ are performed in Lines 8 and 11. The for loop costs $m(6d-1)$ flops and $2d$ flops are performed in Line 15. When $k$ iterations are performed, nlTGCR costs $O(mdk)$ plus the costs of $2k$ function evaluations of $F$. When $m=1$ is used in nonlinear problems, nlTGCR costs $O(dk)$ plus the costs of $2k$ function evaluations of $F$. 

\subsection{The Frechet derivative
}
In vector analysis, derivatives provide local linear approximations. Frechet differentiation can be used to calculate directional derivatives of gradients. We use Frechet Differentiation to compute  the directional derivative of a gradient mapping $f$ at $x$ in direction $h$, which is $v =  J(x_{j+1}) p$ in algorithm \ref{alg:nltgcr}. We define Frechet derivative as follows,

\begin{definition}
Let $(S,\|\cdot\|)$ and $(T,\|\cdot\|)$ be two normed spaces and let $X$ be an open set in $(S,\|\cdot\|)$.

A function $f: X \longrightarrow T$ is Fréchet differentiable at $x_{0}$, where $x_{0} \in X$, if there exists a linear operator $\left(D_{X} f\right)\left(x_{0}\right): X \longrightarrow T$ such that
$$
\lim _{h \rightarrow 0} \frac{\left\|f\left(x_{0}+h\right)-f\left(x_{0}\right)-\left(D_{x} f\right)\left(x_{0}\right)(h)\right\|}{\|h\|}=0
$$
The operator $\left(D_{x} f\right)\left(x_{0}\right): X \longrightarrow T$ is referred to the Fréchet derivative at $x_{0}$.
\end{definition}

\section{Proofs}
\subsection{Optimality for Linear Problem}
We can write the Generalized Conjugate residual
  formally as follows
    \begin{algorithm}[H]
    \centering
    \caption{GCR}\label{alg:gcr}
    \begin{algorithmic}[1]
  \State \textbf{Input}: Matrix $A$, RHS $b$, initial  $x_0$. \\
  Set $p_0=r_0 \equiv b-Ax_0$.
\For{$j=0,1,2,\cdots,$ Until convergence} 
\State  $\alpha_j = (r_j, A p_j) / (A p_j, A p_j)$
\State $x_{j+1} = x_j + \alpha_j p_j$
\State $r_{j+1} = r_j - \alpha_j A p_j$
\State $p_{j+1} = r_{j+1} - \sum_{i=1}^j \beta_{ij} p_i $ \quad where \quad
$\beta_{ij} :=  (A r_{j+1} , Ap_i) / (A p_i, Ap_i)$
\EndFor
\end{algorithmic}
\end{algorithm}

\begin{theorem}[Lemma~6.21 in \cite{Saad-book2}.]\label{thm:shortTGCR}
If $\{p_0,\dots,p_{n-1}\}$ is the basis of the Krylov space $\mathcal{K}_{n}(A, r_0)$ which are also $A^{T}A$ orthogonal . Then $$x_n = x_0 + \sum_{i = 0}^{n-1}\frac{\inner{r_0}{ Ap_i}}{\inner{Ap_{i}}{Ap_i}}p_i$$
minimizes the residual among all the iterates with form $x_0 + \mathcal{K}_{n}(A, r_0)$. Further more, we have
\begin{align*}
    x_n = x_{n-1} +\frac{\inner{r_{n-1}}{ Ap_{n-1}}}{\inner{Ap_{n-1}}{Ap_{n-1}}}p_{n-1}
\end{align*}
\end{theorem}
\begin{proof}
We can write $x_n = x_0 + \sum_{i = 0}^{n-1}\beta_ip_i$ and $r_n = r_0 -  \sum_{i = 0}^{n-1}\beta_iAp_i$. Since $x_n$ minimizes the residual, we know the following Petrov–Galerkin condition must hold
\begin{align*}
    (r_n, Ap_j) = 0, \quad j = 0, \dots, n-1
\end{align*}
The $A^{T}A$ orthogonality gives us
\begin{align*}
    \beta_i = \frac{\inner{r_0}{ Ap_i}}{\inner{Ap_{i}}{Ap_i}}.
\end{align*}
Similarly, we can write $x_n = x_{n-1} + \beta_{n-1}p_{n-1}$ and $r_n = r_{n-1} - \beta_{n-1}Ap_{n-1}$. Agagin, the optimality condition reads
\begin{align*}
    \inner{r_n}{p_{n-1}} = 0
\end{align*}
which gives us 
\begin{align*}
    \frac{\inner{r_{n-1}}{ Ap_{n-1}}}{\inner{Ap_{n-1}}{Ap_{n-1}}}
\end{align*}

\end{proof}

\begin{theorem}
When the coefficient matrix $A$ is symmetric, TGCR(m) generates exactly the same iterates as TGCR(1) for any $m>0$.
\label{thm:appendTGCR}
\end{theorem}
\begin{proof}
Lines 8 to 14 in Algorithm \ref{alg:tgcr} computes the new
direction $p_{j+1}$ -- by ortho-normalizing the vector $Ar_{j+1}$ against all
previous $Ap_i$'s.
In fact the loop of  lines 9--13, implements a modified Gram-Schmidt procedure, which in exact
arithmetic amounts simply to setting $p_{j+1}$ to
\eq{eq:equivAlg}
\beta_{j+1,j} p_{j+1} := r_{j+1} - \sum_{i=i_0}^j \beta_{ij} p_i \quad \mbox{where}\quad
\beta_{ij} = (Ar_{j+1}, A p_i) \ \mbox{for} \ i_0 \le i \le j. 
\en
In the above relation, $\beta_{j+1,j}$ is the scaling factor $\| v\|$ used to normalize
$p$ and $ v$ in Line~14. 
Then, $v_{j+1} \equiv A p_{j+1}  $ is computed accordingly as is reflected  in lines~12
and 14.
The update relation 
$A p_{j+1} = A r_{j+1} - \sum_{i=i_0,j} \beta_{ij}Ap_i$  (from Line 12) shows
that $A p_{j+1} \perp A p_i $ for $i=i_0, ...,j$.  In addition, it can
easily be shown that in this case ($m=\infty$) the residual vectors produced by the algorithm
are $A$-conjugate in that $(r_{j+1}, A r_i) = 0$ for $i\le j$.
Indeed, this requires a simple induction argument exploiting the  equality:
\[ 
  (r_{j+1}, A r_i) = (r_j -\alpha_j A p_j, A r_i) = (r_j , A r_i)  -\alpha_j (A p_j, A r_i) 
\]
and relation \nref{eq:equivAlg} which shows that $A r_{i} = \sum \beta_{k,i-1} A p_k $.

When $A$ is symmetric, exploiting the relation
$r_{i+1} = r_i - \alpha_i A p_i$, we can see that the scalar $\beta_{ij} $ in
Line~11 of Algorithm \ref{alg:tgcr} is
\[
  \beta_{ij} = (A r_{j+1}, A p_i) = \frac{1}{\alpha_i}
  (A r_{j+1}, r_i - r_{i+1}) = \frac{1}{\alpha_i} 
  (r_{j+1}, A r_i - A r_{i+1}) 
\]
which is equal to zero for $i < j$. Therefore we need to orthogonalize
$Ar_{j+1}$ against vector $Ap_j$ only in the loop of lines 9 to 13. This completes the proof.
\end{proof}

\newcommand{\x}{\mathbf{x}}
\newcommand{\bv}{\mathbf{v}}
\newcommand{\bu}{\mathbf{u}}
\newcommand{\y}{\mathbf{y}}
\newcommand{\A}{\mathbf{A}}
\newcommand{\F}{\mathbf{F}}
\newcommand{\G}{\mathbf{G}}
\newcommand{\0}{\mathbf{0}}
\newcommand{\w}{\mathbf{w}}
\newcommand{\br}{\mathbf{r}}
\newcommand{\bb}{\mathbf{b}}


{
\begin{theorem}[]
\label{thm:residual}
Let $\widehat{\mathbf{x}}_t$ be the approximate solution obtained at the t-th  iteration of \methodName~ being applied to solve $\A \x =\bb$, and denote the residual as $\br_t = \bb - \A\widehat{\mathbf{x}}_t$. Then, $\br_t$ is of the form
\begin{equation}
    \br_t=f_{t}(\A)\br_0,
\end{equation}
where
\begin{equation}
    \|\br_t\|_2 = \| f_{t}(\A)\br_0\|_2 = \min_{f_{t}\in \mathcal{P}_{t}}\| f_{t}(\A)\br_0\|_2,
\end{equation}
where $\mathcal{P}_{p}$ is the family of polynomials with degree p such that $f_p(0) = 1, \forall f_p \in\mathcal{P}_{p} $, which are usually called residual polynomials.
\label{resdiue-poly}
\end{theorem}
}

{
\begin{theorem}[Convergence of TGCR (Indefinite Case)]
    Suppose $\mathbf{A}$ is hermitian, invertible, and indefinite. Divide its eigenvalues into positive and negative sets $\Lambda_{+}$and $\Lambda_{-}$, and define
$$
\kappa_{+}=\frac{\max _{\lambda \in \Lambda_{+}}|\lambda|}{\min _{\lambda \in \Lambda_{+}}|\lambda|}, \quad \kappa_{-}=\frac{\max _{\lambda \in \Lambda-}|\lambda|}{\min _{\lambda \in \Lambda_{-}}|\lambda|}
$$
Then $\mathrm{x}_{m}$, the $m$ th solution estimate of TGCR, satisfies
$$
\frac{\left\|\mathbf{r}_{m}\right\|_{2}}{\|\mathbf{b}\|_{2}} \leq 2 \left(\frac{\sqrt{\kappa_{+} \kappa_{-}}-1}{\sqrt{\kappa_{+} \kappa_{-}}+1}\right)^{\lfloor m / 2\rfloor}
$$
where $\lfloor m / 2\rfloor$ means to round $m / 2$ down to the nearest integer.
\end{theorem}
}
{
\begin{proof}
When A is hermitian indefinite, an estimate on the min-max approximation
\begin{equation}
\frac{\left\|\mathbf{r}_{m}\right\|_2}{\left\|\mathbf{b}\right\|_2} \leq \min _{p \in \mathcal{P}_{m}} \max _{k}\left|p\left(\lambda_{k}\right)\right|
\end{equation}
that represents the worst-case TGCR convergence behavior, can be obtained by replacing the discrete set of the eigenvalues by the union of two intervals containing all of them and excluding the
origin, say $\Lambda_{+}$and $\Lambda_{-}$. Then the classical bound for the min-max value can be used to obtain an estimate for the convergence of the residual \cite{Saad-book1}
$$
\begin{aligned}
\min _{p \in \mathcal{P}_{m}} \max _{k}\left|p\left(\lambda_{k}\right)\right| & \leq \min _{p \in \mathcal{P}_{m}} \max _{z \in \Lambda_{+} \cup \Lambda_{-}}|p(z)| \\
& \leq 2\left(\frac{\sqrt{\kappa_{+} \kappa_{-}}-1}{\sqrt{\kappa_{+} \kappa_{-}}+1}\right)^{\lfloor m / 2\rfloor},
\end{aligned}
$$
where $[m / 2]$ denotes the integer part of $m / 2$.
\end{proof}
}

The optimality of TGCR(m) is proved in Theorem \ref{thm:tgcr}. 

\begin{theorem}
\label{thm:tgcr}
Let $r$ be the residual generated by the basic \methodName~(m), the following relations hold:
\begin{enumerate}

\item $\Span \{p_0, \cdots, p_{m-1} \} =
  \Span \{r_0, \cdots, A ^{m-1} r_0 \} \equiv K_m(r_0, A)$

\item If $m \ge j$,  the set of vectors $ \{A p_i \}_{i=1:j} $ is orthonormal.

\item More generally:  $ (A p_i, A p_j) = \delta_{ij}$ for
  $|i-j| \le m-1 $  

\item If $m \ge j$, then  $ \| b - A x_j \| = \min \{ \| b - A x \| \quad \vert \quad  \ x \ \in \
  x_0 + K_m(r_0,A) \}$

  \end{enumerate} 
\end{theorem}

\begin{proposition}
  Assume that  $J(x_j)$ is nonsingular and that $\mu_j  \equiv   v_j^T r_j   \ne 0$.
  Then $\delta_j = \mu_j p_j$ is a descent direction for the function
  $\phi(x) = \half \| F(x) \|^2 $ at $x_j$.
\end{proposition}
\begin{proof}
  It is known  \cite{Brown-Saad2} that the gradient of $\phi(x) $ at $x$ is
  $\nabla \phi(x) = J(x)^T F(x)$. In order for $p_j$ to be a descent direction at $x_j$ it
  is sufficient that   the inner product of $p_j$ and $\nabla \phi(x_j)$  is negative.
  Consider this inner product
  \eq{eq:inProd}
  (\nabla \phi (x_j) , \mu_j p_j) =  \mu_j (J(x_j)^T F(x_j), p_j) = \mu_j 
  (F(x_j), J(x_j) p_j) =  \mu_j (-r_j, v_j) = -\mu_j^2 <0 .
  \en
  which proves the result.
\end{proof}
\subsection{Optimality from Quasi-Newton Viewpoint}
\begin{theorem}[Optimality of nltgcr(m) from Quasi-Newton Viewpoint]
 The matrix $G_j$ is the best approximation to the inverse Jacobi $\inv{J(x_i)}$ of $F(x)$ at $x_i$ among  all the matrices $G$ whose range $\operatorname{Range}(G) = \operatorname{Span}\{ V_j \}$. That is,
\begin{align}
     \label{eq: Hessian-optimality}
     G_j = \underset{\{G \in \mathbb{R}^{d\times d}| GV_j = P_j\}}{\operatorname{arg\min}} \norm{GJ(x_i) - I}.
 \end{align}
 \end{theorem}
 \begin{proof}
 Assume $G$ is an arbitrary matrix satisfying the multisecant condition $   GV_j = P_j$. We have $(G_j - G)G_{j}^{T} = 0$ and $\operatorname{Range(G)} = V_j$. This can be derived as follows
 \begin{align*}
      0 = P_j(P_j^{T} - P_j^{T}) = P_jV_j^{T}(G^{T} - G_j^{T})  = G_j(G^{T} - G_j^{T}).
 \end{align*}
 We also have $(G_j - G)V_{j} = 0$.
 Then set $\Delta = G - G_j$, we have
 \begin{align*}
      \norm{GJ(x_i) - I} &=  \norm{(G_j + \Delta)J(x_i) - I} \\
      &= \norm{G_jJ(x_i) - I} + \norm{\Delta J(x_i)} + 2\operatorname{Trace}((G_jJ(x_i) - I)^{T}\Delta J(x_i))\\
      &\geq \norm{G_jJ(x_i) - I} + 2\operatorname{Trace}((G_jJ(x_i) - I)^{T}\Delta J(x_i)).
 \end{align*}
Then we prove $\operatorname{Trace}((G_jJ(x_i) - I)^{T}\Delta J(x_i)) = 0$. In order to prove this, we compute the trace explicitly. We will denote the natural basis in $\mathbb{R}^{d}$ by $\{e_l\}_{l=1}^{d}$ and $l-th$ column of $J(x_i)$ by $J_l$.
\begin{align*}
  \operatorname{Trace}((G_jJ(x_i) - I)^{T}\Delta J(x_i)) &= \sum_{l=1}^{d}e_{l}^{T}  ((G_jJ(x_i) - I)^{T}\Delta J(x_i))e_l \\
  & = \sum_{l=1}^{d} (J_l^{T}G_j^{T} - e_l^{T})\Delta J_l
\end{align*}
Recall we have $\operatorname{Range}(G) = \operatorname{Range}(G_j) = \operatorname{Span}\{V_j\}$ and $GV_j = G_jV_j = P_j$ , so 
$$
  \left\{
  \begin{array}{cc}
    \Delta J_l = 0   &  J_l \in V_j\\
     J_l^{T}G_j^{T} = 0,~ GJ_l = 0   &  J_l \in V_j^{\perp}
  \end{array}\right\}
 $$
 \end{proof}
\subsection{Convergence Analysis}
Firstly, we will show the global convergence of the Algorithm \ref{alg:nltgcr} from inexact Newton perspective. Usually, some global strategies like line search or residue check are required for inexact Newton method to converge for $\phi(x)$.
\begin{algorithm}
 \caption{Linesearch Algorithm}
 \label{alg:linesearch}
 \begin{algorithmic}[1]
  \State $\beta = \max \{ 1, \epsilon^{*}\frac{|\nabla \phi(x_n)^{T}p_n|}{\norm{p_n}^2}\}$.
  \State If $\phi(x_n + \beta p_n) \leq \phi(x_n) + \alpha\beta \nabla\phi(x_n) $, then set $\beta_n = \beta$ and exit. Else:
  \State Shrink $\beta$ to be $\beta\in [\theta_{\min}\beta,  \theta_{\max}\beta]$ where $0< \theta_{\min}\leq \theta_{\max} < 1$ .Go back to Step. 2.
 \end{algorithmic}
 \end{algorithm}
In \cite{Brown-Saad2}, authors showed with the general line search algorithm \ref{alg:linesearch}, inexact Newton-Krylov method can converge globally under some mild conditions.

 \begin{theorem}[Global Convergence from Algorithm \ref{alg:nltgcr} with linearized update and line search]
\label{thm:linearizedtgcr}
 Assume $\phi$ is continuously differentiable and $F(x)$ is L-lipschitz. Furthre more, the residual check is satisfied $\norm{J(x_n) P_ny_n + F(x_n)} \leq \eta \norm{\nabla f(x_n)}$ where $0\leq \eta_n\leq \eta < 1$. If $J(x_n)$ is nonsingular and its norm is bounded from above for all n, then $P_ny_n$ produced in line 7 of Algorithm \ref{alg:nltgcr} is a descent direction and the iterates $x_n$ produced by Algorithm \ref{alg:nltgcr} with linearized update and line search in Algorithm \ref{alg:linesearch} will converge to the minimizer:
 \begin{align*}
     \lim_{n\rightarrow \infty} \phi(x_n) = 0
 \end{align*}
 \end{theorem}
 \begin{proof}
Since Algorithm \ref{alg:nltgcr} with linearized update is equivalent to inexact Newton Krylov method with TGCR as the solver for the Jacobian system $J(x_n)p_n  = -F(x_n)$, the theorem is just a result of Theorem \ref{thm:Newton-Krylov}.
 \end{proof}

\begin{theorem}[\cite{Brown-Saad2}]
\label{thm:Newton-Krylov}
Assume $\phi$ is continuously differentiable and $F(x)$ is L-Lipschitz and let $p_n$ be such that $\norm{F(x_n) + J(x_n)p_n}_2 \leq \eta_n \norm{F(x_n)}_2$ for each $\eta_n\leq \eta <1$. Further more, let the next iterate be decided by Algorithm \ref{alg:linesearch}  and $J(x_n)$ is nonsingular and bounded from above for all n. Then
\begin{align*}
    \lim_{n\rightarrow \infty} \phi(x_n) = 0.
\end{align*}
\end{theorem}
The proof of this theorem depends on the following lemma in \cite{Brown-Saad2},
\begin{lemma}[Lemma 3.8 of \cite{Brown-Saad2}]
\label{lemma: backtracking property}
Assume $\phi$ is differentiable and $\nabla \phi$ is L-lipschitz. Let $0 < \alpha < 1$ and $p_n$ denote a descent direction. Then the iterates $x_{n+1}  = x_n + \beta p_n$ in Algorithm \ref{alg:linesearch} will generated in finite backtracking steps and $\beta_n$ satisfies 
\begin{align*}
    \beta_n \norm{p_n}_{2}\geq -\frac{\nabla \phi^{T}p_n}{\norm{p_n}}\min\Big(\epsilon^{*}, \frac{1-\alpha}{L}\theta_{\min}\Big).
\end{align*}
\end{lemma}


\begin{theorem}[Global convergence of \methodNamenl~ with  residual check] 
Assume $\phi$ is twice differentiable and $F(x)$ is L-lipschitz. If the residual check is satisfied $\norm{J(x_n) P_ny_n + F(x_n)} \leq \eta_n \norm{F(x_n)}$ where $0\leq\eta_n \leq \eta < 1$ and $J(x_n)$ is non-singular and the norm of its inverse is bounded from above for all n, then $P_ny_n$ produced in line 7 of Algorithm \ref{alg:nltgcr} is a descent direction and the iterates $x_n$ produced by Algorithm \ref{alg:linesearch} will converge to the minimizer $x^{*}$:
\begin{align*}
         \lim_{n\rightarrow \infty} \phi(x_n) = \phi(x^{*}) = 0.
\end{align*}

 \label{thm:inexact-Newton-global}
 \end{theorem}
 \begin{proof}
 Denote $P_ny_n$ by $p_n$, $J(x_n)p_n + F(x_n) = r_n$. Since $\nabla \phi(x_n) = J(x_n)^{T}F(x_n)$ and $\norm{r_n}\leq \eta_n \norm{F(x_n)}$, we have $\nabla \phi(x_n)^{T}p_n = F(x_n)^{T}r_n - F(x_n)^{T}F(x_n) \leq (\eta - 1) \nabla\norm{F}^2 = -2(1-\eta)\phi$ which implies $p_n$ is a descent direction. To see the second part of the theorem, we have
 \begin{align}
     \phi(x_n + \beta_n \alpha p_n) &\leq \phi(x_n) + \beta_n\alpha\nabla\phi(x_n)^{T}p_n\\
     &\leq \phi(x_n) - 2\beta_n\alpha(1-\eta)\phi(x_n) = [1 - 2\beta_n\alpha(1-\eta)]\phi(x_n). 
 \end{align}
 Denote $\min\Big(\epsilon^{*}, \frac{1-\alpha}{L}\theta_{\min}\Big)$ by $C$ then,
 \begin{align*}
     -\beta_n\norm{p_n} \leq C \frac{\nabla\phi(x_n)^{T}p_n}{\norm{p_n}_2}.
 \end{align*}
 Inserting it back to Inequality (27), we have
 \begin{equation}
 \label{ineq:key-ineq}
       \phi(x_{n+1}) \leq \Big(1 + 2\alpha(1-\eta)C \frac{\nabla\phi(x_n)^{T}p_n}{\norm{p_n}_2^2}\Big)\phi(x_n)
 \end{equation}
 Denote $2\alpha(1-\eta)C$ by $\lambda$ and $\frac{\nabla\phi(x_n)^{T}p_n}{\norm{p_n}_2^2}$ by $t_n$, then $\phi(x_{n+1})\leq (1+\lambda t_n)\phi(x_n)$. Since $\phi(x_n)$ is bounded from below and non-increasing by the inequality. It must converge to a finite limit $\phi^{*}$. If $\phi^{*} = 0$, we 're done. Otherwise, dividing the Inequality \ref{ineq:key-ineq} by $\phi(x_n)$ on both sides, we have
 \begin{equation}
     \frac{\phi(x_{n+1})}{\phi(x_n)} \leq (1+\lambda t_n) \rightarrow 1, \quad \text{as}~ n \rightarrow \infty.
 \end{equation}
 We also know $1+\lambda t_n \leq 1$. Therefore, $t_n\rightarrow 0$, as $n\rightarrow \infty$. In the above discussion, we showed $2(1-\eta)\phi(x_n)\leq |t_n|\norm{p_n}_2^{2}$ which implies $\norm{p_n} \rightarrow \infty$. 
 Recall $p_n = J(x_n)^{-1}(r_n - F(x_n))$, we must have $\norm{p_n}$ bounded. This contradicts with the fact $\norm{p_n} \rightarrow \infty$. Therefore, $\phi^{*} = 0$
 \end{proof}
 To proceed to the superlinear and quadratic convergence results, we need the following lemma from \cite{Eisenstat-Walker94}
 \begin{lemma}
 \label{Jacobi-nonsingular-implies-Gradient-zero}
 Assume F is continuously differentiable, $\{x_k\}$ is a sequence such that $F(x_k) \rightarrow 0$, and for each k,
 \begin{align}
     \norm{F(x_{k+1})} \leq \norm{F(x_k)} \quad \text{and} \quad\norm{F(x_k) + J(x_k)p_k} \leq \eta\norm{F(x_k)}
 \end{align}
 where $p_k = x_{k+1} - x_k$ and $\eta >0$ is independent of k. If $x_*$ is a limit point of $\{x_k\}$ such that $J(x_*)$ is nonsingular, then $F(x_*) = 0$ and $x_k \rightarrow x_*$. In this lemma, we don't require $\eta < 1$.
 \end{lemma}

  \begin{theorem}[Superlinear and quadratic convergence of \methodNamenl~]
  With the same setting as Theorem \ref{thm:inexact-Newton-global}. Assume both $\nabla \phi$ and $\nabla^{2}\phi$ are L-Lipschitz. Consider a sequence generated by Algorithm \ref{alg:nltgcr} such that residual check is satisfied $\norm{J(x_n) P_ny_n + F(x_n)} \leq \eta_n \norm{F(x_n)}$ where $0\leq\eta_n \leq \eta < 1$. Moreover, if the following conditions hold
  \begin{align}
  \label{back-track-line-search}
      \phi(x_n +  P_ny_n) &\leq \phi(x_n) + \alpha  \nabla\phi(x_n)^TP_ny_n\\
      \label{back-track-line-search-1}
      \phi(x_n +  P_ny_n) &\geq \phi(x_n) + \beta  \nabla\phi(x_n)^TP_ny_n
  \end{align}
  for $\alpha <\frac{1}{2}$ and $\beta >\frac{1}{2}$. If $x_n \rightarrow x_*$ with $J(x_*)$ nonsingular , then $F(x_*) = 0$. Moreover, there exists $N_s$ such that  $x_n \rightarrow x^{*}$ superlinearly for $n\geq N_s$ if $\eta_n\rightarrow 0$, as $n\rightarrow \infty$.  Furthermore, if $\eta_n = O(\norm{F(x_n)}^2)$, the convergence is quadratic.
 \end{theorem}
 \begin{proof}
 In the proof, we denote $Pny_n$ by $p_n$ for convenience and utilize the proof of Theorem 3.15 in \cite{Brown-Saad2}.
   According to assumptions, $x_n \rightarrow x_*$ with $J(x_*)$ nonsingular, then $J(x_n)$ is nonsingular for $n > n_J$ for some large enough $n_J$. Next, if $F(x_n) = 0$ for some $n \geq n_J$, then residual check condition will imply $p_n = 0$ which means $x_m = x_n$ for all $m\geq n$. Then the results hold automatically because the sequence converges in finite steps. Therefore, we can assume $J(x_n)$ is nonsingular and $F(x_n)$ is nonzero for all $n$.
   
   The residual check condition implies $p_n$ is a descent direction according to Lemma \ref{lemma: backtracking property}. That is, $\nabla \phi^{\top}p_n < 0$. Then we can show
   $$\lim_{n\rightarrow \infty} \frac{\nabla\phi_n^{\top}p_n}{\norm{p_n}} = 0.$$ To show this notice that according to \ref{back-track-line-search}, the following inequality holds
   \begin{align*}
       \phi_{n} - \phi_{n+1} \geq -\alpha\nabla \phi(x_n + p_n)^{\top}(x_{n+1} - x_n)  = \norm{p_n} \frac{\nabla\phi_n^{\top}p_n}{\norm{p_n}}
   \end{align*}
   Since $\phi_n$ is monotone decreasing, thus $\norm{p_n} \frac{\nabla\phi_n^{\top}p_n}{\norm{p_n}} \rightarrow 0$ as $n \rightarrow \infty$. To show $\lim_{n\rightarrow \infty}\frac{\nabla\phi_n^{\top}p_n}{\norm{p_n}} \rightarrow 0$. We also need to apply \ref{back-track-line-search-1}. Firstly, according to mean value theorem, there exists a $\lambda \in (0, 1)$ such that
   \begin{align}
       \phi_{n+1} - \phi_{n} = \nabla phi(x_n + \lambda p_n)^{\top}p_{n}.
   \end{align}
   According to \ref{back-track-line-search-1}, 
   \begin{align}
       \phi_{n+1} - \phi_{n}  = \nabla \phi(x_n + \lambda p_n)^{\top}p_{n} \geq \beta \nabla \phi_{n}^{\top}p_n.
   \end{align}
   This yields
   \begin{align*}
       [\nabla \phi(x_n + \lambda p_n) - \nabla \phi(x_n)]^{\top} p_n \geq (\beta - 1)\nabla\phi_{n}^{\top}p_n > 0.
   \end{align*}
  According to Cauchy-Schwartz inequality,
   \begin{align}
      (\beta - 1) \frac{\nabla\phi_n^{\top}p_n}{\norm{p_n}}\norm{p_n} \leq \norm{p_n} \norm{\nabla \phi(x_n + \lambda p_n) - \nabla \phi(x_n)} \leq L\lambda\norm{p_n}^{2}
   \end{align} 
   Therefore,
   \begin{align}
       \norm{p_n} \geq \frac{(\beta - 1)}{L\lambda} \frac{\nabla\phi_n^{\top}p_n}{\norm{p_n}} > 0.
   \end{align}
   which means we can draw the conclusion that $\norm{p_n}\frac{\nabla\phi_n^{\top}p_n}{\norm{p_n}} \rightarrow 0$ implies $\frac{\nabla\phi_n^{\top}p_n}{\norm{p_n}} \rightarrow 0$.
   If $x_n \rightarrow x_*$ with $J(x_*)$ nonsingular, then by Lemma \ref{Jacobi-nonsingular-implies-Gradient-zero}, we know $F(x_*) = 0$. According to the definition,
   \begin{align}
       p_n = -J_{n}^{-1}F_{n} + J_{n}^{-1}( F_n + J_n p_n).
   \end{align}
   This implies
   \begin{align}
   \label{pn-upper-bound}
       \norm{p_n} \leq \norm{J_n^{-1}}\norm{F_n} + \norm{J_n^{-1}}\norm{F_n + J_np_n} \leq (1+\eta) \norm{J_n^{-1}}\norm{F_n}.
   \end{align}
   We know $\norm{F_n} \rightarrow 0$ as $x_n \rightarrow x_*$. The above inequality implies that $\norm{p_n} \rightarrow 0$ as $x_n \rightarrow x_*$ since $J_n$ is nonsingular. Denote the residual $F_n + J_np_n$ by $r_n$. Then $\norm{r_n} \leq \eta \norm{F_n}$ and $p_n = J_n^{-1}(r_n - F_n)$. Therefore,
   \begin{align}
       \nabla \phi_n^{\top} = (J_n^{\top}F_n)^{\top} J_n^{-1}(r_n - F_n) = F^{\top}r - F^{\top}F.
   \end{align}
   This implies
   \begin{align}
       \frac{|\nabla\phi_n^{\top}p_n|}{\norm{p_n}} = \frac{|F^{\top}r - F^{\top}F|}{\norm{J_n^{-1}(r_n - F_n)}} \geq \frac{|F^{\top}F| - |F^{\top}r|}{\norm{J_n^{-1}(r_n - F_n)}}.
   \end{align}
   Since $\norm{r_n} \leq \norm{F_n}$ implies $|F^{\top}r| \leq \eta \norm{F_n}^2$, we have
   \begin{align}
       |F^{\top}F| - |F^{\top}r| \geq (1 - \eta) \norm{F_n}^{2}.
   \end{align}
   Moreover,
   \begin{align}
       \norm{J_n^{-1}(r_n - F_n)} \leq \norm{J_n^{-1}}\norm{F_n} + \norm{J_n^{-1}r}_{2} \leq (1+\eta)\norm{J_n^{-1}}\norm{F_n}.
   \end{align}
   Finally, we have
   \begin{align}
        \frac{|\nabla\phi_n^{\top}p_n|}{\norm{p_n}} \geq 
    \frac{(1 - \eta) \norm{F_n}^{2}}{(1+\eta)\norm{J_n^{-1}}\norm{F_n}} = \frac{(1 - \eta) \norm{F_n}}{(1+\eta)\norm{J_n^{-1}}}.
   \end{align}
   Using $\norm{\nabla \phi_n} = \norm{J_n^{\top}F_n} \leq \norm{J_n}\norm{F_n}$, we have
   \begin{align}
       \frac{|\nabla \phi_n^{\top}p_n|}{\norm{\nabla \phi_n}\norm{p_n}} \geq \frac{(1 - \eta)}{(1 + \eta)M_n},
   \end{align}
   where $M_n = \operatorname{cond}_2(J_n)$.
   Therefore, we have
   \begin{align}
       \frac{-\nabla \phi_n^{\top}p_n}{\norm{p_n}} \geq \frac{(1 - \eta)}{(1 + \eta)M_n} \norm{\nabla \phi_n} \geq \frac{(1 - \eta)}{(1 + \eta)M_n} \norm{J_n}^{-1}\norm{F_n},
   \end{align}
   This yields,
   \begin{align}
       \norm{F_n} \leq \frac{(1 + \eta)M_n}{(1 - \eta)} \norm{J_n} \frac{-\nabla \phi_n^{\top}p_n}{\norm{p_n}}.
   \end{align}
   Hence,
   \begin{align}
   \label{Fnpn-bound}
       \norm{F_n}\norm{p_n} \leq \frac{(1 + \eta)M_n}{(1 - \eta)} \norm{J_n} (-\nabla \phi_n^{\top}p_n) = -a_n \nabla \phi_n^{\top}p_n),
   \end{align}
  where $a_n = \frac{(1 + \eta)M_n}{(1 - \eta)} \norm{J_n}$ .
  Combining \ref{pn-upper-bound}, we have
  \begin{align}
  \label{pn_bound}
      \norm{p_n}^2 \leq \frac{(1 + \eta)^2M_n^2}{(1 - \eta)^2} (-\nabla \phi_n^{\top}p_n)  = -b_n\nabla \phi_n^{\top}p_n),
  \end{align}
  where $b_n = \frac{(1 + \eta)^2M_n^2}{(1 - \eta)^2}$. Next we show the convergence of the algorithm with the aid of the second order Taylor expansion.
  Notice 
  \begin{align}
  \nabla \phi(x) = J^{\top}F(x) = [\nabla F_{1}(x) \dots \nabla F_{n}(x)]
  \begin{bmatrix}
  F_1(x)\\
  \vdots\\
  F_n(x)
  \end{bmatrix}
  \end{align}
  The Hessian can be computed as follows
  \begin{align}
      \nabla^{2}\phi(x) = J^{T}J +  \sum_{i=1}^{n}\nabla^{2}F_{i}(x)F_{i}(x) = J^{T}J + G(x),
  \end{align}
  where $\norm{G(x)} = \norm{\sum_{i=1}^{n}\nabla^{2}F_{i}(x)F_{i}(x)}\rightarrow 0 $ as $x_n\rightarrow x_*$ since $F(x_*) = 0$.
  using second order Taylor expansion, we have
  \begin{align}
      \phi_{n+1} - \phi_{n} - \frac{1}{2}\nabla\phi_n^{\top} p_n = \frac{1}{2}(\nabla\phi_n + \nabla^{2}\phi(\Bar{x})p_n)^{\top}p_n.
  \end{align}
  where $\Bar{x} = \gamma x_n + (1 - \gamma)x_{n+1}$ for some $\gamma \in (0, 1)$.
  Then we can have
  \begin{equation}
  \begin{aligned}
      | \phi_{n+1} - \phi_{n} - \frac{1}{2}\nabla\phi_n^{\top} p_n| & = \frac{1}{2}|(\nabla\phi_n + \nabla^{2}\phi(\Bar{x})p_n)^{\top}p_n|\\
      &  = \frac{1}{2}|(\nabla\phi_n + \nabla^{2}\phi_n p_n)^{\top}p_n + p_n^{\top}(\nabla^{2}\phi(\Bar{x}) - \nabla^{2}\phi_n )p_n|\\
      & \leq \frac{1}{2} (\norm{J_n^{\top}(F_n + J_n p_n)}\norm{p_n} + \norm{G_n}\norm{p_n}^2 + L\norm{p_n}\norm{p_n}^{2}) \\
      & \leq (\eta\norm{J_n}\norm{F_n}\norm{p_n} + (\norm{G_n} + L\norm{p_n})\norm{p_n}^2)\\
      &\leq -\frac{1}{2}(a_n\eta_n\norm{J_n} + b_n (\norm{G_n} + L\norm{p_n}))\nabla \phi_n^{\top}p_n\\
      & = -\frac{1}{2}\epsilon_n\phi_n^{\top}p_n,
  \end{aligned}
  \end{equation}
 where $\epsilon_n = a_n\eta_n\norm{J_n} + b_n (\norm{G_n} + L\norm{p_n})$.
 Therefore,
 \begin{align}
    \frac{1}{2}(1+\epsilon_n)\nabla\phi_n^{\top} p_n  \leq \phi_{n+1} - \phi_{n} \leq \frac{1}{2}(1 -\epsilon_n)\nabla\phi_n^{\top} p_n
 \end{align}
 Notice that $\norm{J_n}$, $a_n$ and $b_n$ are all bounded from above and $\eta_n$, $\norm{S_n}$ and $\norm{p_n}$ all converges to $0$ as $x_n \rightarrow x_*$. Therefore $\epsilon_n\rightarrow 0$ as $x_n \rightarrow x_*$. And choose lager enough $N$ such that for all $n\geq N$ the following holds
 \begin{align}
     \epsilon_n \leq \min\{1 - 2\alpha, 2\beta - 1\}.
 \end{align}
  Then for all $n\geq N$, the Goldsetin-Armijo condition is satisfied,
  \begin{align}
     \beta \phi_n^{\top} p_n \leq  \phi_{n+1} - \phi_n \leq \alpha \phi_n^{\top} p_n.
  \end{align}
  We then finish the proof following Theorem 3.3 in \cite{Dembo-al}.
  It's easy to see
  \begin{align}
      J(x_*)(x_{k+1} - x_*) = [I + J_*(J_k^{-1} - J_{*}^{-1})](r_k + [J_k - J_*](x_k - x_*) - [F_k - F_* - J_*(x_k - x_*)])
  \end{align}
  Taking norm yields
  \begin{equation}
      \begin{aligned}
      \norm{x_{k+1} - x_*} &\leq [\norm{J_*^{-1}} + \norm{J_*}\norm{J_k^{-1} - J_{*}^{-1}}][\norm{r_k} + \norm{J_k - J_*}\norm{x_k - x_*} +\\
      & \quad \norm{F_k - F_* - J_*(x_k - x_*)}] \\
      &= [\norm{J_*^{-1}} + \norm{J_*}\norm{J_k^{-1} - J_{*}^{-1}}][\norm{r_k} + \norm{J_k - J_*}\norm{x_k - x_*} \\
      & \quad + \norm{F_k - F_* - J_*(x_k - x_*)}]\\
      & = [\norm{J_*^{-1}} + o(1)][o(F_k) + o(1)\norm{x_k - x_*} + o(\norm{x_k - x_*})] 
  \end{aligned}
  \end{equation}
  Therefore, 
  \begin{align}
       \norm{x_{k+1} - x_*} = o(F_k) + o(1)\norm{x_k - x_*} + o(\norm{x_k - x_*}), \quad \text{~} k\rightarrow \infty.
  \end{align}
  where we used the fact that for sufficient small $\norm{y-x_*}$
  \begin{align}
      \frac{1}{\alpha}\norm{y-x_*} \leq \norm{F(y)} \leq \alpha \norm{y - x_*}.
  \end{align}
  which is Lemma 3.1 in \cite{Dembo-al}. Similarly, to show quadratic convergence, juts notice 
  \begin{align}
      \norm{F(y) - F(x_*) - F(x_*)(y - x_*)} \leq L^{'}\norm{y- x_*}^{2} 
  \end{align}
  for some constant $L^{'}$ and sufficient small $\norm{y-x_*}$. For more details, check Lemma 3.2 in \cite{Dembo-al}.

 \end{proof}
\subsection{Stochastic nlTGCR}
Denote the noisy gradient by $F(x;\xi_{\mathcal{G}})$ and the noisy evaluation of Hessian along a vector $p$ by $J(x;\xi_{\mathcal{H}})p$. The subsample exact Newton algorithm is defined in Algorithm \ref{alg: subsample-Newton}. At $k$-th iteration, we uniformly subsample $\SG_k, \SH_k$ from full sample set to estimate the noisy gradient and Hessian, so both of them are unbiased. 
\begin{algorithm}
\caption{subsmaple Exact Newton}
\label{alg: subsample-Newton}
 \begin{algorithmic}[1]
 \For{$i = 1,\dots,k$}
 \State Estimate $F(x_i;\SG_i)$ and $J(x_i;\SH_i)$
 \State $x_{i+1} \leftarrow x_i - s_iJ^{-1}(x_i;\SH_i)F(x_i;\SG_i)$
 \EndFor
 \end{algorithmic}
\end{algorithm}
Before we start the theoretical analysis, we need to make some assumptions which are usual in stochastic setting. 
\paragraph{Assumptions for stochastic setting}
\begin{enumerate}[label=\subscript{E}{{\arabic*}}]
    \item The eigenvalues of Hessian matrix for any sample $|\SH| = \beta$ is bounded form below and above in Loewner order
    \begin{align}
        \mu_{\beta}I \preceq J(x, \SH) \preceq L_{\beta}I.
    \end{align}
    Further more, we require there is uniform lower and upper bound for all subsmaples. That is, there exists $\hat{\mu}$ and $\hat{L}$ such that
    \begin{align}
        0 \leq \hat{\mu} \leq \mu_{\beta} \quad \text{and} \quad L_{\beta} \leq \hat{L}< \infty, \quad \forall \beta \in \mathbb{N}.
    \end{align}
    And the full Hessian is bounded below and above
    \begin{align}
        \mu I \preceq J(x) \preceq LI,\quad \forall x.
    \end{align}
    \item The variance of subsampled gradients is uniformly bounded by a constant $C$.
    \begin{align}
        \operatorname{tr}(Cov(F(x))) \leq C^2, \quad \forall x
    \end{align}
    \item Hessian is M-Lipschitz, that is
    \begin{align}
        \norm{J(x) - J(y)} \leq M\norm{x-y}, \quad \forall x, y
    \end{align}
    \item The variance of subsampled Hessianis bounded by a constant $\sigma$.
    \begin{align}
        \norm{\Expectation{\SH}{(J(x;\SH) - J(x))}} \leq \sigma, \quad \forall x
    \end{align}
    \item There exists a constant $\gamma$ such that
    \begin{align*}
        \mathbb{E}[\norm{x_n - x^{*}}^2] \leq \gamma(\mathbb{E}[\norm{x_n - x^{*}}])^2.
    \end{align*}
\end{enumerate}

Firstly, we recall the few results on subsample Newton method from
\cite{bollapragada2019exact}.
\begin{theorem}[Theorem 2.2 in \cite{bollapragada2019exact}]
Assume $x_n$ is generated by Algorithm \ref{alg: subsample-Newton} with $|\SG_i| = \eta^{i}$ for some $\eta > 1$, $|\SH| = \beta\geq 1$ and $s_i = s = \frac{\mu_{\beta}}{L}$ and Assumptions E1-E2 hold, then
\begin{align}
  \Expectation{k}{\phi(x_k) - \phi(x^{*})} \leq \alpha \tau^{k},  
\end{align}
where
\begin{align*}
    \alpha = \max\Big\{\phi(x_0) - \phi(x^{*}), \frac{C^2L_{\beta}}{\mu\mu_{\beta}}\Big\}\quad \text{and} \quad \tau = \max\Big\{1 - \frac{\mu\mu_{\beta}}{2LL_{\beta}}, \frac{1}{\eta}\Big\}.
\end{align*}
\label{thm:SNewton-global}
\end{theorem}
\begin{theorem}[Lemma 2.3 form \cite{bollapragada2019exact}]
Assume $x_n$ is generated by Algorithm \ref{alg: subsample-Newton} with $s_i \equiv 1$ and Assumptions E1-E3 hold. Then
\begin{align*}
    \Expectation{k}{\norm{x_{n+1} - x^{*}}} \leq \frac{1}{\mu_{|\SH_n|}}\Big[\frac{M}{2}\norm{x_{n} - x^{*}}^{2} + \Expectation{k}{\norm{(J(x_n;\xi_{\SH_n}) - J(x_n))(x_{n} - x^{*})}} + \frac{C}{\sqrt{|\SG_n|}}\Big]
\end{align*}
\label{thm:subsample-Newton}
\end{theorem}
 \begin{lemma}[Lemma 2.4 from \cite{bollapragada2019exact}]
 Assume the assumption E1 and E4 hold. Then
 \begin{align}
     \Expectation{k}{\norm{(J(x_n;\xi_{\SH_n}) - J(x_n))(x_{n} - x^{*})}} \leq \frac{\sigma}{\sqrt{\SH_n}}\norm{x_k - x^{*}}.
 \end{align}
 \label{lemma: Hessian-variance}
 \end{lemma}
 \begin{theorem}[{Convergence of stochastic version of \methodNamenl~}] 
Assume  $|\SH_n| = \beta \geq \frac{16\sigma^2}{\mu},~ \forall n$, residue check is satisfied for $\eta_n\leq \eta \leq \frac{1}{4L}$ and assumptions E1-E5 hold.
 The iterates generated by the stochastic version Algoritrhm \ref{alg:nltgcr} converges to $x^{*}$ if $\norm{x_k - x^{*}}\leq \frac{\mu}{2M\gamma}$.
 \begin{align}
 \label{ineq:stochastic-bound}
     \mathbb{E}\norm{x_{n+1} - x^{*}} \leq \frac{3}{4}\mathbb{E}\norm{x_n - x^{*}}
 \end{align}
 \end{theorem}
 \begin{proof}
 \begin{align*}
      \Expectation{n}{\norm{x_{n+1} - x^{*}}} = \Expectation{n}{\norm{x_{n} - x^{*} - J(x_n)^{-1}F(x_n)}} + \Expectation{n}{\norm{J(x_n)^{-1}F(x_n) + P_nV_n^{T}y_n}}
 \end{align*}
 The first term can be bounded using the Theorem \ref{thm:subsample-Newton} and Lemma \ref{lemma: Hessian-variance}, 
 \begin{align*}
    \Expectation{n}{\norm{x_{n} - x^{*} - J(x_n)^{-1}F(x_n)}} &\leq \frac{1}{\mu_{|\SH_k|}}\Big[\frac{M}{2}\norm{x_{n} - x^{*}}^2 + \\
    &\Expectation{k}{\norm{(J(x_n;\xi_{\SH_n}) - J(x_n))(x_{n} - x^{*})}} + \frac{C}{\sqrt{|\SG_n|}}\Big]\\
    &\leq \frac{1}{\mu}[\frac{M}{2}\norm{x_{n} - x^{*}}^2  +  \frac{\sigma}{\sqrt{\SH_n}}\norm{x_n - x^{*}} ]\\
    & = \frac{M}{2\mu}\norm{x_{n} - x^{*}}^2  +  \frac{\sigma}{\mu\sqrt{\beta}}\norm{x_n - x^{*}}]
 \end{align*}
 We can bound the second term through the line search, recall at each iteration we have
 \begin{align*}
     \Expectation{n}{\norm{J(x_n)^{-1}F(x_n) + P_nV_n^{T}y_n}} &=  \Expectation{n}{\norm{F(x_n) + J(x_n)P_nV_n^{T}y_n}}\\
     & \leq \eta_n \Expectation{k}{\norm{F(x_n)}} \leq \eta_{n} L\norm{x_n - x^{*}}.
 \end{align*}
 The last inequality comes from the assumption that eigenvalues of $J(x)$ is uniformly upper bounded by $L$.
 Finally, combining the above inequalities gives us
 \begin{align*}
     \Expectation{n}{\norm{x_{n+1} - x^{*}}} \leq \frac{M}{2\mu}\norm{x_{n} - x^{*}}^2  +  \frac{\sigma}{\mu\sqrt{\beta}}\norm{x_n - x^{*}} +  \eta_{n} L\norm{x_n - x^{*}}
 \end{align*}
 Taking the total expectation on both sides leads to
 \begin{align*}
     \mathbb{E}\Expectation{n}{\norm{x_{n+1} - x^{*}}}  & =  \mathbb{E}\norm{x_{n+1} - x^{*}} \leq \frac{M}{2\mu}\mathbb{E}[\norm{x_{n} - x^{*}}^2  +  (\frac{\sigma}{\mu\sqrt{\beta}}+  \eta_{n} L)\mathbb{E}[\norm{x_n - x^{*}}]\\
     & \leq \frac{M\gamma}{2\mu}\mathbb{E}\norm{x_{n} - x^{*}}\mathbb{E}\norm{x_{n} - x^{*}} + (\frac{\sigma}{\mu\sqrt{\beta}}+  \eta_{n} L)\mathbb{E}[\norm{x_n - x^{*}}]
 \end{align*}
 We prove the convergence by induction, notice that
 \begin{align*}
     \mathbb{E}\norm{x_{1} - x^{*}} & \leq \frac{M\gamma}{2\mu}\mathbb{E}\norm{x_{0} - x^{*}}\mathbb{E}\norm{x_{0} - x^{*}}  +  (\frac{\sigma}{\mu\sqrt{\beta}}+  \eta_{n} L)\mathbb{E}[\norm{x_0 - x^{*}}]\\
     &\leq \Big(\frac{M\gamma}{2\mu}\mathbb{E}\norm{x_{0} - x^{*}} + \frac{\sigma}{\mu\sqrt{\beta}}+  \eta L \Big)\mathbb{E}[\norm{x_0 - x^{*}}]\\
     &\leq \Big(\frac{M\gamma}{2\mu}*\frac{\mu}{2M\gamma} + \frac{\sigma}{\mu\sqrt{\frac{16\sigma^2}{\mu}}} + \frac{1}{4L}*L \Big)\mathbb{E}\norm{x_{0} - x^{*}} = \frac{3}{4}\mathbb{E}\norm{x_{0} - x^{*}}
 \end{align*}
 Now assume inequality \ref{ineq:stochastic-bound} holds for $n-th$ iteration, we prove it for $n+1$-th iteration
 \begin{align*}
     \mathbb{E}\Expectation{n}{\norm{x_{n+1} - x^{*}}} 
     & \leq \Big(\frac{M\gamma}{2\mu}\mathbb{E}\norm{x_{n} - x^{*}}+\frac{\sigma}{\mu\sqrt{\beta}}+  \eta L\Big) \mathbb{E}[\norm{x_n - x^{*}}]\leq \frac{3}{4}\mathbb{E}[\norm{x_n - x^{*}}]
 \end{align*}
 \end{proof}
 \section{Experimental Details and More Experiments}
\label{Sec:moreexp}
In this section, we first include more experimental details that could not be placed in the main paper due to the space limitation. We then present more experimental results of NLTGCR for different settings and difficult problems.
\subsection{Experimental Details}
We provide codes implemented in both Matlab and Python. All experiments were run on a Dual Socket Intel E5-2683v3 2.00GHz CPU with 64 GB memory and NVIDIA GeForce RTX 3090.

For linear problems considered in Section \ref{sec:linear}, $\mathbf{A}, \mathbf{b}, \mathbf{c}$, and initial points are generated using normally distributed random number. We use $\mathbf{A}^T\mathbf{A}+\alpha \mathbf{I}$ to generate symmetric matrices. The step size is set as 1 after rescaling $\mathbf{A}$ to have the unit 2-norm. For solving linear equations, we depict convergence by use of the norm of residual, which is defined as $\lVert \mathbf{b} - \mathbf{A}\mathbf{x} \rVert$. For solving bilinear games, we depict convergence by use of the norm of distance to optima, which is defined as $\lVert \mathbf{w}^{\ast} - \mathbf{w}_t \rVert$. For most baselines, we use the Matlab official implementation. 

The softmax regression problem considered in Section \ref{sec:softmax} is defined as follows,

\begin{equation}
f =-\frac{1}{s} \sum_{i=1}^{s} \log \left(\frac{e^{w_{y_{j}}^{T} x^{(i)}}}{\sum_{j=1}^{k} e^{w_{j}^{T} x^{(i)}}}\right),
\label{eqn:softmax}
\end{equation}
where $s$ is the total number of sample, $k$ is the total number
of classes, $x^{(i)}$ is vector of all features of sample $i$, $w_{j}$ is the weights for the $j^{t h}$ class, and $y_{j}$ is the correct class for the $i^{th}$ sample.

\subsection{TGCR(1) for linear system}
\label{appendex:lin}
{We first test the robustness of TGCR(1) for solving linear systems by running with 50 different initials. Figure \ref{fig:multipleLinear} indicates \methodName~ converge well regardless of initialization. We then compare the performance on bilinear games with Anderson Acceleration as \cite{https://doi.org/10.48550/arxiv.2110.02457} shows AA outperforms existing methods on such problems. }
\begin{figure*}[ht]
\centering
\begin{subfigure}[b]{0.45\textwidth}
\centering
\includegraphics[width=0.9\linewidth,]{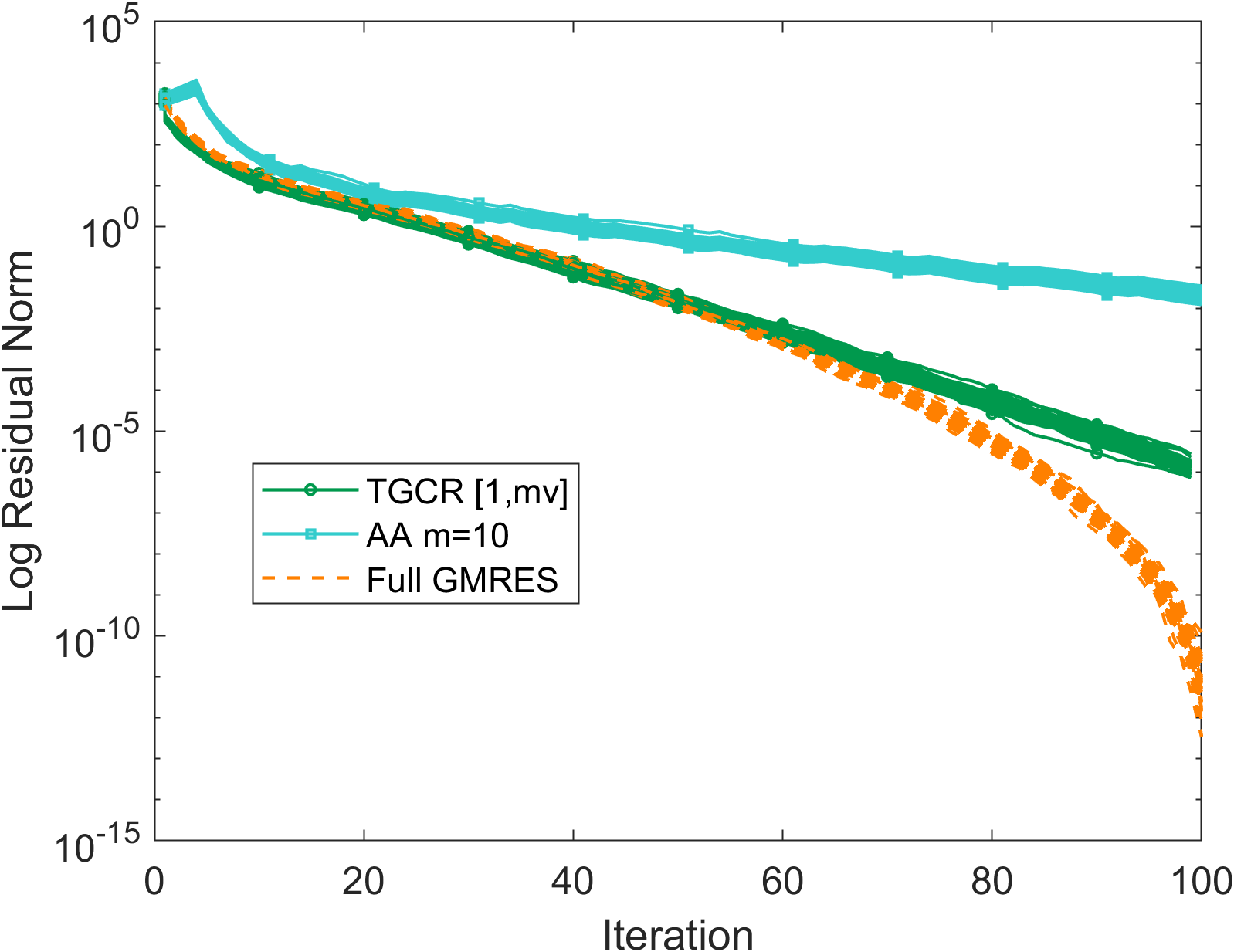}
\subcaption{$\mathbf{A}$ is SPD} 
\label{fig:linmul}
\end{subfigure}
\begin{subfigure}[b]{0.45\textwidth}
\centering
\includegraphics[width=0.9\linewidth]{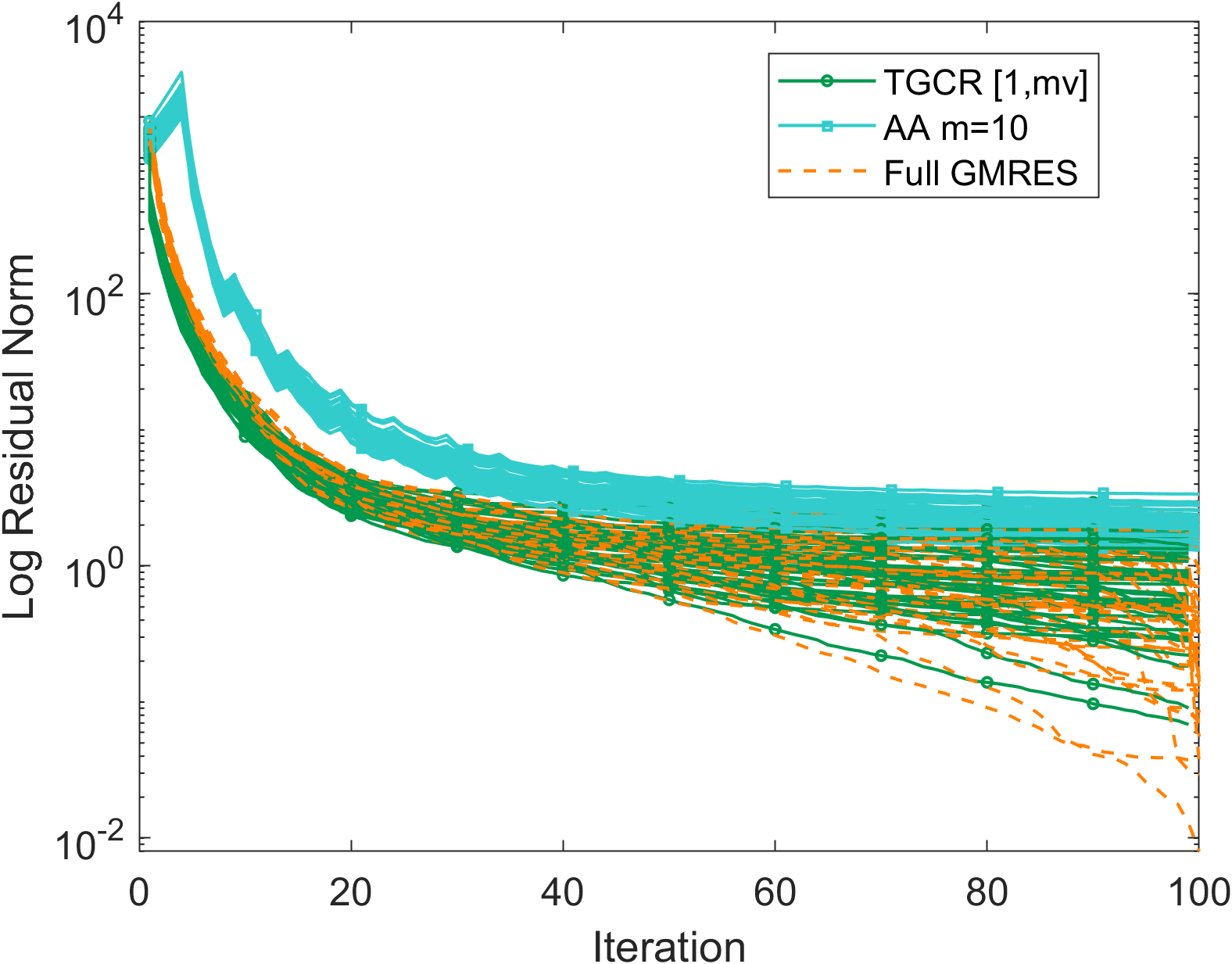}
\subcaption{$\mathbf{A}$ is symmetric indefinite} 
\label{fig:lin2mul}
\end{subfigure}

\caption{{\textbf{Linear Systems}  $\mathbf{A}\mathbf{x}=\mathbf{b}, \mathbf{A} \in \mathbb{R}^{100\times100}$: Comparison in terms of iteration over 50 random runs. We can observe that TGCR(1) match full memory GMRES in the first stage and consistently converge faster than AA(10).}}
\label{fig:multipleLinear}
\end{figure*}
\paragraph{Minimax Optimization.} \ 
Next, we test \methodName~ on the following  zero-sum bilinear games:
\begin{equation}
\label{eq:bilinear}
    \min_{\mathbf{x}\in \mathbb{R}^n}\max_{\mathbf{y}\in \mathbb{R}^n}f(\mathbf{x},\mathbf{y}) =  \mathbf{x}^{T}\mathbf{A}\mathbf{y} + \mathbf{b}^{T}\mathbf{x} + \mathbf{c}^T\mathbf{y}, \quad \mathbf{A}\textrm{ is full rank.}
\end{equation}

Bilinear games are often regarded as an important but simple class 
of problems for theoretically analyzing and understanding algorithms for solving general minimax problems \cite{Zhang2020ConvergenceOG,sbb}. Here we consider simultaneous GDA mapping for minimax bilinear games 
$\bigl(\begin{smallmatrix}
\mathbf{I} & -\eta\mathbf{A}\\
\eta\mathbf{A}^{T} &\mathbf{I}
\end{smallmatrix}\bigr)$ \cite{https://doi.org/10.48550/arxiv.2110.02457}.
Although this mapping is skew-symmetric, TGCR can still exploit the short-term recurrence. It can be observed from Figure \ref{fig:minimax} that Krylov subspace methods such as \methodName~ and AA converge fast for bilinear problem when $A$ is either SPD or random generated.  More importantly,  Figure \ref{fig:minimax} demonstrates that \methodName~$(1)$ exhibits a \emph{superlinear} convergence rate and converges to optimal significantly faster than AA$(m)$ in terms of both iteration number and computation time. 

\begin{figure}[ht]
\centering
\begin{subfigure}[b]{0.32\textwidth}
\centering
\includegraphics[width=0.99\linewidth]{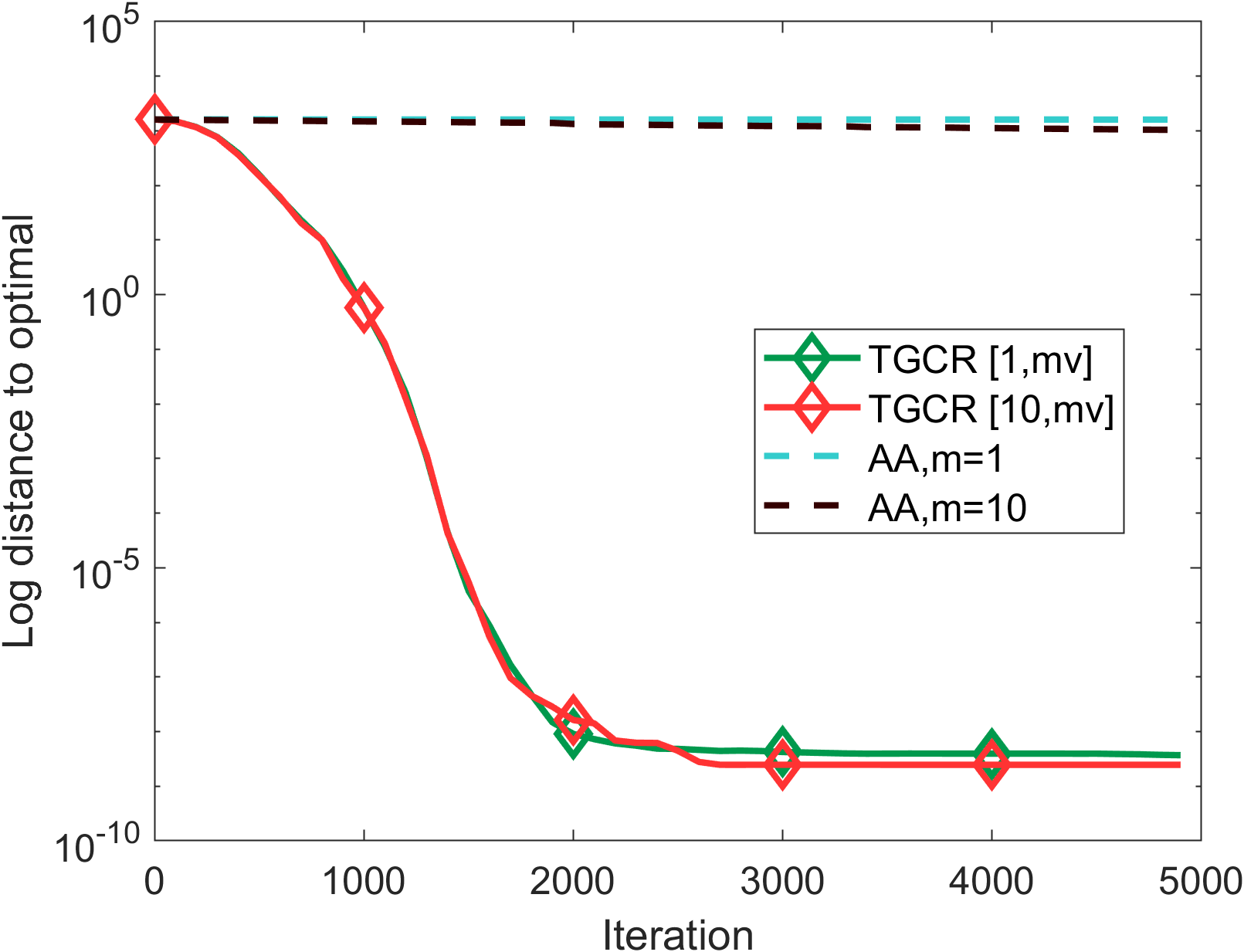}
\subcaption{$\mathbf{A}$ is SPD} 
\label{fig:bi_iter}
\end{subfigure}
\begin{subfigure}[b]{0.32\textwidth}
\centering
\includegraphics[width=0.95\linewidth]{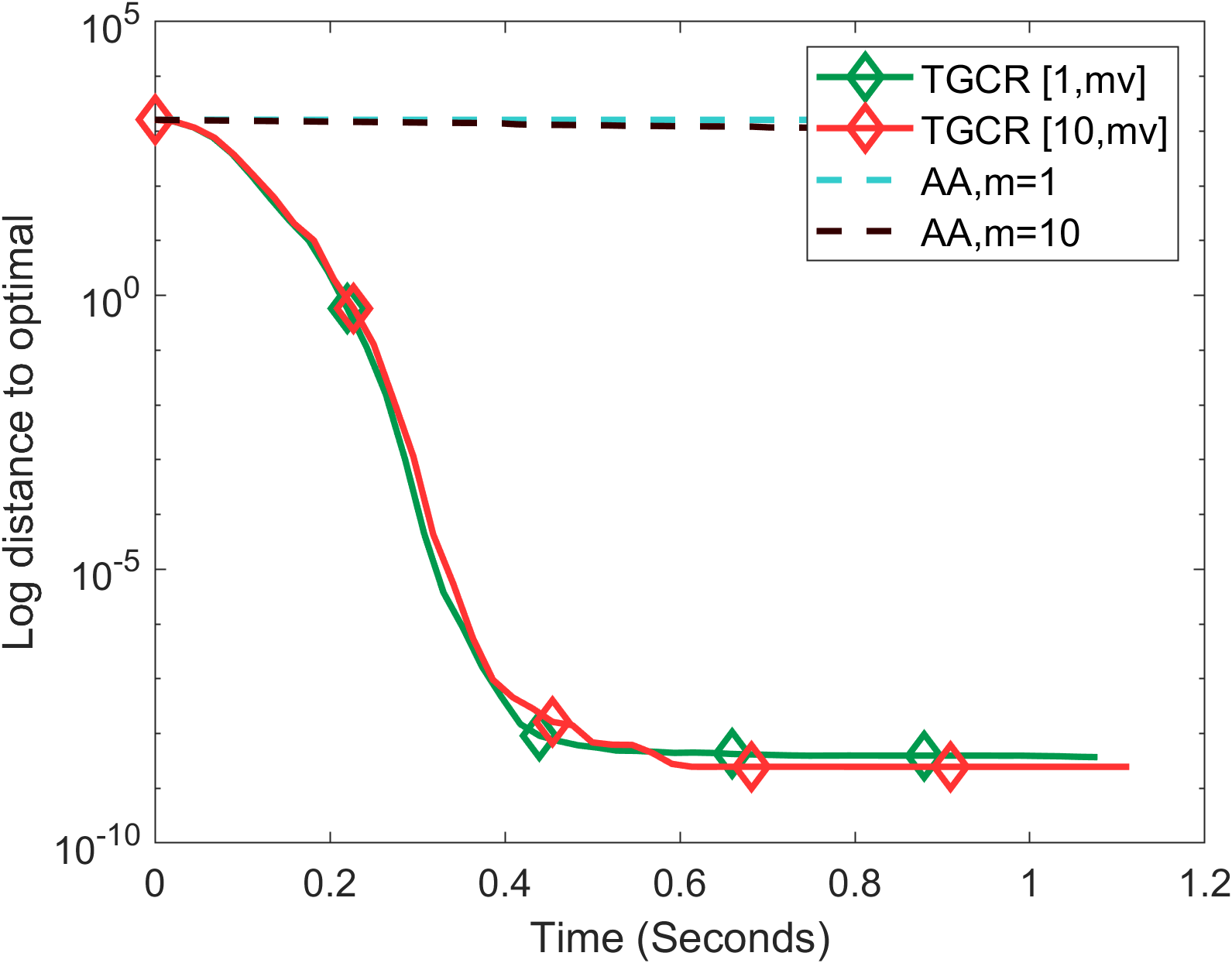}
\subcaption{Time Comparison (\ref{fig:bi_iter})} 
\label{fig:bi_time}
\end{subfigure}
\begin{subfigure}[b]{0.32\textwidth}
\centering
\includegraphics[width=0.99\linewidth]{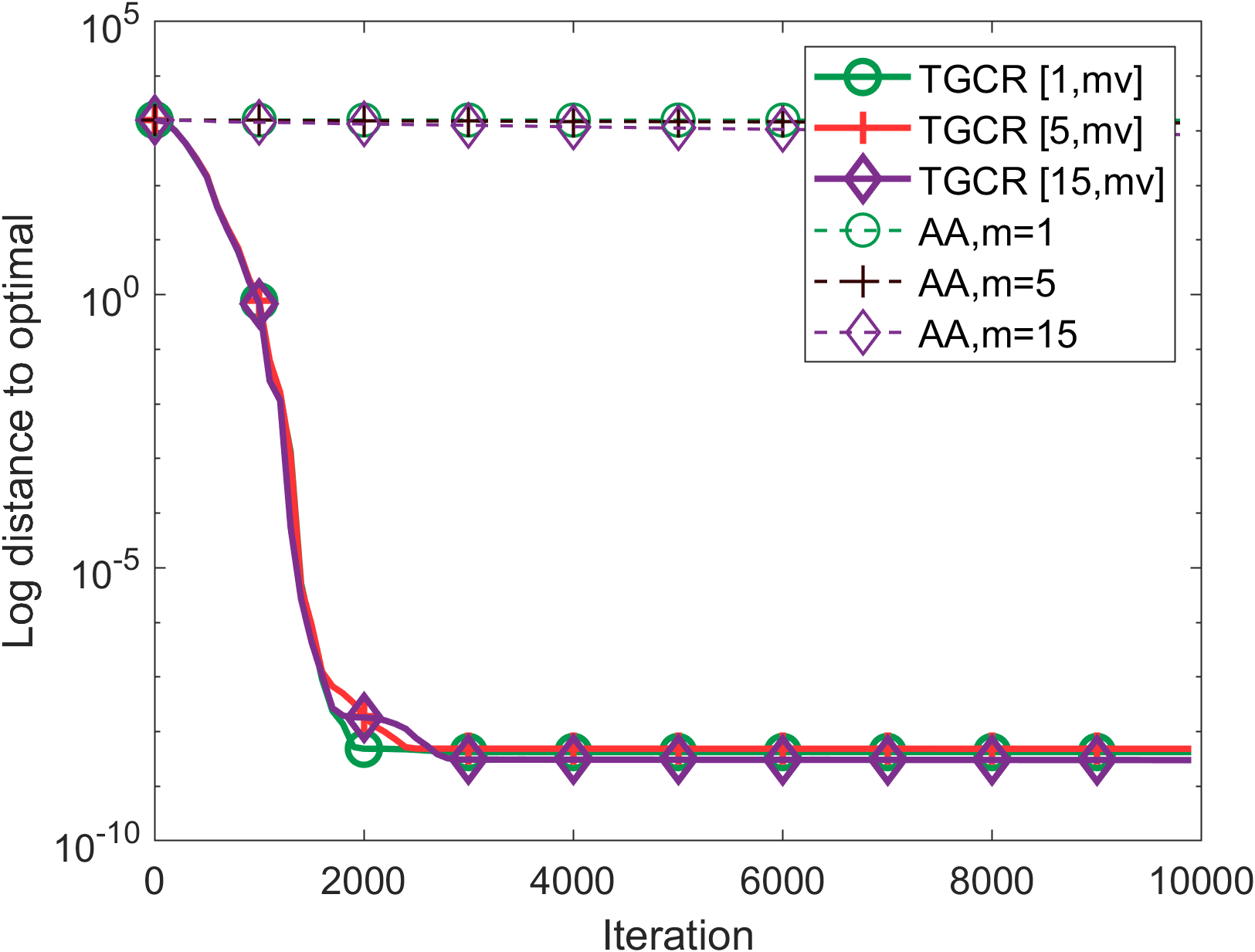}
\subcaption{$\mathbf{A}$ is random generated} 
\label{fig:Arand}
\end{subfigure}

\caption{\textbf{Bilinear Problems:}  \ref{fig:bi_iter}: Distance to optimal vs. Iterations; \ref{fig:bi_time}: Distance to optimal vs. Time; \ref{fig:Arand}: Distance to optimal vs. Iterations. It can be observed that the short-term property (Theorem \ref{thm:m1}) holds as long as the mapping is symmetric or skew-symmetric.}

\label{fig:minimax}
\end{figure}

\subsection{TGCR(1) for nonsymmetric quadratic minimization and linear system}

A quadratic form is simply a scalar, quadratic function of a vector with the form
\begin{equation}
f(x)=\frac{1}{2} \mathbf{x}^{T} \mathbf{A} \mathbf{x}-\mathbf{b}^{T} \mathbf{x}+c,
\label{eqn:quad}
\end{equation}
where $\mathbf{A}$ is a matrix, $\mathbf{x}$ and $\mathbf{b}$ are vectors, and $c$ is a scalar constant. When $\mathbf{A}$ is  symmetric
and positive-definite, $f(x)$  is minimized by the solution to $\mathbf{A} \mathbf{x}= \mathbf{b}$.

\begin{figure*}[ht]
\centering
\begin{subfigure}[b]{0.45\textwidth}
\centering
\includegraphics[width=0.9\linewidth,]{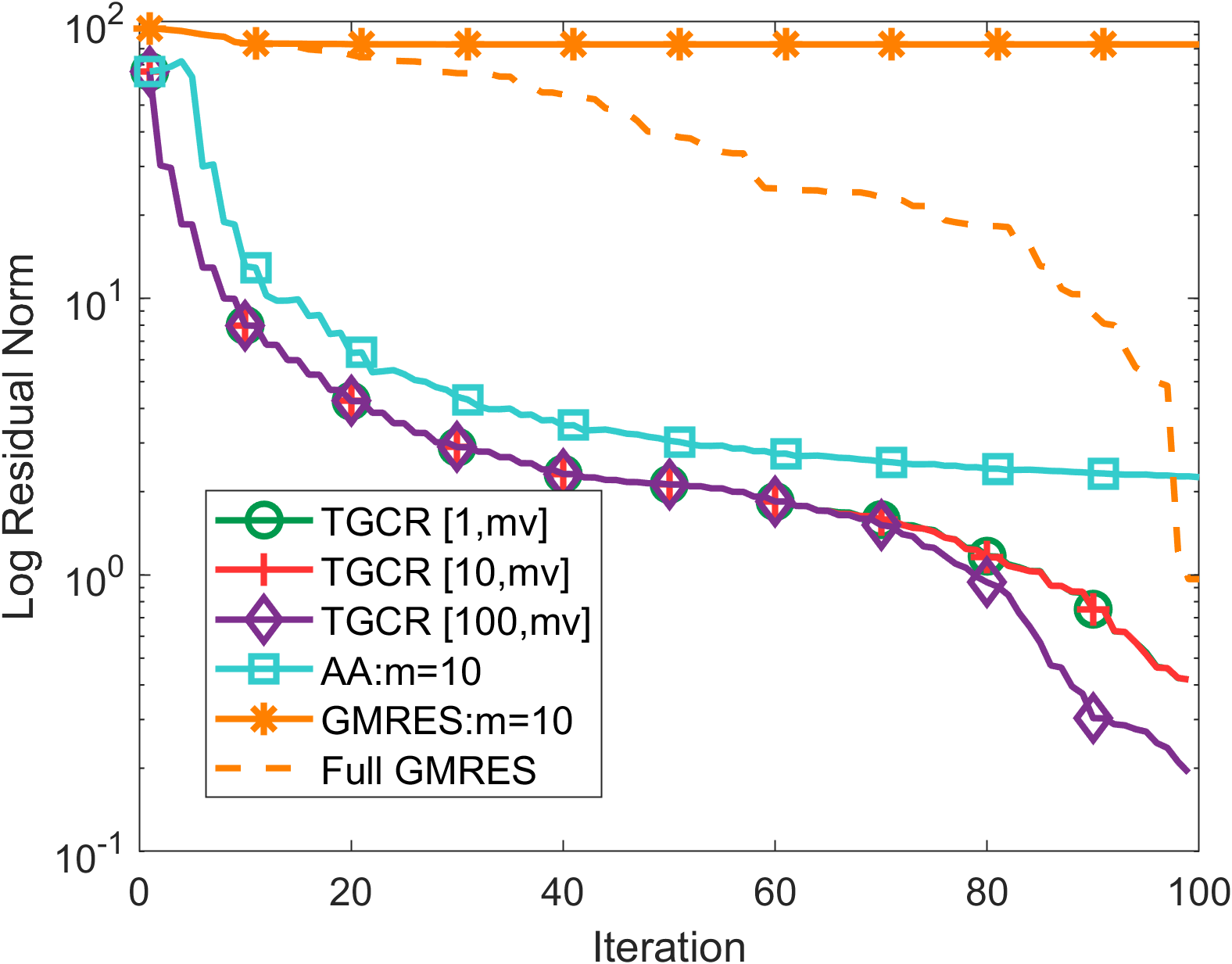}
\subcaption{\textbf{Quadratic form (\ref{eqn:quad})}} 
\label{fig:nonquad}
\end{subfigure}
\begin{subfigure}[b]{0.45\textwidth}
\centering
\includegraphics[width=0.9\linewidth]{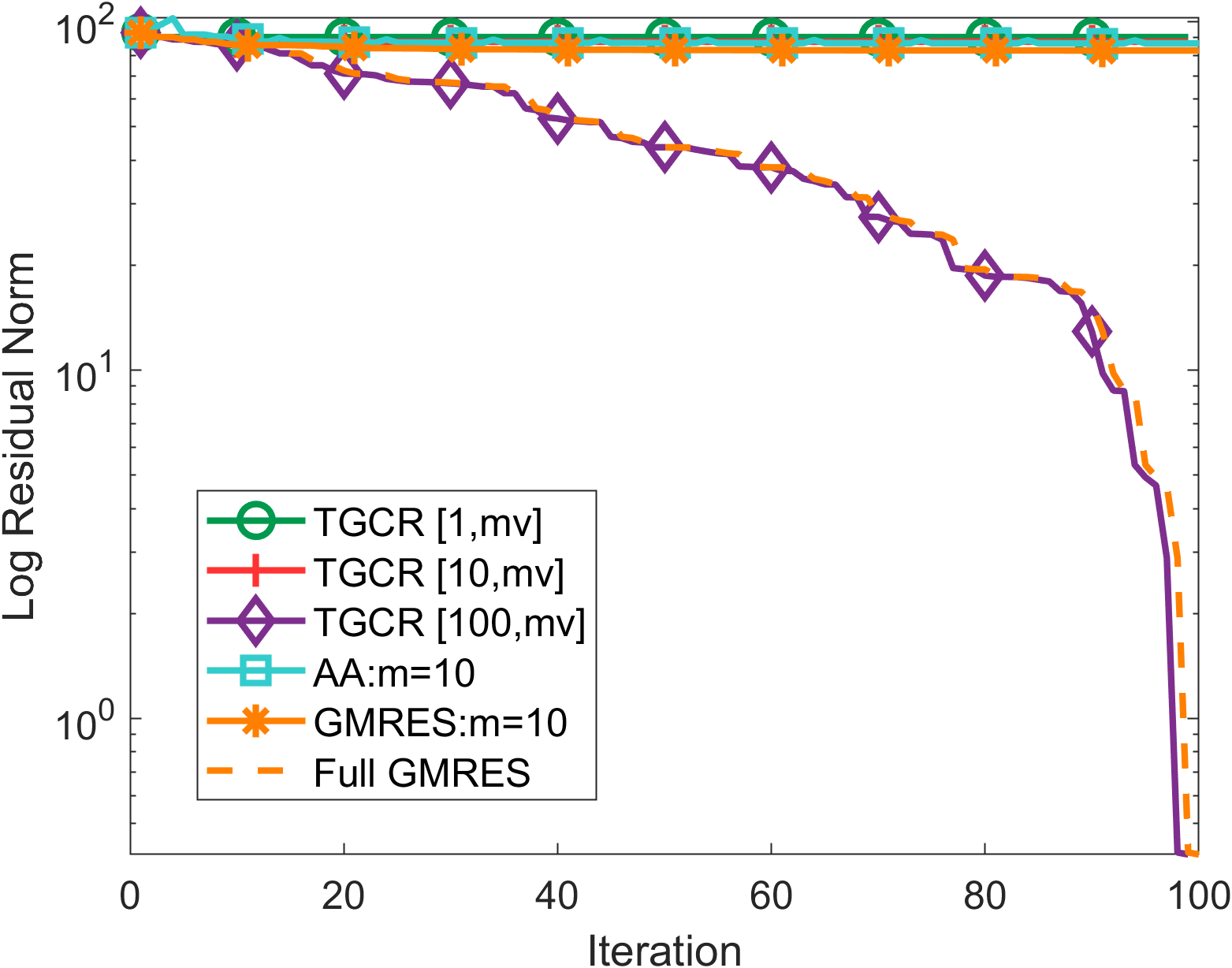}
\subcaption{Linear System $\mathbf{A} \mathbf{x}= \mathbf{b}$} 
\label{fig:nonsymLin}
\end{subfigure}

\caption{$\mathbf{A}$ is nonsymmetric. In Figure \ref{fig:nonquad}, we can find TGCR (1) still converges fast  because the Hessian of Equation \ref{eqn:quad} is $\frac{1}{2}(A^{T}+A)$, which is still symmetric. In Figure \ref{fig:nonsymLin}, we can find TGCR(100) has the same convergence rate with GMRES. This shows that TGCR(m) can still converge for solving nonsymmetric linear systems with $m>1$ and is mathematically equivalent to non-restart GMRES when $m$ is equal to the matrix size.}
\label{fig:randAsup}
\end{figure*}

\subsection{Investigation of Stochastic NLTGCR}
In the main paper, we report the effectiveness of NLTGCR for softmax regression on MNIST. We
give the result about using deterministic and stochastic gradients in Figure \ref{fig:softmax} and find that NLTGCR only requires a small batch size and table size ($1$) . In this section, we further investigate the effectiveness of NLTGCR in a stochastic setting, provide additional experimental results in Figure \ref{fig:sgdsoftmax2} and Figure \ref{fig:sgdsoftmax22}. From Figure \ref{fig:sgdsoftmax2}, we can observe that using the same batch size and table size $m=1$, stochastic NLTGCR consistently outperforms stochastic AA. Also, it can be observed from Figure \ref{fig:sgdsoftmax22} that stochastic NLTGCR outperforms stochastic AA for different table size using a fixed batch size of 1500. Although the short-term property does not strictly hold in a stochastic setting, we can still observe that NLTGCR outperforms AA with a smaller variance. In addition, it is worth noting that a smaller table size works better for both NLTGCR and AA. We suspect this is due to the accumulation of inaccurate gradient estimates. 

\begin{figure*}[ht]
\centering
\begin{subfigure}[b]{0.45\textwidth}
\centering
\includegraphics[width=0.9\linewidth,]{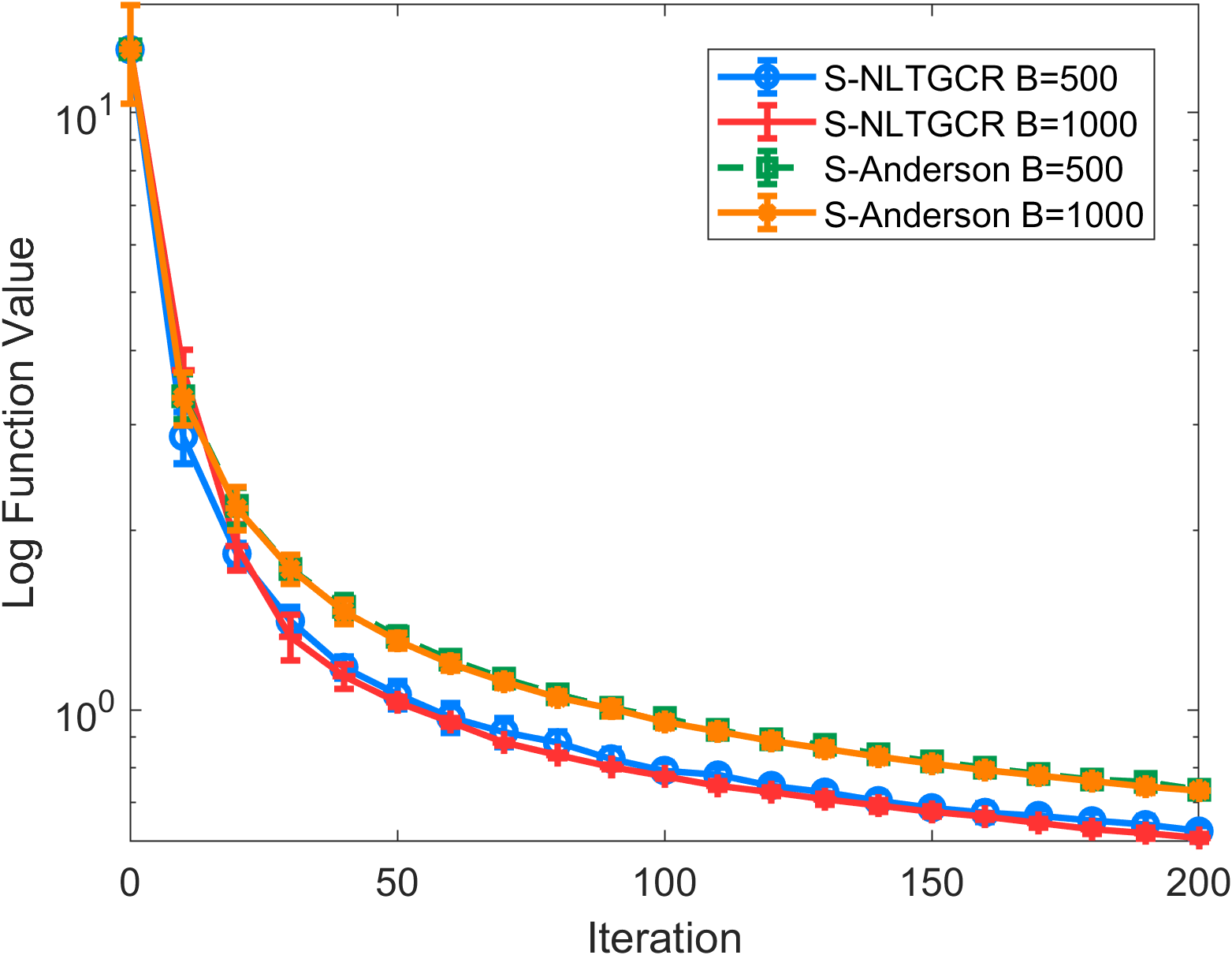}
\subcaption{Loss} 
\label{fig:sgdm1loss}
\end{subfigure}
\begin{subfigure}[b]{0.45\textwidth}
\centering
\includegraphics[width=0.9\linewidth]{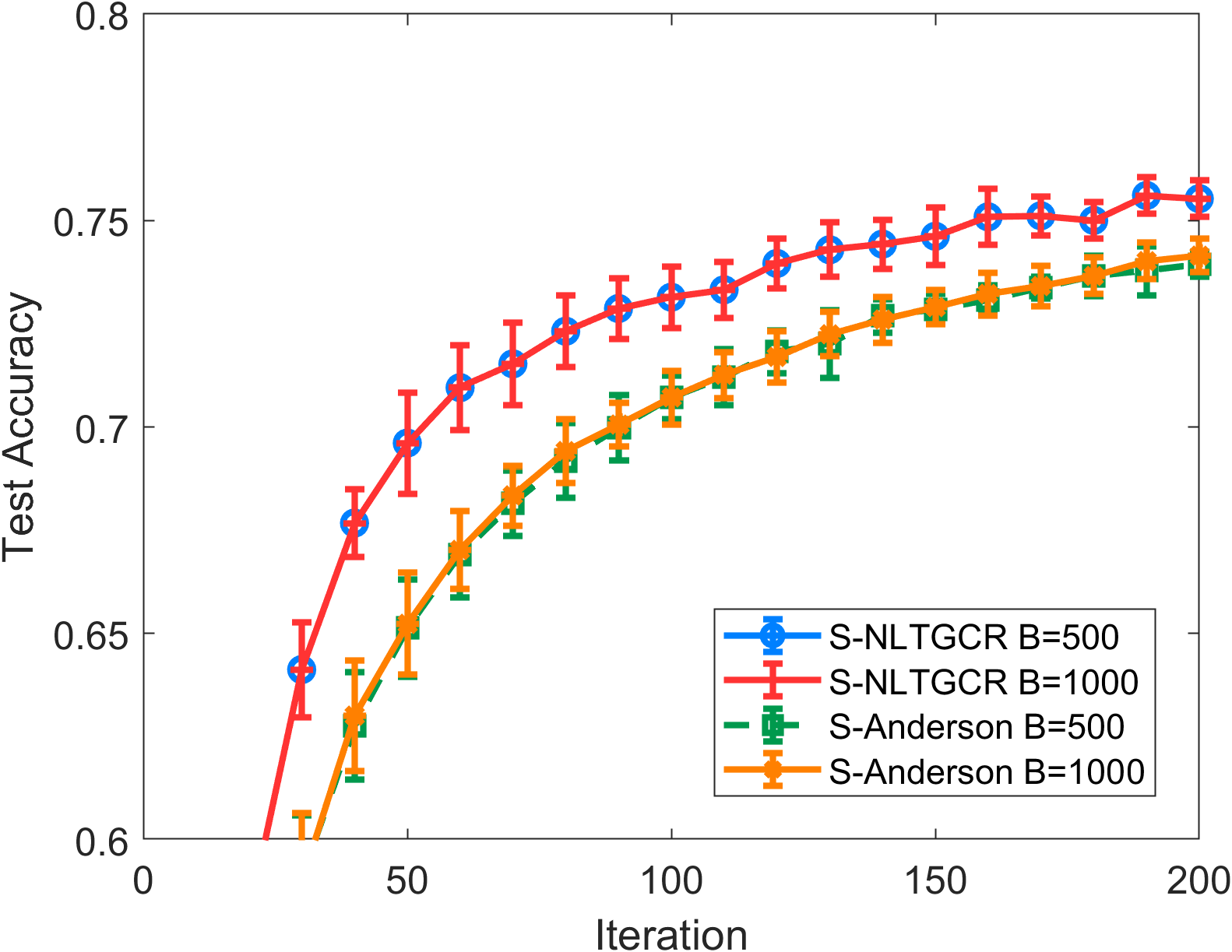}
\subcaption{Test Accuracy} 
\label{fig:sgdm1acc}
\end{subfigure}

\caption{\textbf{Softmax Regression: Effects of batch size}, $m=1$}
\label{fig:sgdsoftmax2}
\end{figure*}

\begin{figure*}[ht]
\centering
\begin{subfigure}[b]{0.45\textwidth}
\centering
\includegraphics[width=0.9\linewidth,]{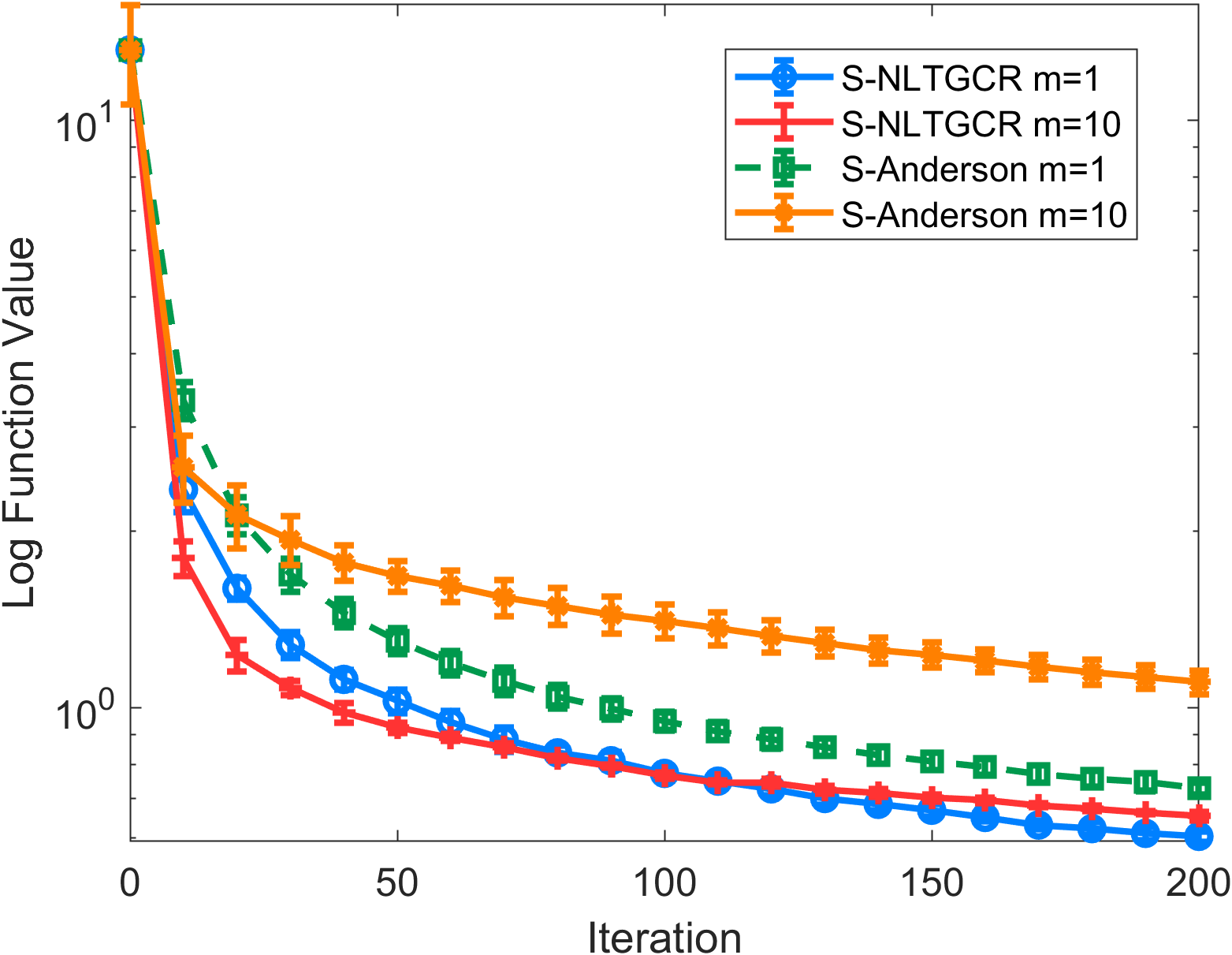}
\subcaption{ Loss} 
\label{fig:sgdm10loss}
\end{subfigure}
\begin{subfigure}[b]{0.45\textwidth}
\centering
\includegraphics[width=0.9\linewidth]{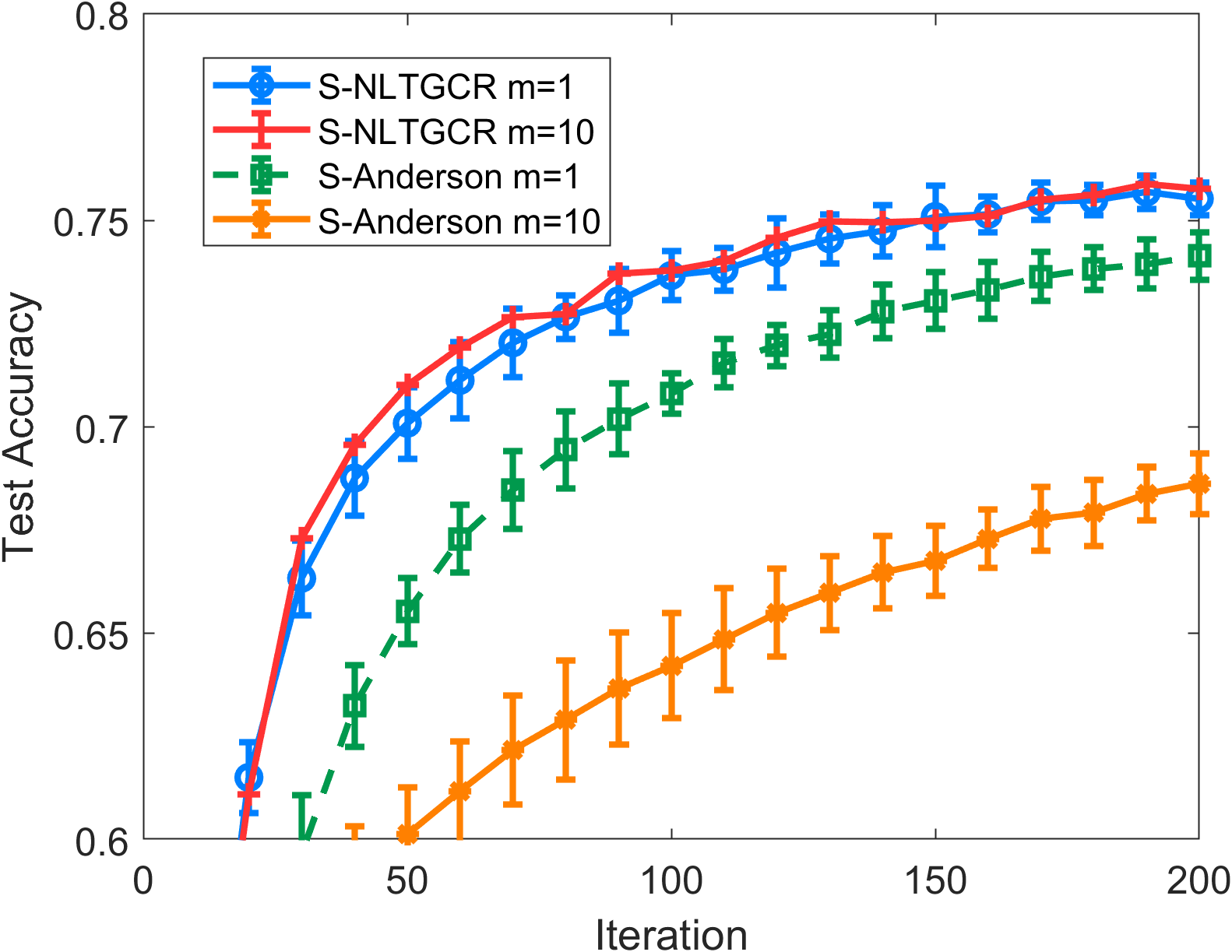}
\subcaption{Test Accuracy} 
\label{fig:sgdm10acc}
\end{subfigure}

\caption{\textbf{Softmax Regression: Effects of table size in a stochastic setting}, $B=1000$}
\label{fig:sgdsoftmax22}
\end{figure*}

\paragraph{Compatible with Momentum}
Another important technique in optimization is momentum, which speeds up convergence significantly both in theory and in practice. We experimentally show that it is possible to further accelerate the convergence of NLTGCR by using Momentum. We run stochastic NLTGCR with different momentum term and present results in Figure \ref{fig:sgdmom22}. It suggests that by incorporating momentum into NLTGCR momentum further accelerates the convergence, although the variance gets larger for a large momentum term $v$. We leave the theoretical analysis of of NLTGCR with momentum for future work. 

\begin{figure*}[ht]
\centering
\begin{subfigure}[b]{0.45\textwidth}
\centering
\includegraphics[width=0.9\linewidth,]{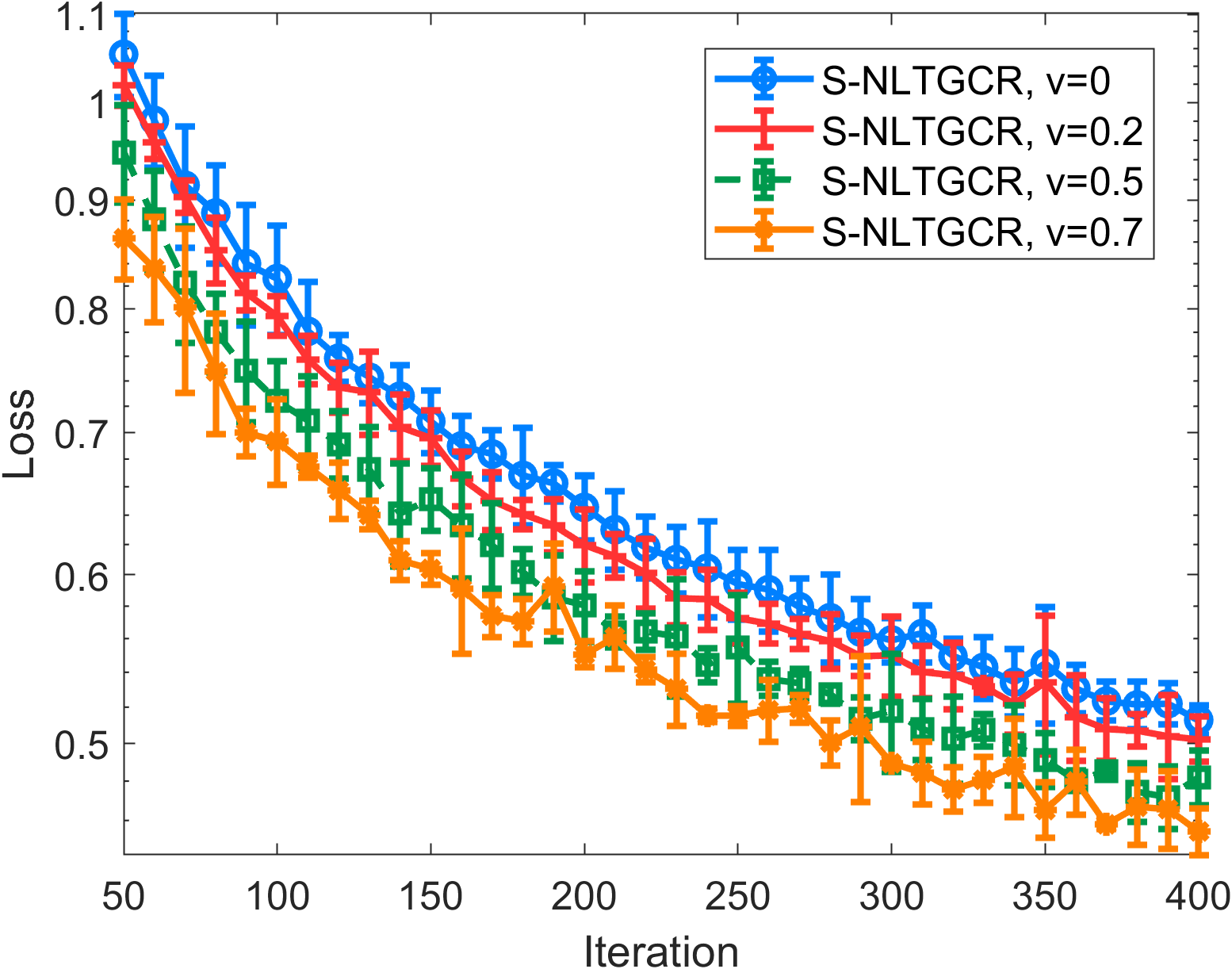}
\subcaption{ Loss} 
\label{fig:sgdmomloss}
\end{subfigure}
\begin{subfigure}[b]{0.45\textwidth}
\centering
\includegraphics[width=0.9\linewidth]{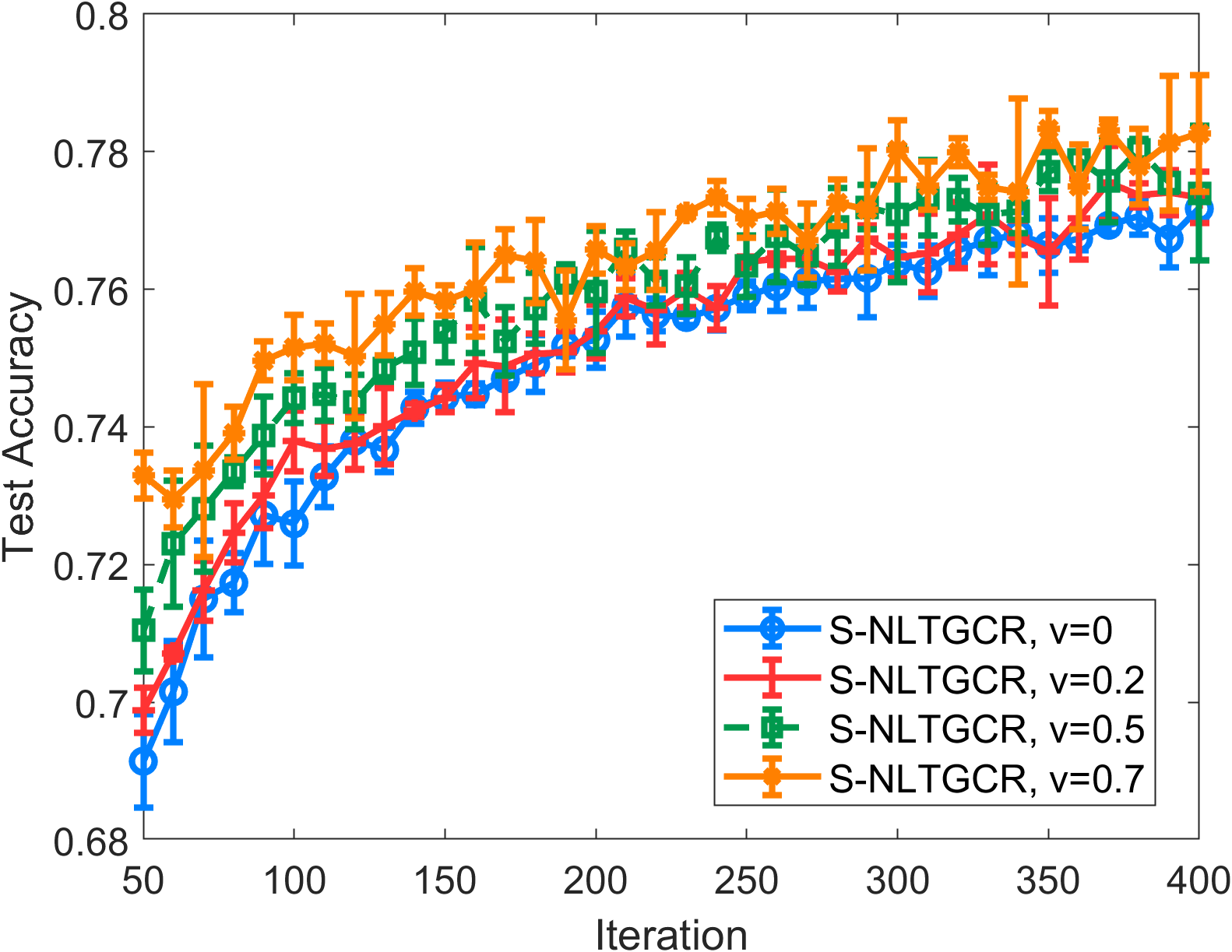}
\subcaption{Test Accuracy} 
\label{fig:sgdmomacc}
\end{subfigure}

\caption{\textbf{Softmax Regression: Compatibile with momentum.} We further test the acceleration effect of momentum on stochastic NLTGCR. It shows that stochastic NLTGCR with momentum converges faster than the one without momentum.}
\label{fig:sgdmom22}
\end{figure*}

\subsection{Results for Deep learning applications }
\label{supsec:DL}
\subsubsection{Image classification using CNN}
In a more realistic setting, we test our algorithm for neural networks on an image classification task. Particularly, we use the
standard MNIST dataset \footnote{http://yann.lecun.com/exdb/mnist/}. The architecture of the network is based on the official PyTorch implementation \footnote{implementation https://github.com/pytorch/examples/blob/master/mnist.}. We tried our best to ensure that the baselines had the best performance in the tests.  Hyperparamters are selected after grid search. We use a batch size of $64$. For SGD, Nestrov ($v=0.9$),  Adam (default $\beta_1$ and $\beta_2$), and NLTGCR $(m=1)$, we use a learning rate of $1 \times 10^{-2}$, $1 \times 10^{-1}$, $1 \times 10^{-3}$, and $1 \times 10^{-3}$, respectively. Figure \ref{fig:mnistlossDL} shows the curves of loss for training the neural network on MNIST.  Figure \ref{fig:mnistaccDL} shows the curves of test accuracy on MNIST. It can be found that NLTGCR outperforms SGD and Nestrov and is comparable to Adam. In addition, we conduct experiments on the effects of table size for NLTGCR and present results in Figure \ref{fig:mnistDL2}. Although Figure \ref{fig:mnistaccDL2}, shows $m=10$ does slightly better, we found that $m=1$ generally yields robust results. In addition, $m=1$ significantly reduces the memory and computation overhead. As a result, we would like to suggest $m=1$ for general experiments. These preliminary results provide insights of the effectiveness of our algorithm for training neural networks. Our algorithm is comparable with the widely used optimizer, Adam. In addition, our algorithm is more memory and computation efficient than other nonlinear acceleration methods including AA and RNA. As a result, it is worth investigating the performance of NLTGCR for different tasks and more complex networks. We leave it for our future work. 
\begin{figure}[ht]
\centering
\begin{subfigure}[b]{0.45\textwidth}
\centering
\includegraphics[width=0.9\linewidth,]{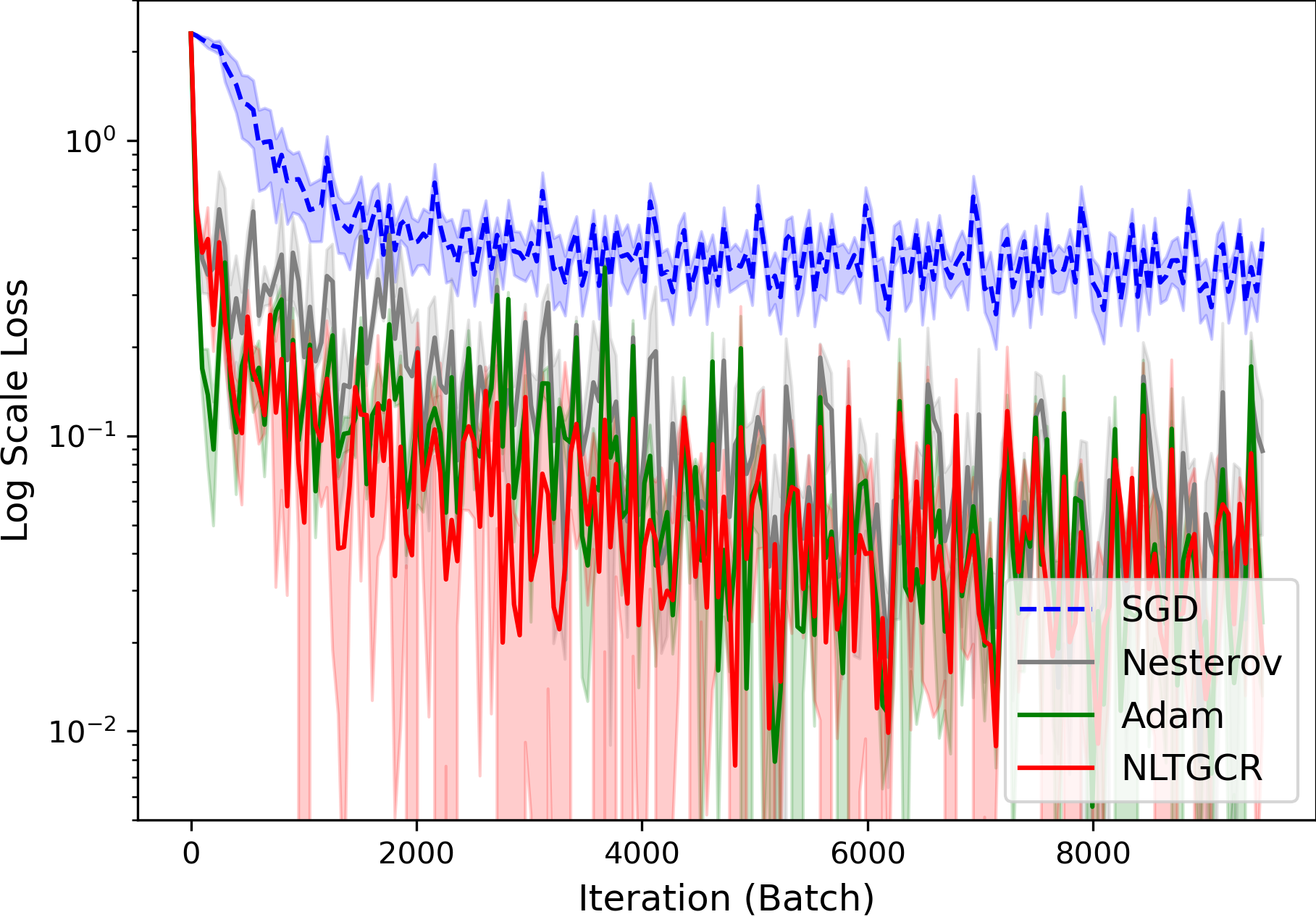}
\subcaption{Loss} 
\label{fig:mnistlossDL}
\end{subfigure}
\begin{subfigure}[b]{0.45\textwidth}
\centering
\includegraphics[width=0.9\linewidth]{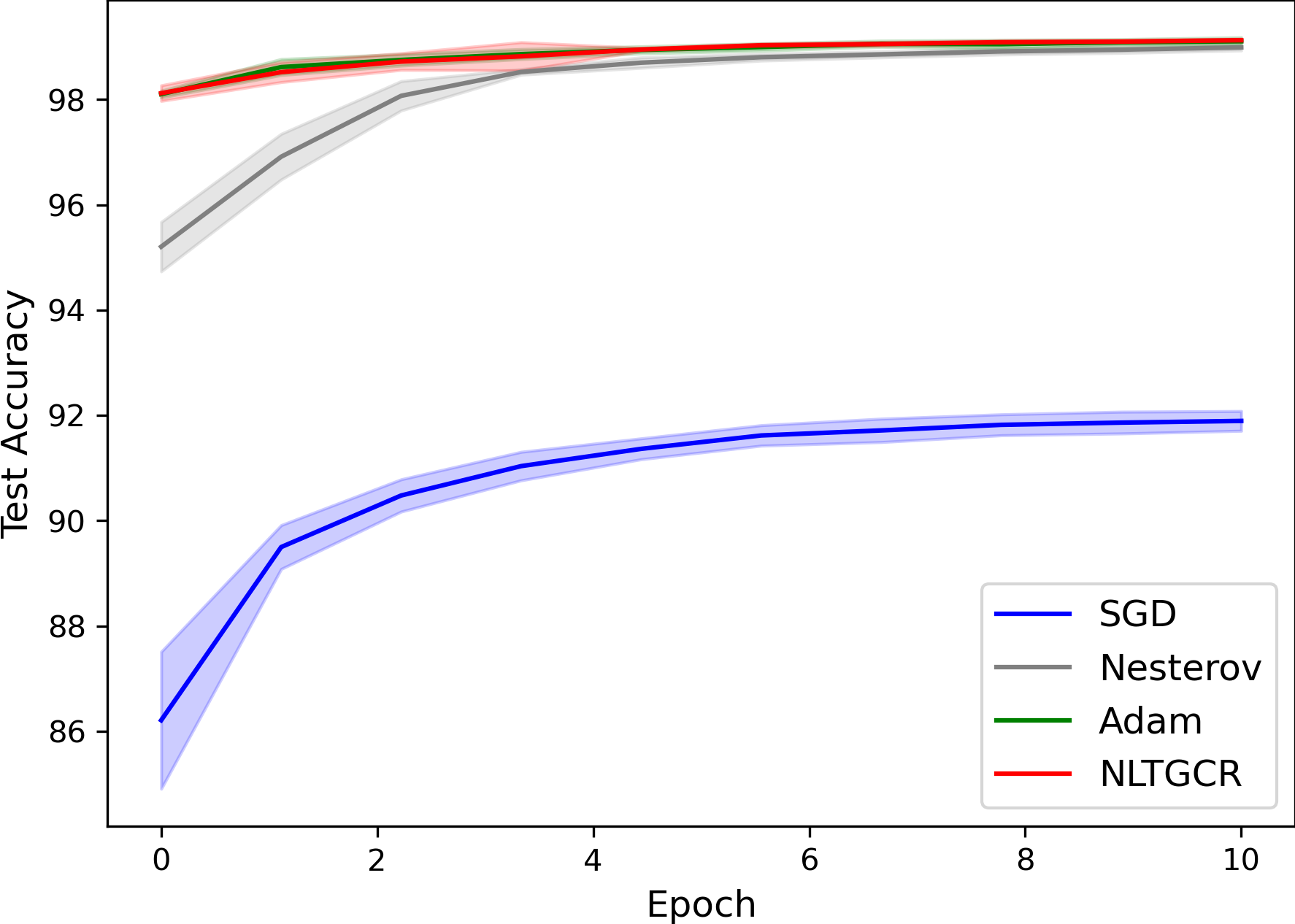}
\subcaption{Test Accuracy} 
\label{fig:mnistaccDL}
\end{subfigure}

\caption{\textbf{Training on MNIST.} Averaged on 5 runs, $m=1$ for our algorithm. }
\label{fig:mnistDL}
\end{figure}

\begin{figure}[ht]
\centering
\begin{subfigure}[b]{0.45\textwidth}
\centering
\includegraphics[width=0.9\linewidth,]{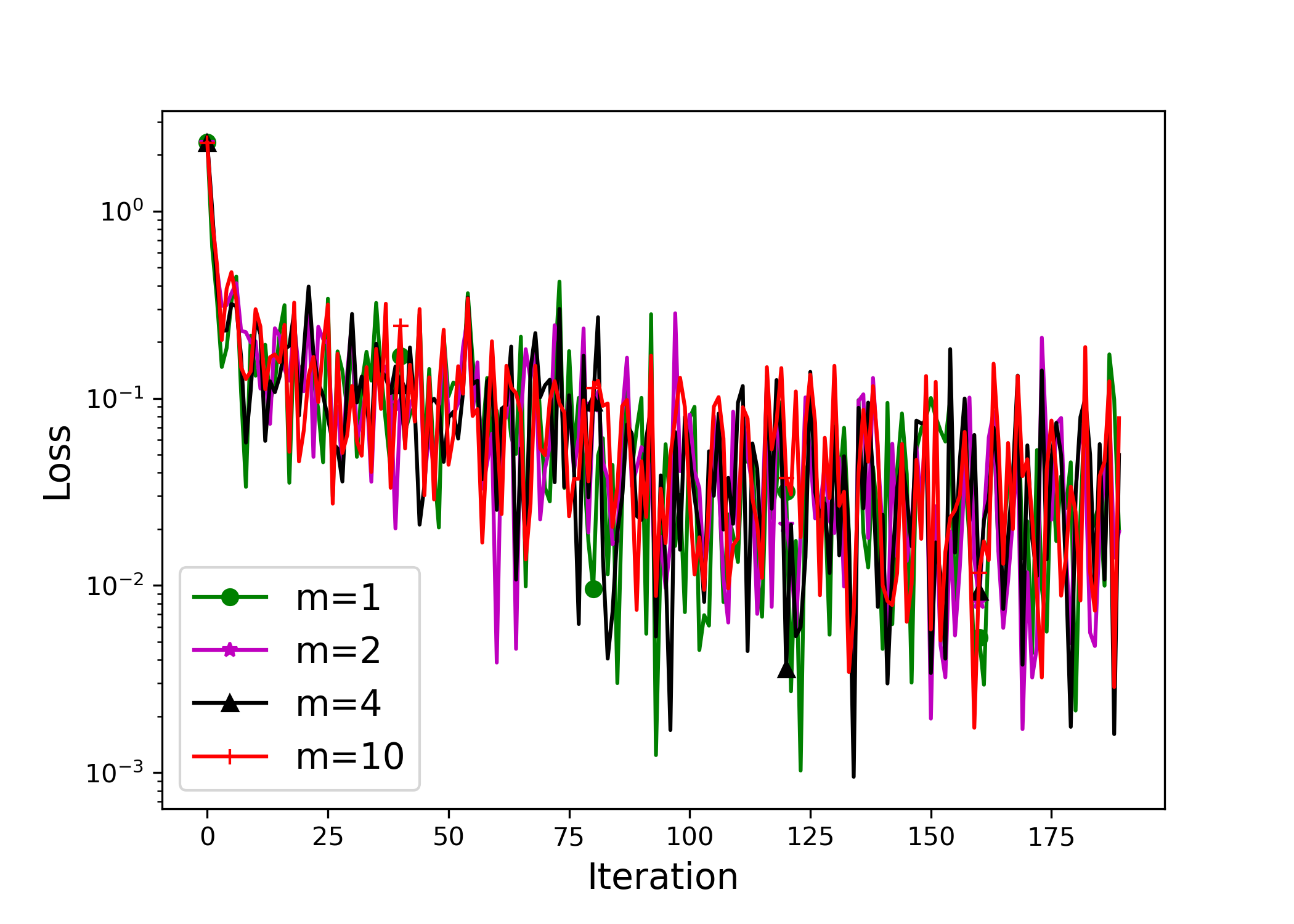}
\subcaption{Loss} 
\label{fig:mnistlossDL2}
\end{subfigure}
\begin{subfigure}[b]{0.45\textwidth}
\centering
\includegraphics[width=0.9\linewidth]{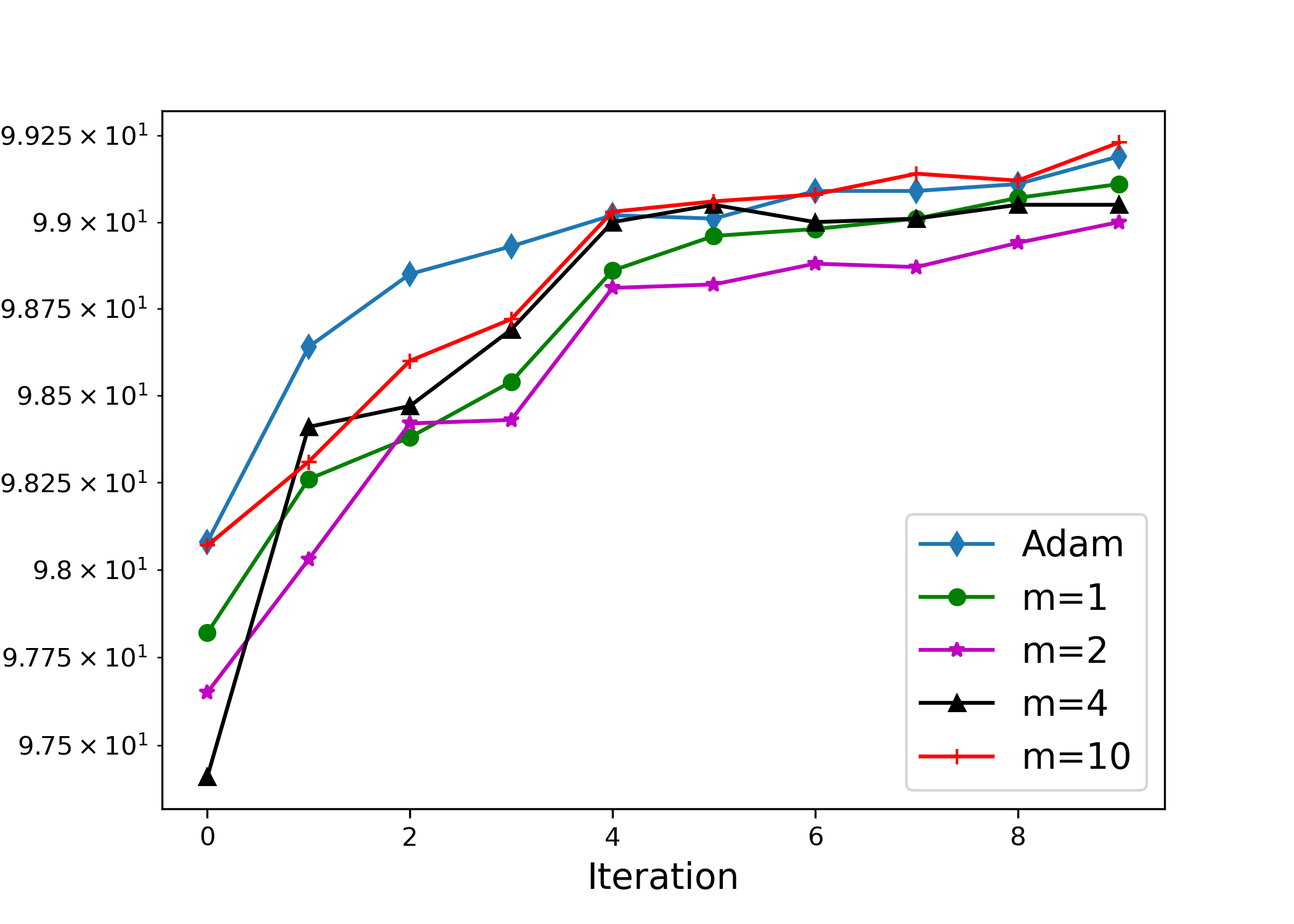}
\subcaption{Test Accuracy} 
\label{fig:mnistaccDL2}
\end{subfigure}

\caption{\textbf{Effects of table size $m$} We run experiments with a fixed seed. It shows S-NLTGCR(10) does slightly better than Adam and S-NLTGCR($m<10)$. We would suggest to use $m=1$ for saving memory and computation.}
\label{fig:mnistDL2}
\end{figure}

\subsubsection{Image classification using ResNet}
{We now perform our tests on ResNet32 \cite{resnet} \footnote{\url{https://github.com/akamaster/pytorch_resnet_cifar10}} using CIFAR10 \cite{cifar}. We randomly split the training set of all the datasets into two subsets, train and validation. The former is used to train the neural network, whereas the latter is used for measuring the performance of the learned model.  Hyperparameters are selected after a grid search. We use a batch size of $128$. For Nesterov ($v=0.9$),  Adam (default $\beta_1$ and $\beta_2$), and NLTGCR $(m=1)$, we use a learning rate of  $3 \times 10^{-4}$, $1 \times 10^{-3}$, and $1 \times 10^{-1}$, respectively. For better visualization, figure \ref{fig:resnetDL} shows the curves of training loss and validation accuracy using 50 epochs. It can be observed that nlTGCR(1) consistently outperforms baselines and has a smaller variance.}
\begin{figure}[ht]
\centering
\begin{subfigure}[b]{0.45\textwidth}
\centering
\includegraphics[width=0.9\linewidth,]{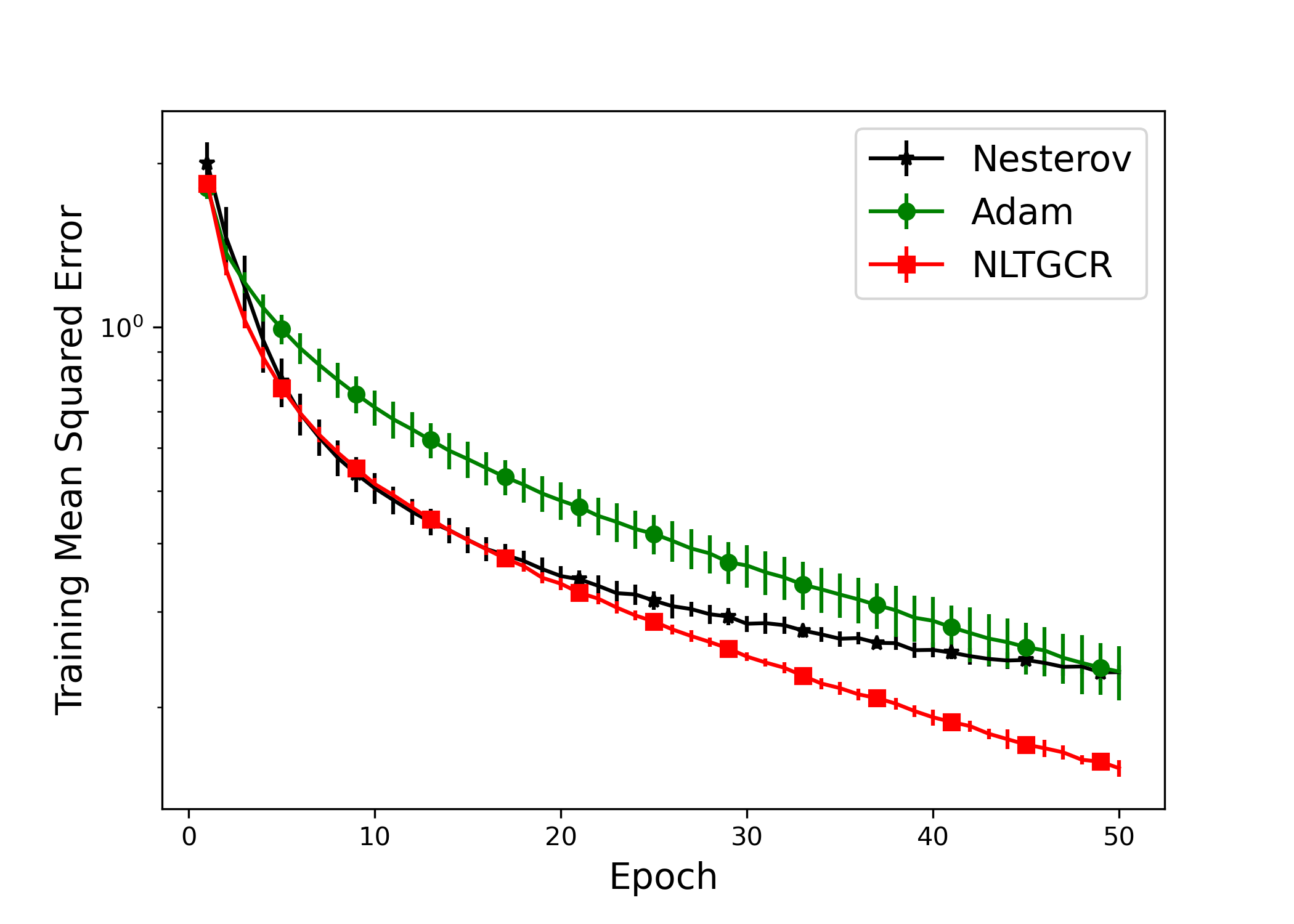}
\subcaption{Training MSE} 
\label{fig:resnetloss}
\end{subfigure}
\begin{subfigure}[b]{0.45\textwidth}
\centering
\includegraphics[width=0.9\linewidth]{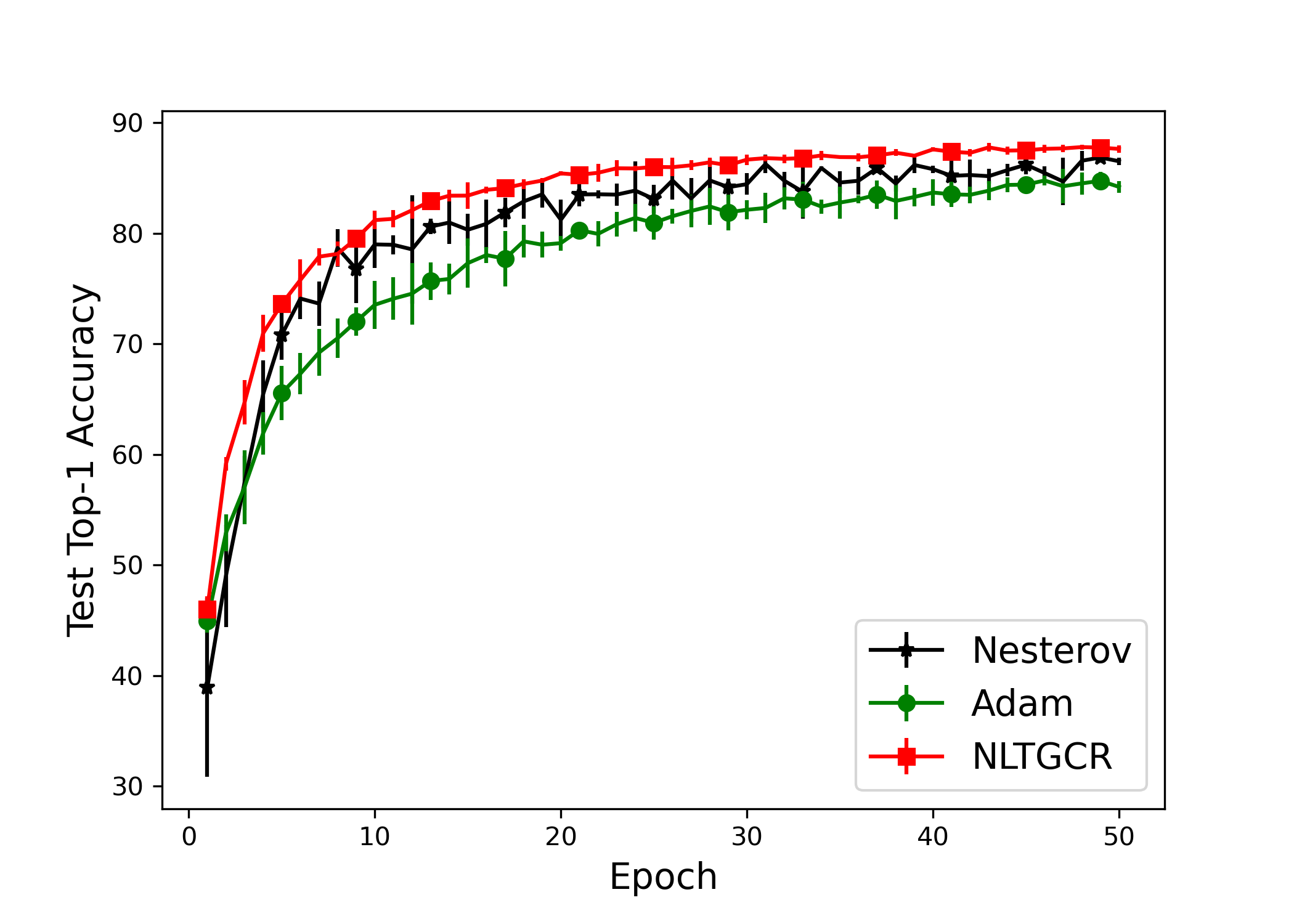}
\subcaption{Validation Accuracy} 
\label{fig:resnetacc}
\end{subfigure}
\caption{{\textbf{Validation MSE of training ResNet32 on CIFAR10.} Averaged on 5 runs (manual random seed 0 to 5 for all methods), $m=1$ for our algorithm. nlTGCR(1) consistently outperforms baselines and has a smaller variance.}}
\label{fig:resnetDL}
\end{figure}


\subsubsection{Time series forecasting using LSTM}
{Next, we test our algorithm for Long Short-Term Memory \cite{lstm} on time series forecasting task using Airplane Passengers and Shampoo Sales Dataset \footnote{\url{https://github.com/spdin/time-series-prediction-lstm-pytorch}}. We use a learning rate of 0.04 and the mean squared error (MSE) as our loss function and evaluation metric. Figure \ref{fig:lstmDL} depicts the MSE on validation set during training. It shows \methodName~ converges better than baselines. It also suggests \methodName~ is capable of optimizing complex deep learning architectures.}
\begin{figure}[ht]
\centering
\begin{subfigure}[b]{0.45\textwidth}
\centering
\includegraphics[width=0.9\linewidth,]{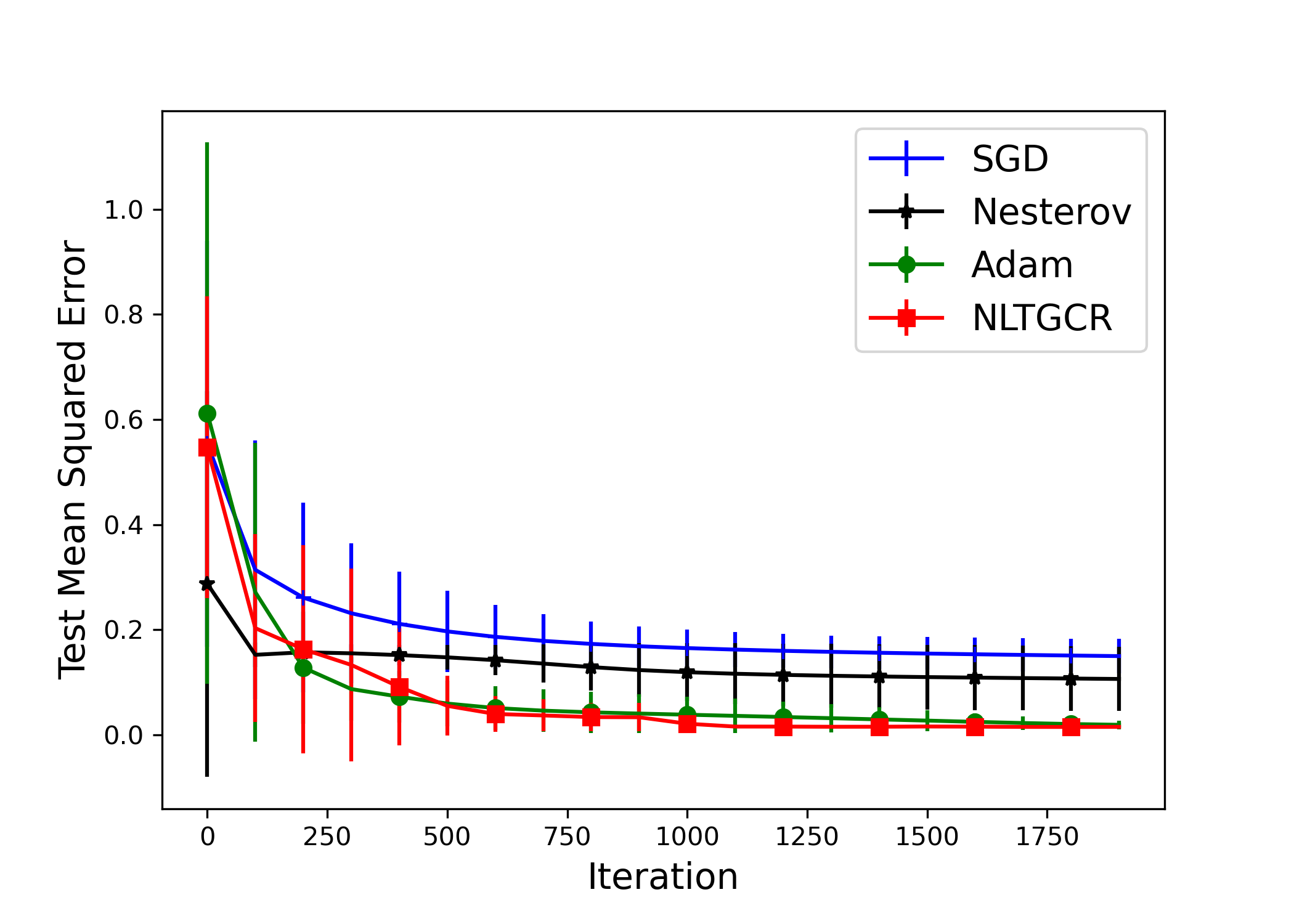}
\subcaption{Airline-passengers} 
\label{fig:lstm1DL}
\end{subfigure}
\begin{subfigure}[b]{0.45\textwidth}
\centering
\includegraphics[width=0.9\linewidth]{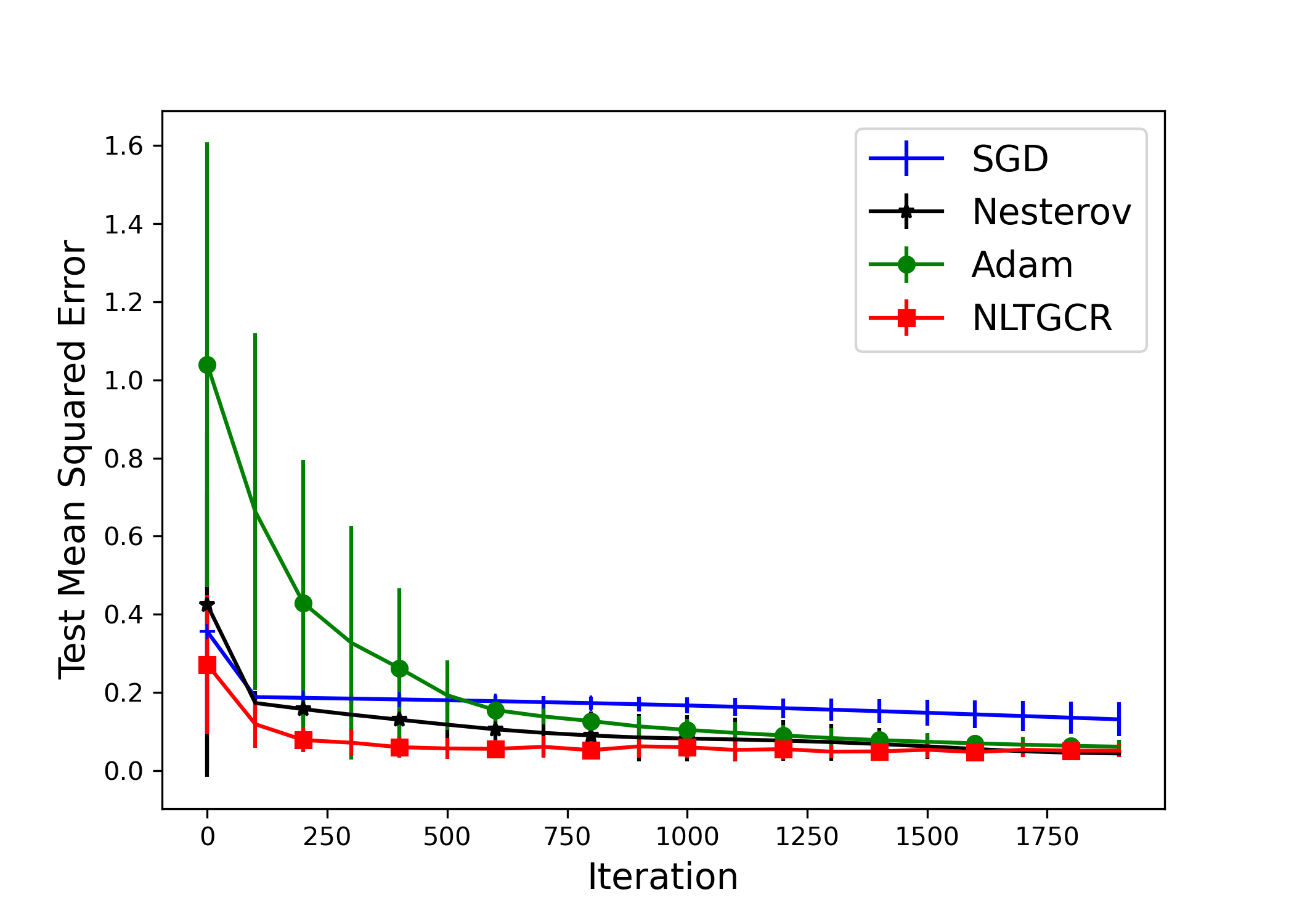}
\subcaption{shampoo} 
\label{fig:lstm2DL}
\end{subfigure}
\caption{{\textbf{Validation MSE of LSTM on two datasets.} Averaged on 5 runs, $m=1$ for our algorithm. It can be observed that nlTGCR(1) outperforms baselines.}}
\label{fig:lstmDL}
\end{figure}

                
    
\subsubsection{Semi-supervised classification of graph data using GCN}
We also test the effectiveness of nlTGCR on GCN \cite{kipf_semi-supervised_2017}  using Cora Dataset. It consists of 2708 scientific publications classified into one of seven different classes. The citation network consists of 5429 links. Each publication in the dataset is described by a 0/1-valued word vector indicating the absence/presence of the corresponding word from the dictionary.
The objective is to accurately predict the subject of a paper given its words and citation network, as known as node classification.  Hyperparameters are selected after a grid search. For Nesterov ($v=0.9$),  Adam (default $\beta_1$ and $\beta_2$), and NLTGCR $(m=1)$, we use a learning rate of $1 \times 10^{-1}$, $1 \times 10^{-2}$, and $1 \times 10^{-1}$, respectively. Figure \ref{fig:gcnLoss} and \ref{fig:gcnAcc} depict the training loss and validation accuracy averaged on 5 runs. It shows \methodName~ converges faster and achieves higher accuracy than baselines, which demonstrates the effectiveness of \methodName~ optimizing complex deep learning architectures.
\begin{figure}[ht]
\centering
\begin{subfigure}[b]{0.45\textwidth}
\centering
\includegraphics[width=0.9\linewidth,]{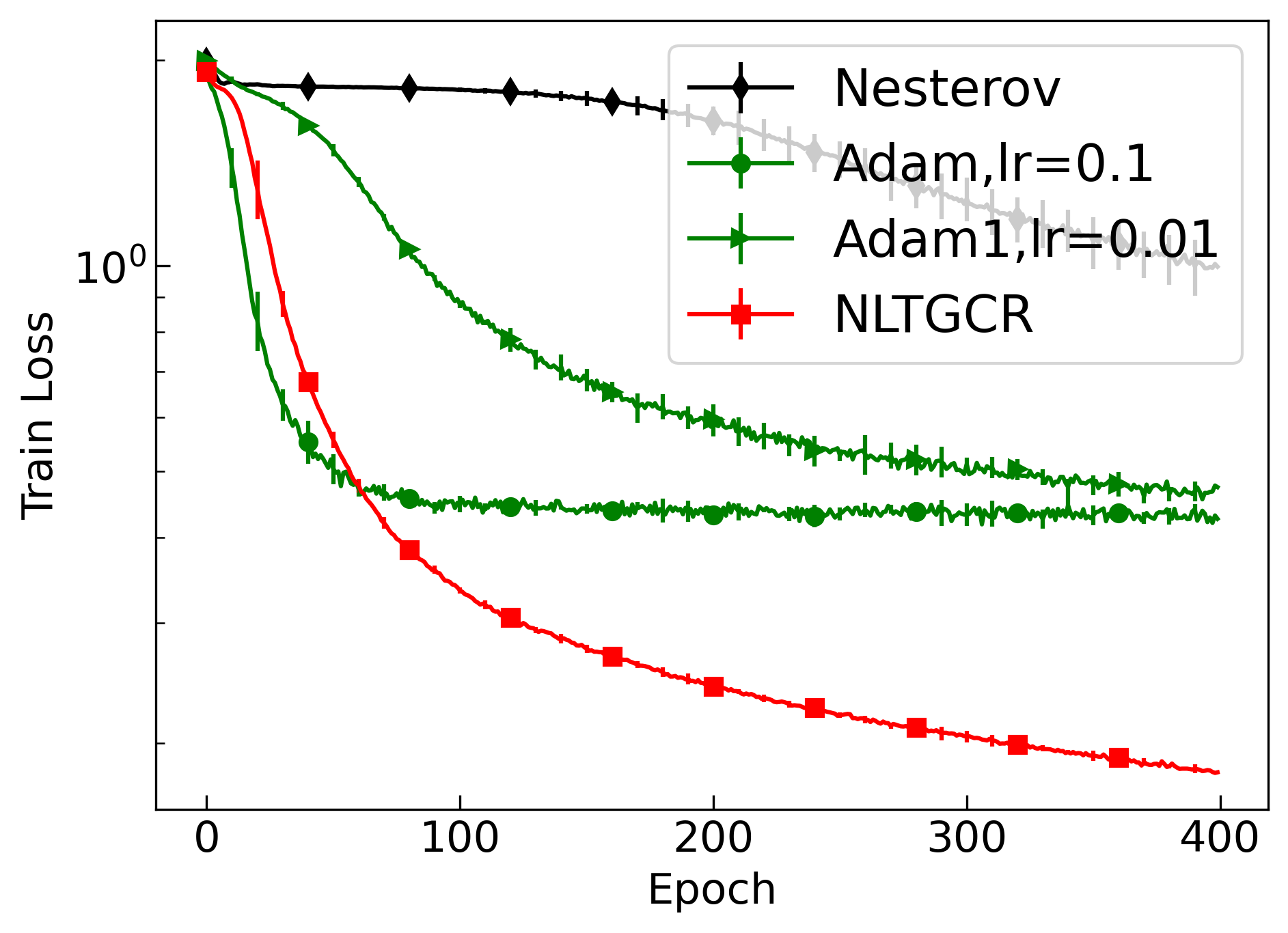}
\subcaption{Train Loss} 
\label{fig:gcnLoss}
\end{subfigure}
\begin{subfigure}[b]{0.45\textwidth}
\centering
\includegraphics[width=0.9\linewidth]{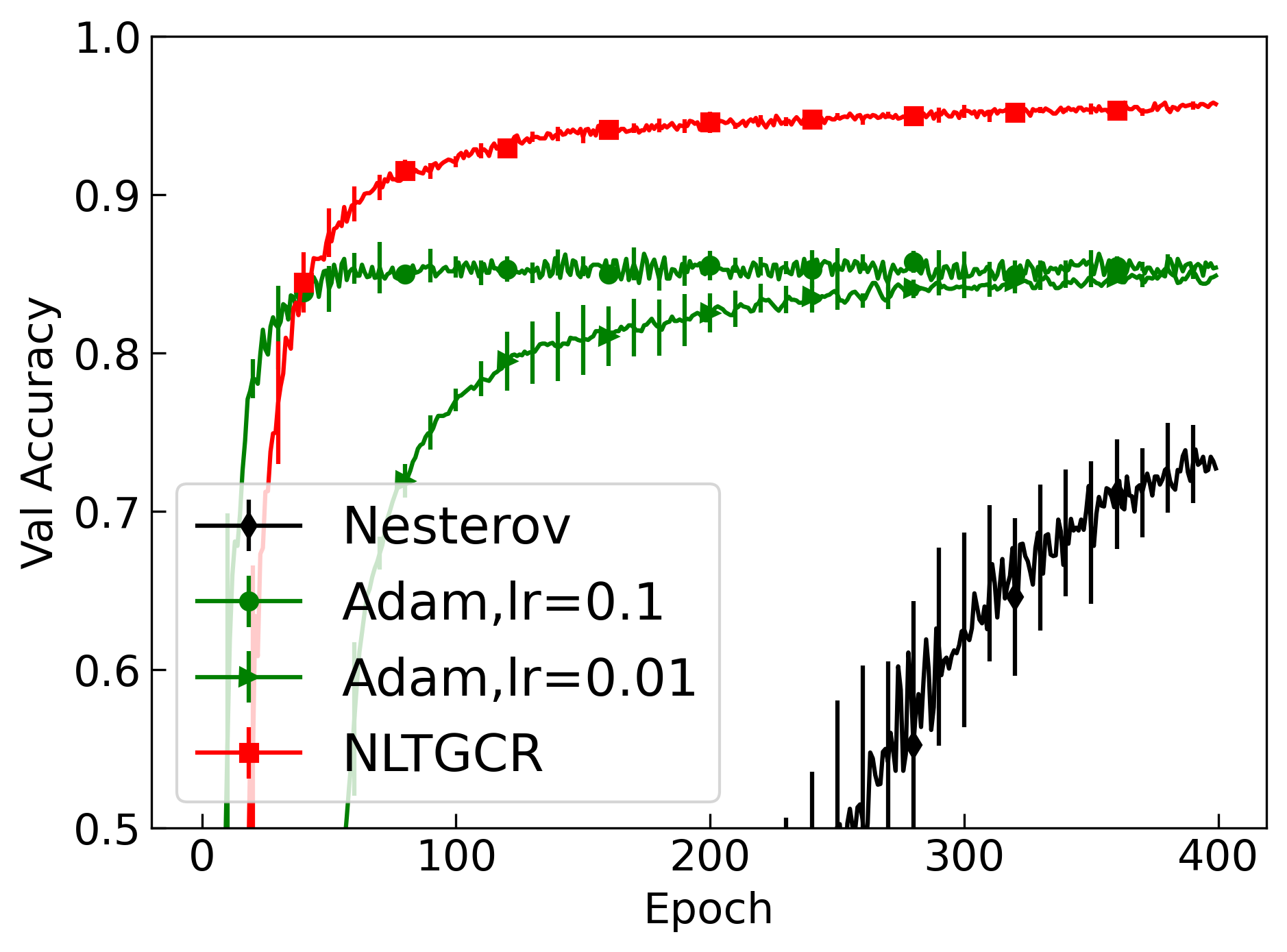}
\subcaption{Validation Accuracy} 
\label{fig:gcnAcc}
\end{subfigure}
\caption{{\textbf{Validation accuracy on CORA using GCN.} Averaged on 5 runs, $m=1$ for our algorithm. It can be observed that nlTGCR(1) significantly outperforms baselines.}}
\label{fig:GCNDL}
\end{figure}
}
 \end{appendices}
\vfill


\end{document}